\documentclass[11pt, a4paper, logo, copyright]{deepmind}

\usepackage[authoryear, sort&compress, round]{natbib}
\usepackage[capitalize]{cleveref}
\usepackage{subcaption}
\usepackage{amsmath}
\usepackage{ctable} 
\usepackage{xspace}
\usepackage{url}

\usepackage{nicematrix}
\usepackage{multirow}

\usepackage{array}

\title{Diversifying AI:\\ Towards Creative Chess with AlphaZero}

\correspondingauthor{tomzahavy@deepmind.com}

\renewcommand{\today}{2023-16-03}

\let\vv=\v 
\newcommand{\setpolicies}{\mathbf{\Pi}}

\newcommand{\cz}{\text{AZ}\textsubscript{\text{db}}\xspace}
\newcommand{\ie}{{\it i.e.}}
\newcommand{\eqdef}{=}  
\newcommand{\nplayers}{N_{\text{Players}}}
\newcommand{\ntd}{N_\text{TD}}

\DeclareMathOperator*{\argmax}{arg\,max}
\DeclareMathOperator*{\argmin}{arg\,min}

\author[a]{Tom Zahavy}
\author[a]{Vivek Veeriah}
\author[a]{Shaobo Hou}
\author[b,*]{Kevin Waugh}
\author[a]{Matthew Lai}
\author[a]{Edouard Leurent}
\author[a]{\\Nenad Toma\vv{s}ev}
\author[c,*]{Lisa Schut}
\author[a]{Demis Hassabis}
\author[a]{Satinder Singh}

\affil[a]{Google DeepMind, London, United Kingdom}
\affil[b]{Sony AI, New York, NY, 10010}
\affil[c]{University of Oxford, Oxford, United Kingdom}
\affil[*]{Work done at Google DeepMind.}
\newcommand{\Expect}{\mathbb{E}}
\newcommand{\Prob}{\mathbb{P}}

\begin{abstract}
Artificial Intelligence (AI) systems have surpassed human intelligence in a variety of computational tasks.
However, like humans, AI systems make mistakes, have blind spots, and struggle to generalize to new situations. This research investigates whether a diverse team of AI systems can mitigate these limitations and outperform a single AI by generating a broader range of ideas and selecting the most promising ones. We explore this concept in the game of chess, 
and extend  AlphaZero (AZ) to represent a team of agents via a latent-conditioned architecture, which we call \text{AZ}\textsubscript{\text{db}}. We train \text{AZ}\textsubscript{\text{db}} to generate a wider range of ideas using behavioral diversity techniques and select the most promising ones with sub-additive planning. Our experiments reveal that players in \text{AZ}\textsubscript{\text{db}} displayed superior puzzle-solving abilities, successfully tackling even the challenging Penrose positions. Their diverse chess playstyles and specialization in various openings contributed to a 50 Elo advantage over AZ.
\end{abstract}

\begin{document}

\maketitle 

\begin{figure}[h]
\vspace{1cm}
\centering
    \includegraphics[ width=0.38\linewidth]{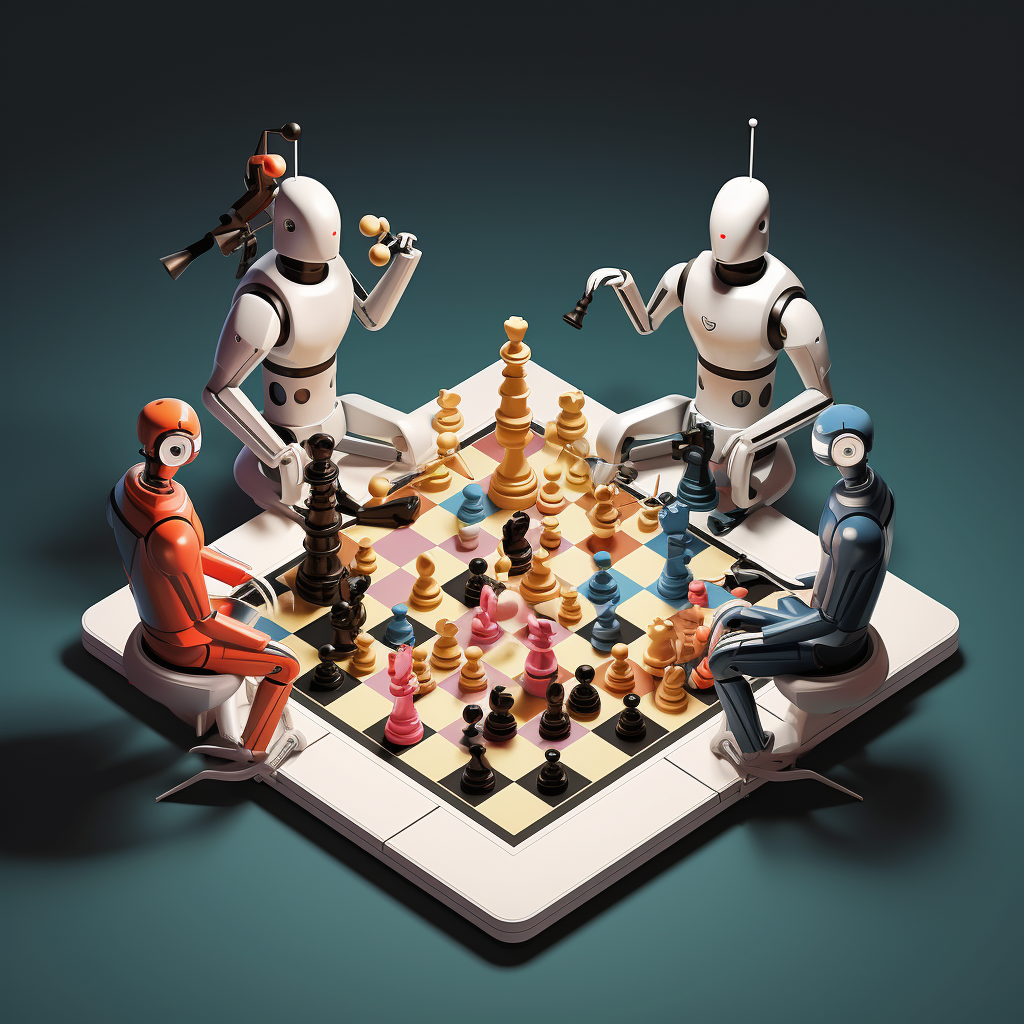}\hspace{0.1\linewidth}
    \includegraphics[clip, trim=7 0.5cm 0 0.5cm, width=0.4\linewidth]{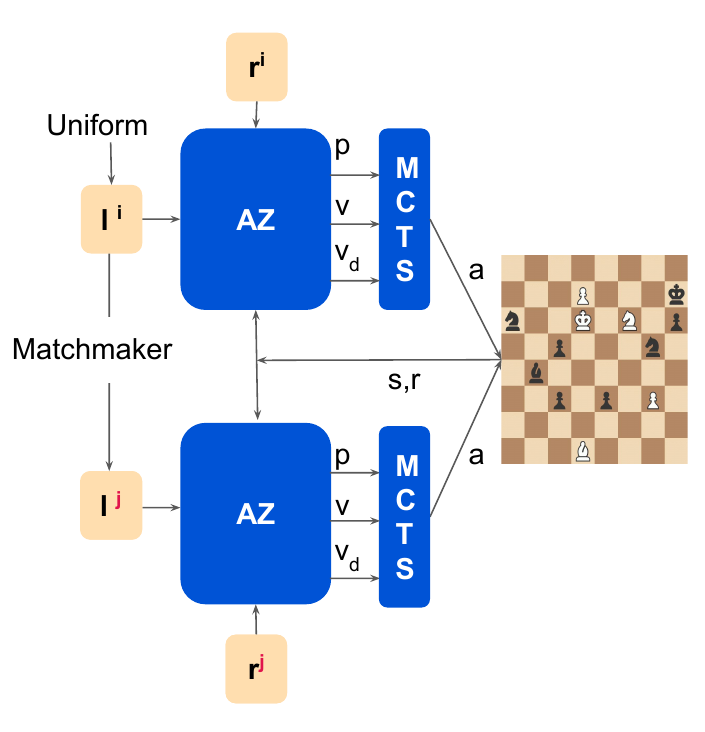}\vspace{0.5cm}
    \caption{Left: an illustration of diverse AI systems playing chess. Right: \cz architecture. For each player $i$, a matchmaker selects an opponent $j$. A latent variable $l^i$ associated with the player conditions the network outputs $p,v,v_d.$ A score for each position is computed using $v,v_d$ and the diversity intrinsic reward $r^i$ and used with the prior $p$ for MCTS.}
    \label{fig:expvic}
\end{figure}
\newpage
\tableofcontents
\section{Introduction}

Artificial Intelligence (AI) systems have achieved superhuman performance in a variety of fields, including playing games like chess \& Go \cite{schrittwieser2020mastering}, poker \cite{moravvcik2017deepstack, brown2019superhuman}, and Atari \cite{mnih2015human} , and more recently, in retrieving knowledge, reasoning over it and answering questions in medical \cite{singhal2023expertlevel}, \href{https://edition.cnn.com/2023/01/26/tech/chatgpt-passes-exams/index.html}{law, and business} domains \cite{katz2023gpt,openai2023gpt4}. AI can outperform humans in many ways; it has greater computational power and ability to experience and learn from more data. Nevertheless, AI is limited by the amount of data and compute available to it, as well as the quality of human feedback. Using Reinforcement learning (RL) AI systems can further improve by interacting with the environment, learning from trial and error, exploring new possibilities, and discovering new knowledge.

AlphaZero (AZ) \cite{az}, an RL algorithm, learned to play Go, chess, and shogi from scratch, without any prior knowledge of the games. It not only outperformed human experts, but also played moves that were considered to be creative \cite{move37b}. AZ has influenced the style of professional chess players \cite{gonzalez2022alphazero} and is associated with increased novelty in the decision-making of Go players \cite{shin2023superhuman}. Indeed, AZ has had a profound impact on the world of chess, inspiring most contemporary chess engines, such as \href{https://lczero.org}{Leela Chess Zero} (Lc0) and \href{https://stockfishchess.org/}{StockFish} (SF), to incorporate neural networks into their evaluation functions.

Despite the prowess of chess engines, we have likely not reached the pinnacle of optimal gameplay. Analyzing the outcomes of computer chess tournaments, such as the \href{https://tcec-chess.com/}{Top Chess Engine Championship} (TCEC), reveals a steady progression in the capabilities of leading engines. A key factor influencing their strength is the computational resources available for search.  For instance, picking moves according to AZ's policy network results in a 1000 ELO weaker performance compared to AZ. Similarly, superhuman AI in Go has shown susceptibility to adversarial attacks \cite{timbers2022approximate,wang2023adversarial} and \href{https://en.wikipedia.org/wiki/Anti-computer_tactics}{anti-computer tactics} highlight blind spots in computers' reasoning.

Chess puzzles present an additional challenge for computers, as they may involve counter intuitive moves, require long-term planning, or involve querying positions that deviate from the training data (out-of-distribution (OOD)). Finding positions that computers don't understand attracted the interest of researchers and chess players \cite{guid2012detecting,penrose,10positions}. While these positions become rarer over time, there remain positions that confuse modern chess engines like Lc0 \cite{steingrimsson2021chess}. In addition, despite being considered weaker players overall, humans can solve most of these puzzles, demonstrating a gap between human and machine thinking \cite{penrose}. 

This work aims to bridge the gap between human and artificial intelligence in chess by investigating  positions that humans can solve but computers find challenging. While in principle, RL agents like AZ can solve the game of chess via trial and error, in practice, the game is too large and it cannot find the truly optimal policy. Instead it finds a particular (hopefully) near-optimal solution whose choice is determined by the nature of the agent's bounds (memory, compute, training experience). This singular sub-optimal choice can potentially generalise poorly when faced with situations it did not encounter during its training. Humans, on the other hand, seem capable of considering multiple good solutions when attempting to solve a new problem instance \cite{page2019diversity}, ruling out solutions that had worked before \cite{stokes2005creativity}, and accepting the notion of failure \cite{kasparov2000kasparov}.  Diverse thinkers can outperform homogenous groups in solving complex tasks, producing “diversity bonuses” \cite{page2019diversity}: improved problem solving, increased innovation, and more accurate predictions (\href{https://priceonomics.com/how-many-people-take-credit-for-writing-a-hit-song/}{Priceonomics}, \href{https://www.wired.com/2009/09/how-the-netflix-prize-was-won/}{Wired}). This leads to a kind of creativity through diversity (of solutions) that is missing in current RL agents. We therefore investigate whether a team of diverse superhuman chess agents can solve more chess puzzles than a single super human AI. 

To test this hypothesis we designed an algorithm to train a team of high-quality, diverse AZ agents, which we call \cz. The team is represented as a single latent-conditioned architecture (\cref{fig:expvic}) such that each player $i$ in the team is represented via a latent variable $l^i$. We encourage diversity via behavioral and response diversity techniques, which complemented each other in other domains \cite{liu2021towards}. Behavioral diversity is implemented via intrinsic motivation: each player $i$ receives an intrinsic reward $r^i$ that motivates the agent to visit different positions from the others. Response diversity is implemented via a matchmaker that samples opponent $j$ for each player $i$.

We find that the players in the \cz team play chess differently: there's more variance in whether they castle long or short, develop different pawn structures, prefer different moves in common openings, and more. 

On five challenging chess puzzle data sets, \cz significantly outperformed AZ in terms of performance. Notably, \cz's single player (player $0$) performed better than AZ, and more solutions were found by different players in each position (max-over-latents). Furthermore, sub-additive planning techniques enabled efficient selection of the correct solution among the multiple solutions. In our ablations, we also observed diversity bonuses from training with different seeds, with both AZ and \cz, and that diverse teams solve more puzzles than more homogeneous teams. During our ablation studies, we also noticed diversity bonuses from training with multiple seeds, with both AZ and \\cz. Lastly, we also observed that teams with diverse members solved more puzzles compared to more homogeneous teams.

We also see diversity bonuses in chess matchups: The best player in the \cz team is $30$ ELO stronger than AZ indicating that \cz benefits from diversity bonuses via its training pipeline. But even more importantly, we observe that different players in the team specialize to different openings: With sub-additive-planning \cz is $50$ ELO stronger than AZ. 

\section{Background}
In Reinforcement Learning (RL), an agent learns how to map situations to actions so as to maximize a cumulative numerical reward signal. The learner is not told which actions to take, but instead must discover which actions lead to the most cumulative reward \cite{sutton2018reinforcement}. Mathematically, the RL problem can be written as finding a policy that maximizes the expected long term cumulative reward  $\pi = \argmax_{\pi \in \Pi} \mathbb{E} \left[\sum _{t=0}^{\infty }\gamma ^{t}r_e(s_{t+1})\right].$ $r_e(s_{t+1})$ denotes an extrinsic reward that the agent receives from the environment upon reaching state $s_{t+1}$, and $\gamma$ is a discount factor. In the case of chess, the extrinsic reward will be the outcome of the game: $-1$ if the agent losses, $1$ if it wins, and $0$ in a draw. It is also $0$ in any intermediate state and the discount factor is set to $1$. Note that the policy affects the objective by selecting actions in each state, as we will describe below, and therefore controls the distribution over next states. 

AZ uses a deep neural network $(p, v) = f_\theta(s)$ with parameters $\theta$. This neural network $f_\theta(s)$ takes the board position $s$ as an input and outputs a vector of move probabilities $p$ with components $p_a = \Pr(a|s)$ for each action $a$, and a scalar value $v$ estimating the expected outcome $z$ of the game from position $s$ ($\mathbb{E}[z|s]$). AZ learns these move probabilities and value estimates entirely from self-play (playing vs. itself); these are then used to guide its search in future games. The board $s$ is represented as an image stack composed of a concatenation of planes of size $8x8$, and also includes metadata about the position (e.g., history/castling rights, etc.). Further details can be found in \cite{silver2018general}.

AZ uses a Monte Carlo tree search (MCTS) algorithm as described in \cite{silver2017mastering}. Each state-action pair $(s, a)$ stores a set of statistics, $\{N(s, a), W(s, a), Q(s, a), P(s, a)\}$, where $N(s, a)$ is the visit count, $W(s, a)$ is the total action-value, $Q(s, a)$ is the mean action-value, and $P(s, a)$ is the prior probability of selecting $a$ in $s$. Each simulation begins at the root node of the search tree, $s_0$, and finishes when the simulation reaches a leaf node $s_L$ at time-step $L$. At each timestep, $t < L$, an action $a_t$ is selected using the PUCT algorithm \cite{rosin2011multi}:
\begin{align}
    a_t = \arg \max_a Q(s_t, a) + U(s_t, a), \label{eqn:puct_action_selection}
\end{align}
where $U(s, a) = C(s)P(s, a)\sqrt{N(s)}/(1 + N(s, a)),$ where $N(s)$ is the parent visit count and $C(s)$ is the exploration rate, which grows slowly with search time,
$C(s) = log ((1 + N(s) + c_{base})/c_{base}) + c_{init}$. Note that in case of ties, $a_t$ is selected uniformly at random from the actions that maximize Eq. (1), making the search stochastic. The leaf node $s_L$ is added to a queue for neural network evaluation, $(p, v) = f_\theta(s_L)$. The leaf node is expanded and each state-action pair $(s_L, a)$ is initialized to $\{N(s_L, a) =
0, W(s_L, a) = 0, Q(s_L, a) = 0, P(s_L, a) = p_a\}.$ The visit counts and values are then updated in a backward pass through each step $t \le L, N(s_t, a_t) = N(s_t, a_t) + 1, Q(s_t, a_t) = W(s_t,a_t)/N(s_t,a_t)$ and
\begin{equation}
    \label{eq:mcts}
    W(s_t, a_t) = W(s_t, a_t) + v.
\end{equation} 
The search returns a vector $\pi$ representing a probability distribution over moves, $\pi_a = Pr(a|s_{\text{root}})$.
Further details can be found in \cite{silver2018general}. Note that \cref{eq:mcts} is highlighted since the only change to MCTS in \cz (compared to AZ) is in the value estimate $v.$
AZ is trained by RL from self-play games, starting from randomly initialized parameters $\theta$. Each game is played by running an MCTS from the current position $s_\text{root} = s_t$ at turn $t$, and then selecting a move. To encourage exploration, in the first $30$ moves of each game, the move is selected at random by drawing a sample from a soft max distribution over the visit counts, $a_t \sim \text{SoftMax}(N(s_t,a_t))$ with temperature $1$. After $30$ moves have been played, the actions are chosen greedily $a_t = \argmax N(s_t,a_t)$ and ties are broken at random. Greedy action selection is also used in all final evaluations (puzzles and matchups), regardless of the move number. At the end of the game, the terminal position $s_T$ is scored according to the rules of the game to compute the game outcome $z \in \{-1,0,1\}$ (for loss, draw or win). The neural network parameters $\theta$ are adjusted by gradient descent to minimize the error between the predicted outcome $v$ and the game outcome $z$, and to maximize the similarity of policy vector $p$ to the probabilities $\pi$:
\begin{equation}
    \label{eq:az-loss}
    (p,v) = f_\theta(s), \ell = (z-v)^2 - \pi^T \log p.
\end{equation}

\section{Methods}
On a high level, \cz is composed of a latent-conditioned architecture (\cref{fig:expvic}), which is commonly used for modeling diverse policies, see for example, \cite{gregor2016variational,eysenbach2018diversity} for behavioral diversity and \cite{liu2022neupl} for response diversity. The architecture has a shared torso and three heads: policy, value, and intrinsic value. The policy and the value heads have identical architecture to the ones used by AZ and are trained in the same manner. The agent uses a single set of weights $\theta$ to represent the  team by conditioning the network on a latent variable $l^i$ that is associated with each player $i \in \setpolicies \eqdef \{0,\ldots, \nplayers-1\}$. We represent these latent variables as one-hot vectors and concatenate them to the input as additional constant planes \footnote{We have also experimented with FILM embeddings \cite{perez2018film} as a more expressive architecture to model player specific behaviours. However, we have not found FILM to yield significant gains over the simpler one-hot representation and have decided not to use it for simplicity}. 

Our objective is to discover a set of Quality-Diverse (QD) \cite{pugh2016quality,mouret2015illuminating} policies -- policies that play chess well but do so differently from each other. We use response and behavioral diversity techniques, which have been demonstrated to complement each other \cite{liu2021towards}. The next section explains how we implemented behavioural diversity in \cz and how we combine the policies via sub additive planning. In the supplementary material, we further provide details regarding matchmaking techniques for response diversity. \footnote{The  advanced matchmaking techniques that we explored in this paper have not provided substantial gains over the simpler self-play and PSRO \cite{lanctot2017unified} matchmakers. One potential explanation for this is that chess is a transitive game, and therefore, does not gain from more sophisticated matchmaking techniques that were designed for non transitive games \cite{balduzzi2019open}. We discuss these results in further detail in the Appendix.}

\paragraph{Behavioral diversity.}
To optimize the diversity of the policies in \cz we focus on behavioural diversity. We refer to the behavior of a policy $\pi$ as the \emph{state-action occupancy measure} $d_\pi$, which measures how often a policy visits each state-action pair when following $\pi$.
We define a diversity objective $\text{Diversity}(d_{\pi^1}, \ldots, d_{\pi^n})$ over the state occupancies of the policies in \cz. To optimize this objective, we rely on the  \textit{intrinsic reward equation}:
\begin{equation}
    \label{eq:ire}
    r_d(s,a) = \nabla_{d_\pi} f(d_\pi)(s,a),
\end{equation}
which states that one can optimise the function $f = \text{Diversity}(d_{\pi^1}, \ldots, d_{\pi^n})$ by locally maximizing its gradient as an intrinsic reward \cite{zahavy2021reward}. This approach allows us to use the AZ algorithm with only a few changes to maximize diversity. We will now provide more formal details.

Let $\Prob_\pi(s_t = \cdot)$ be the probability measure over states at time $t$ under policy $\pi$, then $d_\pi^{\mathrm{avg}}(s,a) = \lim_{T\rightarrow \infty}\frac{1}{T}\Expect\sum_{t=1}^T \Prob_\pi(s_t = s) \pi(s,a), \enspace d_\pi^{\mathrm{\gamma}}(s,a) =  (1-\gamma)\Expect\sum_{t=1}^\infty \gamma^t \Prob_\pi(s_t = s) \pi(s,a),$ for the average reward case and the discounted case respectively. It is worth noting that $d_\pi$ can be calculated for any stationary policy $\pi$, and that the RL problem of reward maximization can be written as finding a policy whose state occupancy has the largest inner product with the extrinsic reward ($r_e$), \ie , $d_\pi\cdot r_e$, which is known as the dual formulation of RL.

We use $\phi(s,a)$ to denote the feature representation of a state-action pair, which represents the chess board after taking action $a$ from state $s$. It includes the locations of the pieces on the board as well as castling rights (see examples in \cref{fig:sfs}). Note that the board representation $s$ used as input for AZ and \cz is slightly different from $\phi$, and includes history, side to play and a $50$ moves counter (see more details in the background section). We removed these features from $\phi$ because the first is redundant and the latter two occur in a predictable and relatively independent manner.

Given $\phi(s,a),$ we define $\psi = \mathbb{E}_{s,a \sim d_\pi}[\phi(s,a)]$ to be the expected features under the state occupancy $d_\pi$. With this notation, we define our objective as  maximizing a non-linear quality-diversity objective over occupancies:
\begin{equation}
    \max_{\Pi^n} \enspace \text{Diversity}(d_{\pi^1}, \ldots, d_{\pi^n}) + \sum_i \lambda_i d_{\pi^i} \cdot r_e,
\end{equation}
where $\lambda_i$ is a policy-specific scalar, $r_e$ is the extrinsic reward vector and $\text{Diversity}(\Pi^n)$ is a non-linear utility function that we will soon define. To solve this problem, our agent maximizes a reward that combines the extrinsic reward $r_e$ (win/draw/loss) and a policy-specific diversity intrinsic reward $r_d^i$:
\begin{equation}
    r(s,a) \eqdef (1-\lambda_i )r_d^i(s,a) + \lambda_i r_e(s,a).
\end{equation}
For simplicity, $\forall i>0$ we set $\lambda_i=\lambda$, but for policy $i=0$ we set $\lambda_0=1$ so it only maximizes extrinsic reward. 

The non-linear diversity utility function is defined over the expected features $\psi$ of the policies in the set. Diverse policies, under this definition, are policies that have different piece occupancies:
\begin{equation}
\label{eq:vdw}
\text{Diversity}(\Pi^n) \eqdef \sum_{i=1}^n (1-\lambda_i)\Big(\underset{\text{Repulsive}}{\underbrace{0.5 || \psi^i-\psi^{j^*_i}||_2^2 }}  \underset{\text{Attractive}}{\underbrace{- 0.2 \big(|| \psi^i-\psi^{j^*_i}||_2^5/\ell_0^3\big)}}\Big),
\end{equation}
where index  $j^*_i$ refers to the player with the closest expected features to player $i$, \ie,  $j^*_i = \arg\min_{j\neq i}||\psi^i-\psi^j||^2_2.$ The utility function, $\text{Diversity}(\Pi^n)$, is designed to balance the QD trade-off, \ie, to discover policies that are diverse and optimal. 

Simply maximizing a linear combination of a diversity objective and the extrinsic reward, often results in policies that are either optimal or diverse. Instead, Eq. (8) introduces a hyperparameter $\ell_0$ to control the QD trade-off. $\ell_0$ defines an equilibrium between two distance-dependent forces: one force that attracts the policies to behave similarly to each other and another that repulses them \cite{zahavy2022discovering}.  The quantity $||\psi^i-\psi^{j^*_i}||$ denotes the Hausdorff
distance of policy $i$ from the set of all the other policies and have  \cite{zahavy2022discovering}. 

We can see that Eq. (8) is a polynomial in the Hausdorff
distance, composed of two forces with opposite signs and different powers. The different powers determine when each force dominates the other. We note that the coefficients in Eq. (8) are chosen to simplify the reward in Eq. (7), one could consider other combinations of powers and coefficients \footnote{Ablations studies in previous work have suggested that the reward is not sensitive to the choice of specific powers \cite{zahavy2022discovering}}. 

According to the \textit{intrinsic reward equation} (\cref{eq:ire}), the gradient of \cref{eq:vdw} can be used as an intrinsic reward for maximising it. Differentiating it gives:  
\begin{equation}
    \label{eq:intrinsic_r}
    r_d^i(s,a) \eqdef
    (1-(|| \psi^i-\psi^{j^*_i}||_2/\ell_0)^3)\phi(s,a) \cdot (\psi^i-\psi^{j^*_i}).
\end{equation}
As we can see, once a set of policies satisfies a diversity degree of $|| \psi^i-\psi^{j^*_i}||_2=\ell_0$, the intrinsic reward is zero, which allows the policies to focus on maximizing the extrinsic reward. 

To maximize this reward, we learn an intrinsic value function $v_d$ (in addition to the policy and value functions) by first computing an intrinsic value target $z_d(s_t)$ and then adding a loss $(z_d-v_d)^2$ to \cref{eq:az-loss}.  To compute the target we only take into account diversity rewards for the player that is currently playing. This is implemented by zeroing out the reward in all of the opponent transitions and using an even number of TD steps $\ntd$ when computing the intrinsic value target $z_d(s_t) \eqdef \sum_i^{\ntd/2} r_d(s_{t+2i}) + v_d(s_{t+\ntd})$ (as black and white alternate moves in chess). 

Similarly, during MCTS, the value estimate $v$ in \cref{eq:mcts} is modified in two ways. Firstly, each value estimate is replaced by a combination of the intrinsic and extrinsic value estimates: $v = \lambda^i v^i + (1-\lambda^i)v_d^i$. Secondly, in each MCTS simulation, we accumulate all the intrinsic rewards from the root to the leaf and add them to $v$. We also zero out the intrinsic reward in states that correspond to the turn of the other player as we did for learning. Furthermore, during MCTS the same latent (associated with the player in the root) is used throughout the entire search, \ie, the planning agent searches against itself. This latent conditions the value, prior and intrinsic value as well as the intrinsic reward. Therefore, the planning player does not have access to privileged information about the opponent.

\paragraph{Sub-additive planning.}
To choose a player from its team to solve a chess puzzle, \cz produces $\nplayers$ moves, one for each player. The first method,  which we call \textit{max-over-latents}, is an oracle that selects the player who has successfully solved the puzzle. If at least one player has solved the puzzle, the score is one. Teams that propose diverse and creative ideas that are also useful have high \textit{max-over-latents} scores.

The \textit{max-over-latents} score, however, is not a feasible method, because the solution to a puzzle is obviously unknown and it cannot be used to select the player who solved it. Instead, we propose a method to choose a player without seeing the solution. We limit our discussion to sub-additive planning methods that involve two steps. (i) a \textit{planning} step, in which each player $ i \in \Pi = \{ 0, \cdots, \nplayers - 1 \} $ performs MCTS and returns the following search-statistics: 
\begin{align*}
    \eta^{\text{Players}} = \{ &N^{i}(s, a), Q^i(s, a), U^i(s, a) \} _{i=0}^{\nplayers-1},
\end{align*} 
where $N$ stands for visitation counts, $Q$ for value estimates, and $U$ for exploration terms (see the background section for more details). (ii) a \textit{selection} step, in which a player is chosen based on the search statistics $\eta^{\text{Players}}$. In other words, sub-additive planning uses a heuristic to decide which player has the best chances to solve the puzzle, similar to how humans use heuristics as a way of reducing the complexity of decision making \cite{gigerenzer1991tools}.
These two steps can be written together as:
\begin{align*}
    i,\ \ \text{score}_i \gets \texttt{sub-additive-fn}(\eta^{\text{Players}}),
\end{align*}
where $\texttt{sub-additive-fn}$ outputs a player's index $i$ and its score in a puzzle $score_i$. Let $V^j(s) = \max_{a} Q^j(s, a)$, the player's selection is based on one of the following rules:
\begin{align}
\label{eq:sub-additive}
    \text{(Visit) :} i & \gets \argmax_{j} \ \ \max_{a} N^j (s, a).\nonumber\\
    \text{(Value) :}  i & \gets \argmax_{j} \ \ V^j(s).\nonumber\\
    \text{(LCB) :} i & \gets \argmax_{j} \ \ \max_{a} Q^j(s, a) - U^j(s, a). \nonumber\\
    \text{(Gap) :} i & \gets \argmin_{j \in \{j: V^j(s) \ge \argmax_{k} V^k(s) - \text{Gap}(s) \}} \ \ V^j(s).
\end{align}
These equations correspond (top to bottom) for selecting a player using visit counts, values, and Lower-Confidence-Bound (LCB) values. The LCB is obtained by switching the sign of the exploration term in the PUCT action-selection formula in \cref{eqn:puct_action_selection}. The Gap method defines a set of candidate policies -- all the policies whose value is within a $\text{Gap}(s)$ from the maximal value policy -- and  selects the policy with the lowest value in this set. Since the gap sets a confidence interval around the value, we selected $\text{Gap}(s)$ to be $\max_{a,j} U^j(s, a)$ (where U is AZ's exploration bonus, defined in the background section). Empirically, we observed that the gap tends to be a small number, making this action selection mechanism similar to the Value action selection. Yet, it seems to help in positions with low confidence. Note that the maximization in the gap is done over both players and actions, but the $\arg\min$ is taken only over the players. Therefore, it can be viewed as a technique to reduce over estimation, similar to double Q learning \cite{van2016deep}.

The cumulative sum of solutions in the team (\textit{max-over-latents}) always upper bounds the performance of any sub-additive planning method (it also explains why these methods are called sub-additive). In addition to that, a good sub-additive method should perform better than the best player in the team. In the case of \cz this is usually player $0$, which is trained without intrinsic reward ($\lambda^0 = 1$). For these reasons, we will always compare sub-additive planning with \textit{max-over-latents} and the performance of player $0$. 

\section{Experiments}
Our experiments are destined to answer the following questions:
\begin{enumerate}
    \item Do \cz's policies play chess differently from each other? 
    \item What makes puzzles hard for AZ and why?
    \item Do diversity bonuses emerge in \cz when solving puzzles and playing chess? 
\end{enumerate}

We begin by presenting technical details of our experimental setup. Next, we delve into question 1 in \cref{subsec:diverse_chess} by visualizing and examining the piece occupancy of \cz  and analyzing the opening preferences of its players. In \cref{subsec:puzzles_hard} we explore question 2 using a comparative analysis approach. We train different variations of AZ that can start from puzzle positions and gradually relax the exploration and generalization challenges at varying levels. We compare the puzzle-solving rates of these variants with those of the original AZ and visualize their understanding of the puzzles by presenting their predictions (policy, value, and planned value) for selected positions. Lastly, in \cref{subsec:div_bonus}, we address question 3 by examining the puzzle-solving rate and playing strength of \cz.

\paragraph{Setup.}
We trained AZ and \cz while keeping almost all the hyperparameters identical to those used in \cite{silver2018general}. We now explain which hyperparameters were changed in each of our training configurations.
Firstly, AZ's state representation includes history states, which are not available in puzzles (which typically only have a position to solve and not the preceding states). To address this, we trained AZ with history dropout, which randomly removes history states with a probability of $0.2$.

The \textit{full configuration} trains a team of $10$ players. It uses a batch size of $4096$ during training and is trained for $1.6$M training steps. The agent is trained with a distributed actor-learner setup. It plays a number of asynchronous self-play games and these games are stored in a replay buffer, along with the policy and raw value targets for each state transition from the game. These targets are used to train their respective estimates. In this configuration, the agent only stores 10\% of state transitions from those games (randomly chosen) in the replay buffer and discards the rest. The replay buffer can store at most $1$M positions. At each training step, the agent samples a batch of $4096$ samples to make one training update. More details can be found in the Appendix.

The \textit{fast configuration} is identical to the full configuration, except for the following three changes: the team is of size $5$, the batch size during training is reduced to $1024$ and all positions generated from self-play games are stored in the replay buffer, \ie, no positions are discarded. This configuration allows an experiment with \cz to complete in less than a day compared to a week with the full configuration, but has significantly weaker performance in both puzzles and matches. We use this configuration to conduct a number of careful ablations on \cz. We evaluate it with 1 million MCTS simulations in each position, unless otherwise specified.

To provide a comparable AZ baseline, we retrained AZ using these exact configurations. AZ trained with the full configuration, performed as well as the published AZ \cite{silver2018general} in terms of win rate against the SF8 reference player. Concretely, AZ and \cz in this paper play the same number of games and learn with the same algorithm. But for \cz, these games are shared played between different players. 

We use the PSRO \cite{lanctot2017unified} matchmaker throughout all of our experiments. In the supplementary material we present a detailed study comparing the different matchmakers. However, we did not find the more advanced match makers to outperform the simpler ones across the puzzle data sets, and the decision to use PSRO is arbitrary. One potential explanation for this is that chess is a transitive game, and therefore, does not require clever matchmaking techniques that were designed for none transitive games. 

\begin{figure}[b]
\centering
\vspace{-0.4cm}
    \includegraphics[clip, trim=8.5cm 0 6.5cm 0, width=0.85\linewidth]{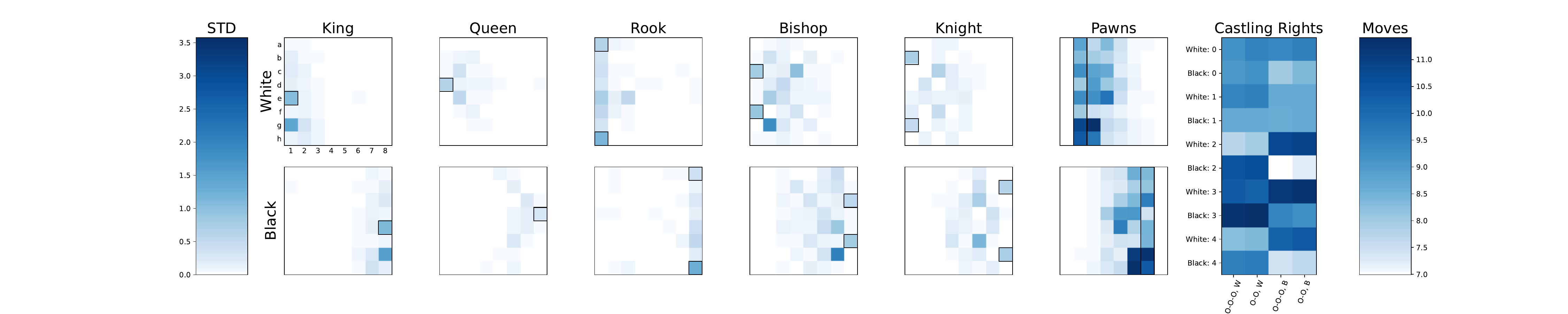}
    \vspace{-0.2cm}
    \caption{Left: Standard deviation across players of the piece occupancies of \cz. Black rectangles highlight the starting position of each piece. Right: average Castling Rights of different players. x-axis corresponds to castling to the Queen (O-O-O) or King (O-O) side as White (W) and Black (B), y-axis corresponds to players. Color units (in the color bars) correspond to occupancy in $\%$ multiplied by the average number of moves in a game (65). For orientation, the chess coordinate system is displayed on the left top. }
    \label{fig:sfs_std}
\end{figure}

\subsection{Do \cz policies play different chess?}
\label{subsec:diverse_chess}
To answer this question, we examined the piece occupancies of the different \cz  policies and assessed their opening diversity. Recall that the expected feature vectors $\psi = \mathbb{E}_{s\sim d_\pi}[\phi(s)]$ represent the features that a policy $\pi$ observes on average when playing chess. The features $\phi$ represent the chess board and also include castling rights. Our diversity objective is designed to discover policies with different associated vectors $\psi$ (\cref{eq:intrinsic_r}). Therefore, we now inspect these vectors to gain insights on the behavior of \cz's policies.

\begin{figure}[t!]
\centering
    \includegraphics[width=\textwidth]{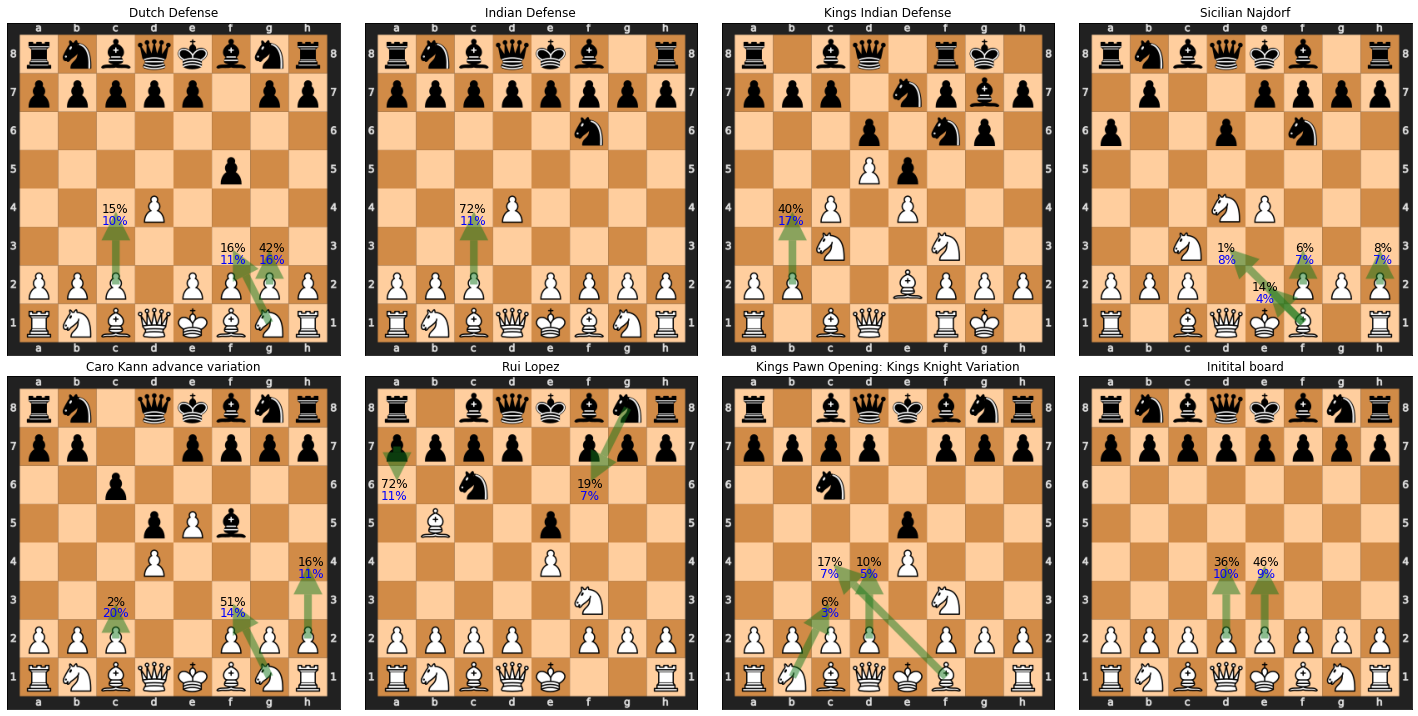}
    \caption{Opening diversity with \cz. We selected $8$ popular openings where humans tend to play different moves and present with green arrows the moves chosen by \cz policies with LCB action selection and $100k$ simulations. For each move, we present in black numbers (top) the $\%$ in which this move was played by chess GMs (taken from the Lichess Masters database), and in blue numbers (below them) the relative win percentage ($\%$ wins - $\%$ losses).}
    \label{fig:opening_diversity_8}
\end{figure}

In \cref{fig:sfs_std} we present standard deviation (across players) of vectors $\psi$. We also present the raw vectors for all the players and a mean subtracted visualization in the Appendix (\cref{fig:sfs} \& \cref{fig:sfs_diff}). These Figures show notable differences in the behavior of the players. Firstly we can see high variability in castling rights on \cref{fig:sfs_std}, right. Players $2$ \& $4$ have lower occupancy for castling, which means that they either castle earlier or move the king or rook so they lose their castling rights earlier in the game. Inspecting \cref{fig:sfs_diff} in Appendix we can indeed see that when player $2$ plays as white, its king occupies $g1$ for more moves (the square where it gets to after castling kingside). Player $3$ on the other hand, castles later, and occasionally castles queenside. In \cref{fig:sfs_diff} we see that it spends more time on $e1$.

Inspecting the pawn structure in \cref{fig:sfs}, we can see that most players push the $e$ and $d$ pawns to take control over the center. Unlike the other players, player 5 favours the Fianchetto and pushes the $g$ pawn ($g2-g3$) to get its kingside bishop to $g2$ (\cref{fig:sfs}). We can also see most players push the $a$ and $h$ pawns more than the $b,f,g$ pawns and that the $c$-pawn is pushed fairly frequently too (the $c2$ square is often lighter than other squares).

There is also variability across players in the average time that each piece gets to stay on the board (see the black numbers above each sub-figure in \cref{fig:sfs}). For example, the white queen stays on the board $48\%$ of the game with player $0$ as white and $52\%$ with player $4$ as black.  The rook is the piece that survives the most, $62\%$ with player $3$ but only  $60\%$ with other players. We can also see that rooks tend to stay on the first rank and the second most common rank for them is the 7th (with lower occupancy). One exception is player $2$ that has a rook at the 3rd rank. 

To study the \textbf{opening diversity} of \cz we performed two analyses. Firstly, we study the diversity in $8$ popular chess opening positions in \cref{fig:opening_diversity_8}. Inspecting the results, we can see that \cz's players play different moves. All of these moves are played by chess grandmasters (GMs) at the highest levels, although some of them are considered niche (eg. 6. Bd3 in the Najdorf, 4. c3 in the Caro-Kann advanced variation). We report the percentage in which each move was played by GMs (black numbers) as well as their relative win rate (blue numbers).

For a quantitative evaluation, we measured the number of different moves played by \cz's policies in $3233$ openings from the Encyclopaedia of Chess Openings using \href{https://github.com/lichess-org/chess-openings}{lichess-org chess-openings} (we excluded irregular openings (A00)). \cz discovered $1.26 \pm 0.01$, $2.45 \pm 0.02,$ and $1.65 \pm 0.01$ moves on average in each position, for $1, 400, 100k$ simulations respectively. Note that \cz discovered the most opening moves when it used $400$ simulations, which is the number of simulations it used for training. We further discuss this finding at the end of this section.

\subsection{Why are puzzles hard for AZ?}
\label{subsec:puzzles_hard}
In the previous section, we observed that the players in the \cz team play chess differently from each other. We would like to evaluate these players on challenging puzzles and examine if diversity bonuses emerge. But before we do so, we examine why chess puzzles make a challenge for computer chess agents like AZ. 

For this goal, we have collected two data sets of challenging puzzles that can trick modern chess engines. The \textit{Challenge set} includes $15$ multi-step puzzles that are challenging to modern chess engines, and the \textit{Penrose set} includes $53$ challenging puzzles including fortresses and Penrose positions to measure the value accuracy of AZ. More details about these data sets can be found in the Appendix. 

In the first row of the top panel in \cref{table:train_on_test_results} we can see the solve rate of AZ with $1M$ MCTS simulations in these data sets; it only solved 11.76\% of the Challenge set and 3.64\% of the Penrose set. Although AZ is a very strong chess player, these puzzles confuse it. We now attempt to understand what makes these puzzles difficult for AZ through a few experiments and visualizations. Note that for all the experiments in this section, we used the fast configuration of AZ (see setup at the beginning of this section). 

\begin{table}[t!]
\centering
    \begin{tabular}{p{0.46\linewidth} | p{0.2\linewidth} p{0.17\linewidth} } 
     \specialrule{.2em}{.1em}{.1em} 
      & Challenge Set &  Penrose Set \\
     \specialrule{.2em}{.1em}{.1em} 
     (1) AZ & 11.76\% & 3.64\% \\
     (2) Self-play from puzzles & 76.47\% & 96.23\% \\ 
     \specialrule{.2em}{.1em}{.1em} 
     (3) In distribution & --- & 88.24\% \\
     (4) Out of distribution & --- & 11.76\% \\
     \specialrule{.2em}{.1em}{.1em} 
     (5) + intermediate positions & 88.23\% & 96.23\% \\
     (6) + exploration & 94.11\% & 96.23\% \\
     (7) + half-move clock & 94.11\% & 100\% \\
     \specialrule{.2em}{.1em}{.1em} 
    \end{tabular}
\caption{Self-play from puzzle positions. From top to bottom.  (1) AZ. (2) AZ with self-play from puzzle positions. (3) Randomly splitting the Penrose set to train and test sets. (4) Training on the first Penrose position \& testing on the second (OOD). (5-7) AZ with self-play from puzzle positions and additional exploration techniques.}
\label{table:train_on_test_results}
\end{table}

Our first hypothesis is that these puzzle sets present a hard generalization challenge for AZ, highlighting its weaknesses and blind spots. To verify this prediction, we allow AZ to begin self-play games from puzzle positions. It is important to note that the solution is not provided to AZ; only the starting position of the puzzle. By presenting the position to AZ, it can learn to solve the puzzle by trial and error and potentially improve at doing so via its RL pipeline.
In the second row of  the top panel in \cref{table:train_on_test_results} we can see the solve rate of AZ when it begins self-play games from the puzzle positions. We can see that it solved most of the Penrose set and significantly improved its performance on the Challenge set (from 11.76\% to 76.47\%). 

We further explore if this generalization challenge is in distribution generalization or OOD generalization in the middle panel of \cref{table:train_on_test_results}. For in distribution generalization, we randomly split the Penrose set to train and test sets. We allow AZ to begin self-play from the train set and evaluate its performance on the test set. In the OOD case, we train on one Penrose position and evaluate on the other (details in Appendix). We can see that AZ generalizes pretty well in distribution while it struggles to generalize to OOD positions.

However, it remains unclear what prevents AZ from reaching 100\% solve rate when it begins self-play games from these puzzle positions. Recall that many of these puzzles involve a long sequence of actions and an incorrect action at any point results in the agent failing to solve the puzzle. Thus, we hypothesize that these positions also present an exploration challenge to AZ.
To verify this second hypothesis, we performed three more experiments which we present in the lower panel of \cref{table:train_on_test_results}. We only briefly explain them here and refer the reader to \cref{subsec:details_train_on_test} for more details. (1) we extended the puzzle positions from the Challenge set to also include intermediate positions (76.47\% $\rightarrow$ 88.23\% on the Challenge set), (2) we change AZ's action selection mechanism to sample actions from a softmax distribution over the MCTS visits (88.23\% $\rightarrow$ 94.11\%) and (3) we manually augmented the half-move counter (100\% in the Penrose set). Each one of these changes have improved AZ's performance on the challenging puzzles until it finally solve all but one of them. 
 
We note that these puzzles were collected and composed by chess players deliberately to trick computers and it's not likely that AZ had seen such positions when it learned to play chess by playing against itself. Our findings in this section suggest that the challenges for AZ are OOD generalization and exploration. In the next section we will inspect if different players in the \cz team generalize differently and solve more puzzles as a result. However, we would like to emphasize that there is no fundamental or incomputable problem to solve these puzzle positions. The challenges are OOD generalization and exploration which are well known challenges in AI.

\paragraph{AZ's understanding of puzzles.}
In \cref{fig:puzzles_hard_p1} and \cref{fig:puzzles_hard_p2} in the Appendix, we visualize AZ's policy $ p(s) $, raw value $v(s)$, and MCTS value estimates $ Q(s, a) $ for four puzzle positions. We compare the estimates of AZ with the fast configuration with the variation of AZ that begins self-play games from the puzzle positions. 
\cref{fig:puzzles_hard_p1} shows two Penrose positions from the Penrose set and \cref{fig:puzzles_hard_p2} in the Appendix shows two steps from a puzzle from the Challenge set (recall that these are multi-step puzzles and to get a score of one the agent has to find the correct solution in all the steps). 
The right column in these figures shows three bar plots: the prior policy's probability $p$ of legal actions from the given position (top), the raw values $v$ (center), and the MCTS values $ Q$ (bottom). We use blue arrows and blue x-ticks to indicate correct actions, and red for incorrect ones.

Inspecting the first row in \cref{fig:puzzles_hard_p1}, we see one of the Penrose positions. It is white's turn to play, and there are only two legal actions available (Kb2 and Kb1). We can see that the prior is roughly uniform across the legal actions, which is fine since both of them are correct. However, when we inspect the raw values (shown in black bars) we see that AZ estimates the white player to win when it plays either of these two actions. After MCTS, the value estimate reduces and predicts a draw, which is the correct evaluation. The self-play from puzzles agent (blue) accurately predicts the position to be a draw in both raw and MCTS values indicating better understanding.

The second puzzle in \cref{fig:puzzles_hard_p1} is the original Penrose position (white's turn to play). The solution to this puzzle is to move the white king to any of its neighboring white squares (blue color) so that it can eventually draw the game. Any other move leads to a losing position (red color). Inspecting the figure, we see similar findings to the previous puzzle: AZ is confused regarding the correct move to play. The prior puts an equal probability on any of the legal actions (c.f. prior barplot), raw value predicts a win (c.f. raw value barplot), and MCTS value predicts a loss (c.f. MCTS value barplot). The self-play from puzzles agent, on the other hand, have accurate raw and MCTS values for this position (a draw for the correct actions and a loss for the incorrect actions), demonstrating  understanding. 

To summarize, in \cref{fig:puzzles_hard_p1} (and \cref{fig:puzzles_hard_p2} in the Appendix), we see that AZ fails to understand positions from the Penrose and the Challenge sets when trained with the fast configuration. AZ's ability to understand difficult chess positions (in terms of prior, raw value, and MCTS value) improves when it is allowed to train from those puzzle positions, which also translates to better performance on those puzzle sets.

\begin{figure}[t]
\centering
    \includegraphics[width=0.5\linewidth]{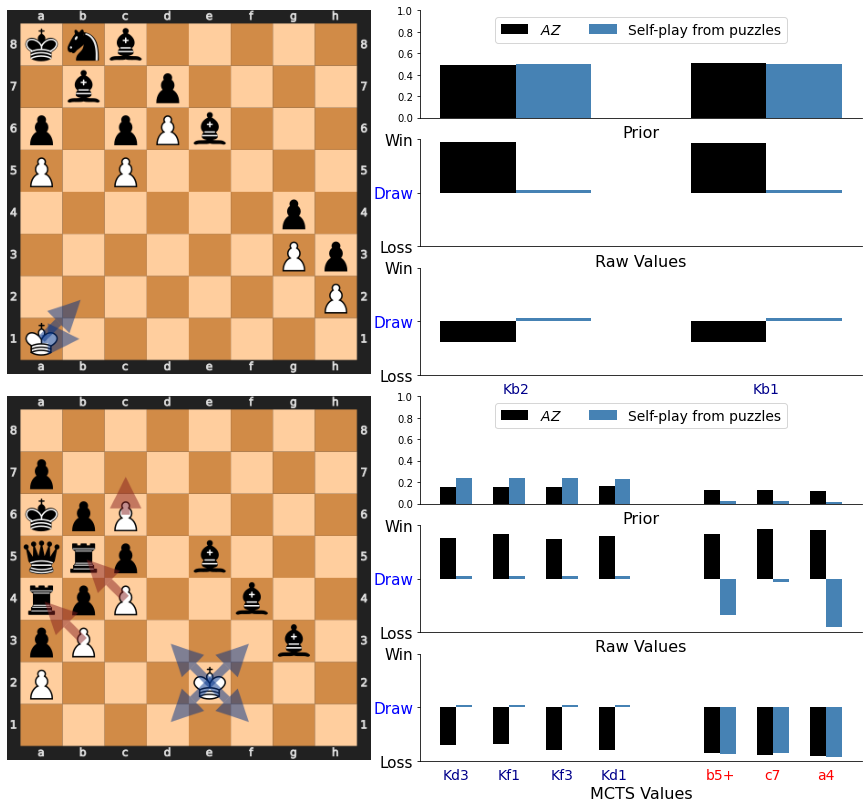}
    \caption{{\em Left:} Two Penrose puzzle positions from the Penrose set. {\em Right:} Visualizations of prior probability from a subset of legal actions, and its corresponding raw value and MCTS value estimates for AZ and AZ trained on puzzle sets. Correct and Incorrect moves are colored in Blue and Red respectively.}
    \label{fig:puzzles_hard_p1}
\end{figure}

\subsection{Diversity bonuses in \cz.}
\label{subsec:div_bonus}
In the previous sections we have seen that \cz trains a team of strong chess players that play chess differently. We have also seen that chess puzzles are a challenging problem for strong chess agents like AZ. In this section we examine if diversity bonuses emerge in \cz when solving chess puzzles. We then let \cz to play chess games against AZ from different opening positions and see if diversity bonuses emerge there and conclude with ablative analysis on diversity bonuses. From this point on, the puzzle sets are not available to any of the agents and they are not allowed to start self-play games from them.

\paragraph{Diversity bonuses in solving puzzles.}
We now compare the performance of AZ and \cz on five data sets: the Challenge and Penrose sets from before, and three larger scale data sets. STS contains around $1000$ multiple-choice puzzles, while the Lichess data set consists of roughly $500$ multiple-step puzzles from  \href{https://database.lichess.org/#puzzles}{lichess.org} that were identified to be more challenging to chess engines. Lastly, the \href{https://talkchess.com/viewtopic.php?t=72902}{Hard Talkchess} data set contains $200$ puzzles that were collected by computer chess enthusiasts and were found to be challenging to different chess engines over the years. More information about these data sets is provided in the Appendix.

\begin{figure}[t!]
\centering
\includegraphics[width=0.31\linewidth]{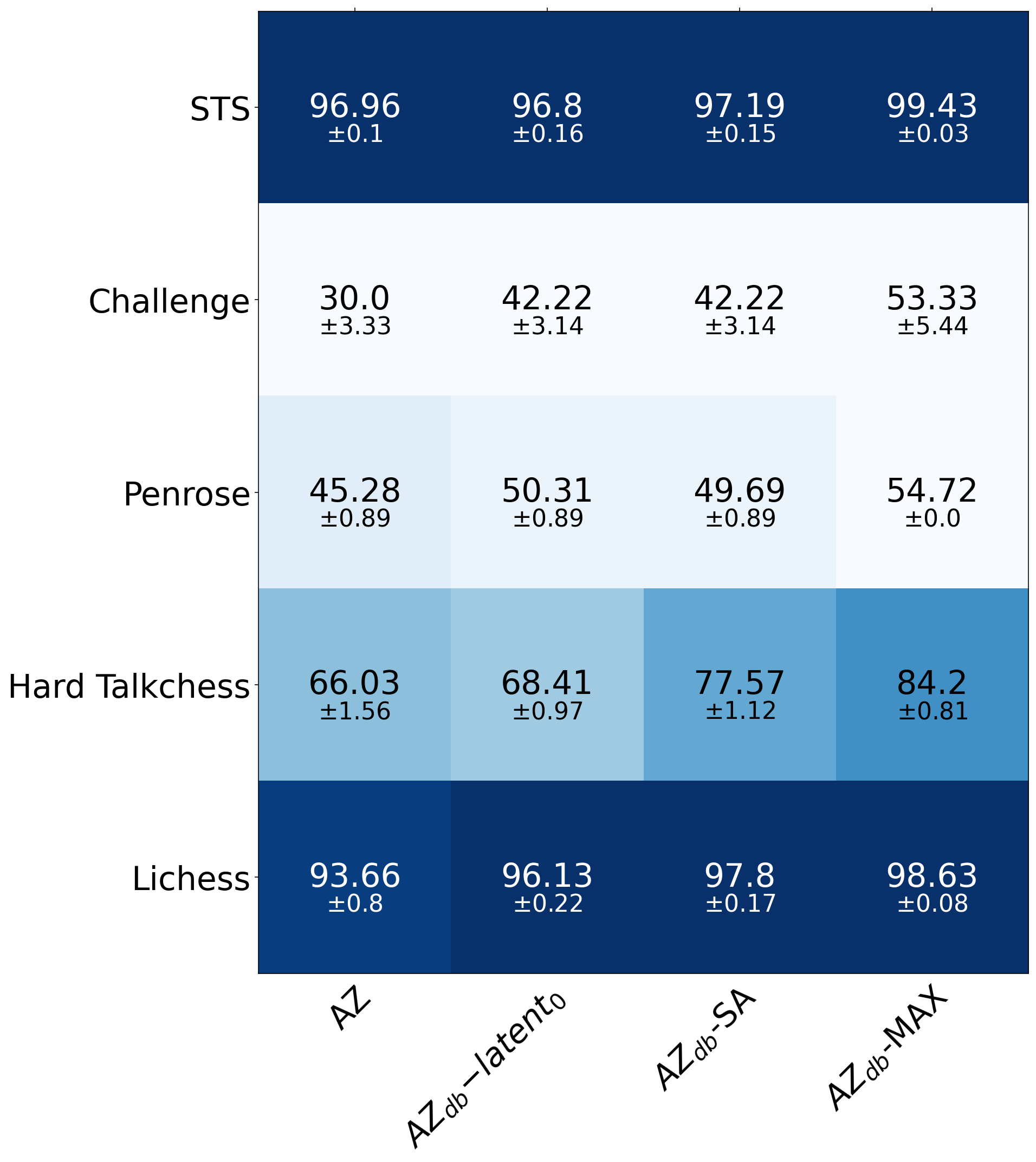}
\includegraphics[width=0.237\linewidth]{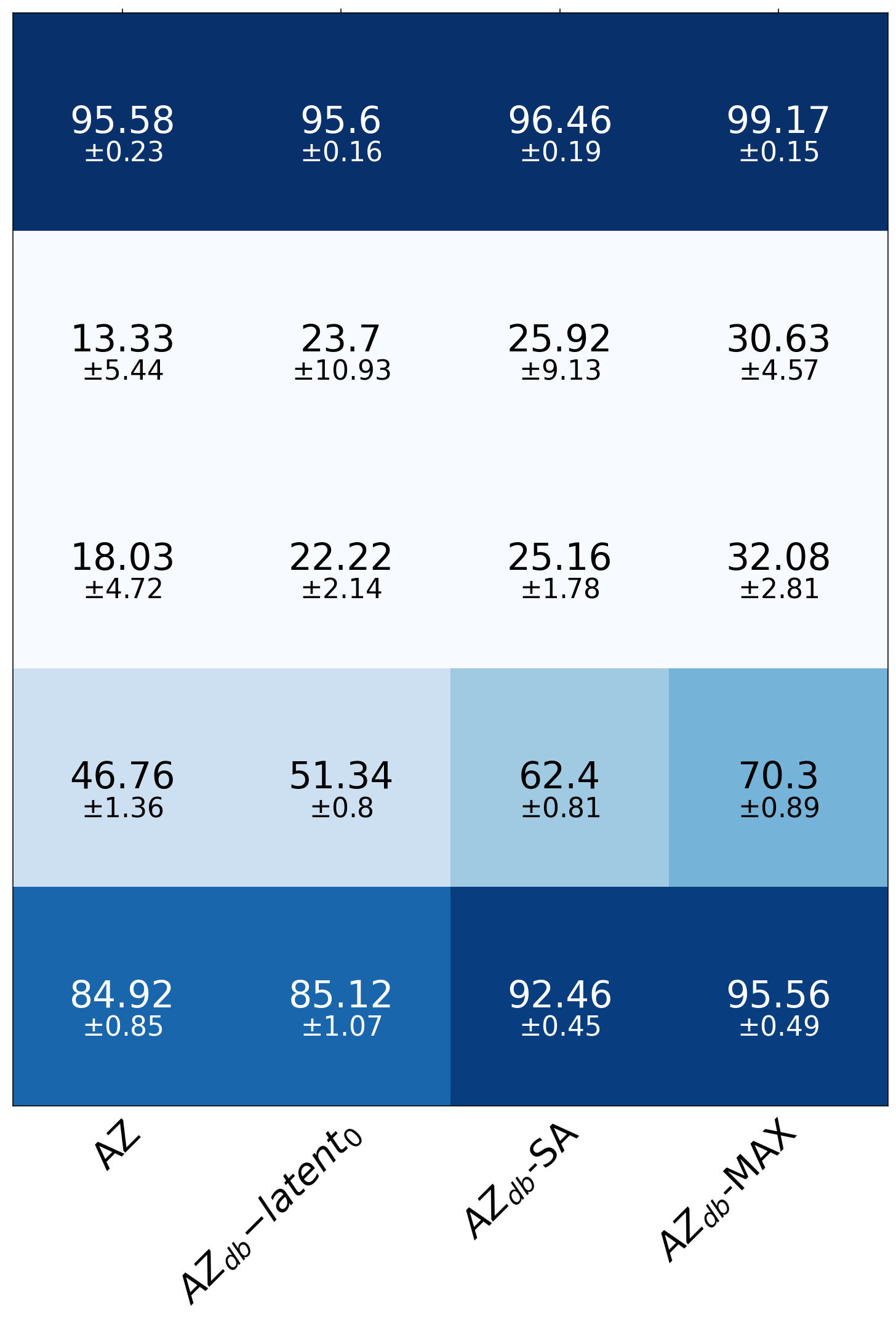}
\includegraphics[width=0.361\linewidth]{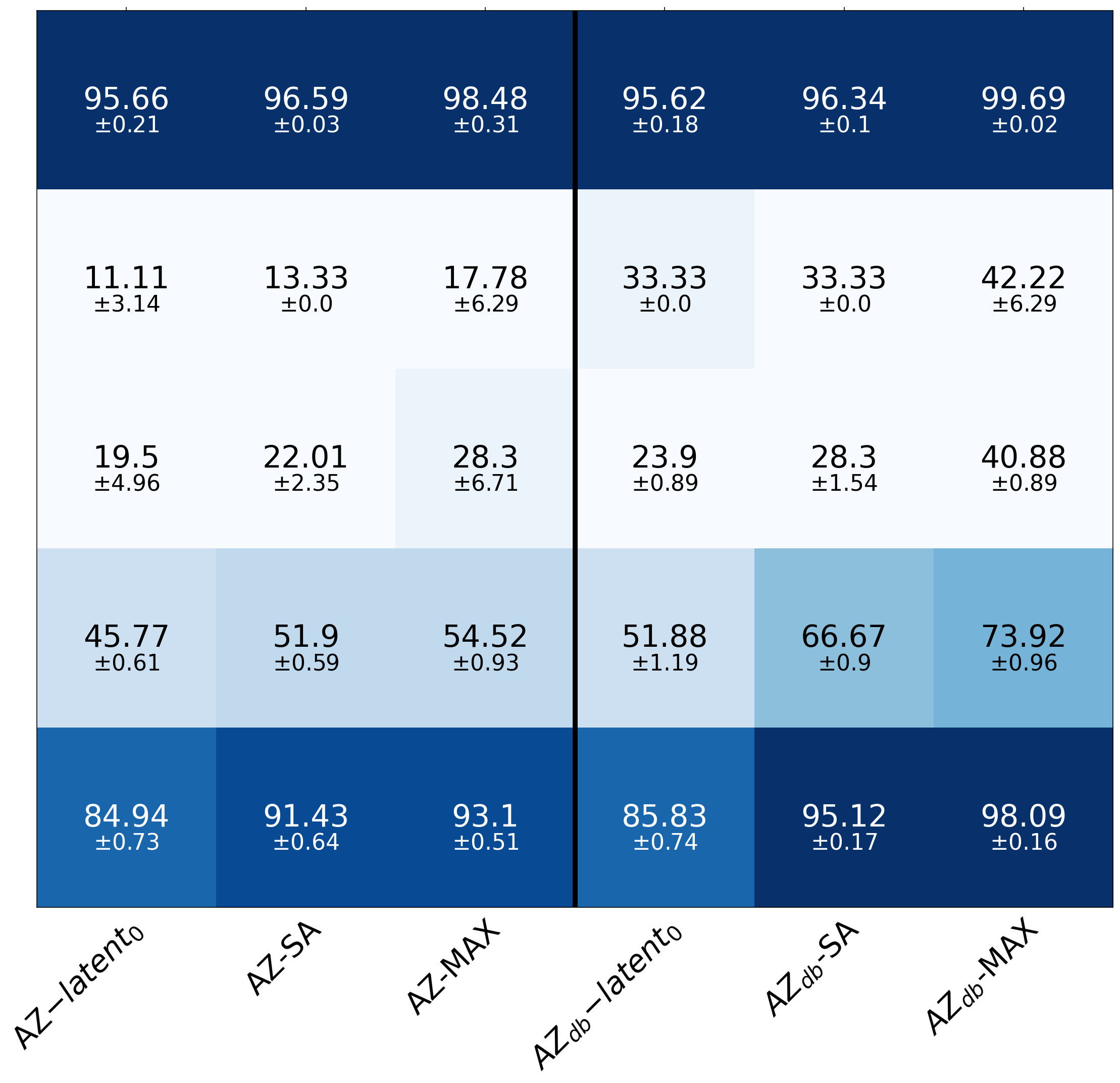}
    \captionof{figure}{The solve rate of AZ and \cz with sub additive planning and max over latents in different puzzle sets. \textbf{Left:} 1 training seed with the full configuration and 100M simulations. Results are averaged over 3 evaluation seeds. \textbf{Center:} 3 training seeds with the fast configuration and 1M simulations. Results are averaged over 3 evaluation seeds and the 3 training seeds. \textbf{Right:} 3 training seeds with the fast configuration and 1M simulations. Results are averaged over 3 evaluation seeds. Sub additive planning and max-over-latents are taken over the 3 training seeds.}
  \label{fig:leaderboard}
\end{figure}

To evaluate the performance of our models, we conducted extensive training and evaluation experiments. Three seeds of both \cz and AZ models were trained using the fast configuration. Each seed was evaluated using $1$ million simulations and three evaluation seeds. Furthermore, we trained AZ and \cz models with the full configuration, which allowed us to search for $100$ million simulations per position. This was the largest search budget we could afford due to the limited memory available due to the size of the search tree. However, in this setup, we could only train one seed but used three evaluation seeds, similar to the previous setup. Unless explicitly stated otherwise, we employed the GAP action selection rule for all of our experiments.

One might assume that this distribution would result in a lower performance for a single player in \cz compared to AZ. Contrary to this assumption, player $0$ in \cz performs better than AZ on most data sets (or comparably to it) as can be observed by inspecting the the performance of AZ on the five puzzle data sets in the left most column of \cref{fig:leaderboard} with player $0$ in \cz (second from left). These results hold with both the full configuration (left figure in \cref{fig:leaderboard}) and with the fast configuration (center figure in \cref{fig:leaderboard}).

While AZ and \cz have the same computational budget during training and play the same number of games, \cz distributes this budget across multiple players. It might be assumed that this distribution would result in lower performance for individual players in \cz compared to AZ. However, contrary to this assumption, player $0$ in \cz outperforms AZ (or performs comparably to it) on most data sets. For instance, inspecting \cref{fig:leaderboard} (left sub figure for the full configuration and middle sub figure for the fast configuration), we observe that the performance of AZ on the five puzzle data sets (leftmost column of each sub figure) is inferior to that of player $0$ in \cz (second column from left). 

We can also observe that by using sub-additive planning \cz can solve significantly more puzzles as a team. For instance, with the fast configuration (center), \cz with sub additive planning (second column from right) solves $0.86\%,$ $2.22\%,$ $2.94\%,$ $11.06\%,$ and $7.34\%,$ more puzzles than player $0$ (second column from left) on STS, Challenge, Penrose, Hard Talkchess and Lichess respectively. With the full configuration (left), we see smaller gains but still a positive improvement ($0.39\%,$ $0\%,$ $-0.61\%,$ $9.16\%,$ and $1.67\%$,  where the onlu decrease in performance is in the Penrose set, but it is within the statistical error bound ($0.89$). 

Additionally, we can observe that with \textit{max-over-latents} (right most column), relatively to sub additive planning (second column from right), than player $0$ (second column from left): with the fast configuration (center), it solves $3.57\%,$ $6.93\%,$  $9.86\%,$ $9.86\%,$ $18.96\%$ more puzzles, and with the full configuration (left) it solves $2.63\%,$ $11.1\%,$  $4.41\%,$ $15.79\%,$ $2.52\%$  more puzzles. The large improvement over player $0$ indicates that as a team, \cz discovers many more solutions to puzzles. However, the fact the \textit{max-over-latents} is much better than sub-additive planing indicates that it is not always easy to tell which player found the correct solution, and that there is room to improvement with sub-additive planning (as a reminder, \textit{max-over-latents} is an upper bound for any sub-additive planning method). Notably, \cz successfully solved two notoriously challenging Penrose positions.

Next, in Figure \ref{fig:leaderboard} (right), we delve deeper into the diversity bonuses of both AZ and \cz across training seeds. We conducted an experiment involving the training of three training seeds with the fast configuration, followed by averaging over three evaluation seeds, as previously done. Notably, we computed the sub-additive planning and \textit{max-over-latents} scores across training seeds, treating each training seed as a distinct player.

In the provided figure, the first three columns display the results for AZ, followed by the results for \cz. For AZ, "latent 0" signifies the performance of the initial training seed, while the sub-additive and max-over latents figures encompass the three training seeds. Similarly, for \cz, "latent $0$" aligns with latent $0$ of the first training seed, and the sub-additive and max-over latents encompass the latents of the three training seeds.

The figure illustrates diversity bonuses across different training seeds in all datasets. The performance of both AZ and \cz models shows significant improvement with sub-additive planning and max over latents compared to their performance with only latent $0$. Notably, the sub-additive planning and max-over-latents scores of \cz are considerably higher than those of AZ, indicating that creativity and diversity bonuses stem from both training-induced diversity and intrinsic behavioral diversity. It is important to note that training \cz involves a computational complexity equivalent to training AZ. In contrast, training $n$ seeds necessitates $n$ times more compute.

Lastly, we note that in addition to the sub-additive planning methods discussed in Equation \cref{eq:sub-additive}, we explored alternative approaches that involve selecting a player to solve the puzzle before engaging in planning. These methods offer the advantage of eliminating the need to perform planning with all players simultaneously. Examples of such methods include player selection based on raw value estimates, the prior probability distribution, the entropy of the prior, and combinations of these statistics. However, the performance of these methods was comparable or inferior to that achieved by player $0$. Similarly, we experimented with super additive planning methods, that switch between the players in a single search, and decide which player should evaluate a given position using the above "none planning" heuristics. However, these methods have not performed well neither.
In addition to the sub-additive planning methods discussed in Equation \cref{eq:sub-additive}, we investigated alternative approaches involving player selection before planning. These methods eliminate the need for simultaneous planning with all players. Examples include player selection based on raw value estimates, the prior probability distribution, the entropy of the prior, and combinations thereof. However, their performance was comparable or inferior to that of player $0$.

Furthermore, we experimented with super-additive planning methods, which switch between players within a single search and use "none planning" heuristics to determine which player should evaluate a given position. However, these methods also under performed.

\paragraph{Diversity bonuses in chess matchups.} In this section, we assess the impact of diversity bonuses on the performance of \cz in chess matches. We matched each player in the \cz team against AZ and let them play chess from 74 opening positions. For each opening, AZ played one game as black and one as white. The opening positions were carefully chosen by GM Matthew Sadler to favor one player by giving them a strong opening advantage (up to 1500 centipawns). This was done in order to produce more decisive matches between agents, with one player winning rather than the match ending in a draw. The games were played in a bullet time control, where each agent used 400 MCTS simulations to select a move in a match. This roughly translates to 100 milliseconds of thinking time per move. Recall that the same amount of MCTS simulations were used in training \cz.

\begin{table}[h]
\caption{Summary of chess matchups between \cz and AZ. The standard error, in winrate, is $0.34$ (for each Player) up to quantization error of $0.01$.}
\centering
    \begin{tabular}{|p{0.1\linewidth} |p{0.15\linewidth} | p{0.3\linewidth} | p{0.22\linewidth}|} 
     \specialrule{.2em}{.1em}{.1em} 
      &Player 0 & \textit{Sub-additive planning}  & \textit{Max-over-latents} \\
    \specialrule{.2em}{.1em}{.1em} 
     Winrate & 54.25 & 57.18 & 58.6 \\
     \specialrule{.2em}{.1em}{.1em} 
     Elo & 29.6 & 50.3 & 60.3 \\
     \specialrule{.2em}{.1em}{.1em} 
    \end{tabular}
\label{table:chess_matchups}
\end{table}

\cref{table:chess_matchups} presents a summary of the games in terms of overall win rate and relative Elo score for player 0, \textit{Max-over-latents} \& \textit{Sub-additive planning}. \cref{fig:matchup_summary} in the Appendix additionally presents the number of wins, draws and losses (as black and as white) for all the players. These statistics are obtained by averaging the results from each position over $500$ random seeds. The win rate and relative Elo improvements show that most of the players in the \cz team are stronger than AZ. This is likely due to implicit diversity bonuses, such as seeing more diverse data during training and playing against each other. In particular, player 0, the strongest player, achieves a win rate of 54.25\% and a relative Elo of 29.6 compared to AZ. It makes sense that player 0 is the strongest in the team, as it is a “special player” that is trained without the diversity intrinsic reward.

We can also see that the \textit{max-over-latents} method performs better than player $0$. But unfortunately, \textit{max-over-latents} is not practical for selecting a player to play in an opening position, since we do not know the outcome of the game in the opening (when we select the player to play). On the other hand, sub-additive planning is practical, and we now discuss how we implemented it. For each opening and seed, we calculated the average score of each player across all other seeds (leaving the current seed out) and selected the player with the highest score. We then reported the score of this player against the current hold-out seed (more details in the Appendix). \cref{table:chess_matchups} shows the results for the Gap sub-additive planning method (\cref{eq:sub-additive}), where the gap is defined to be $2/500$. This value assures that the candidates are all the policies that can become optimal if we add one more game in which they win and the optimal policy loses. As we can see, this improves \cz's win rate to 57.18\%, which translates to a 50.3 ELO improvement over AZ. The results for the other sub-additive rules were $40$ for the value and visit counts, and $41$ for the LCB, \ie, they all improve by around $10$ Elo over player 0, but the gap method was better than all of them.

\begin{figure}[t]
\centering
    \hspace{-5mm}
    {\includegraphics[width=0.8\linewidth]{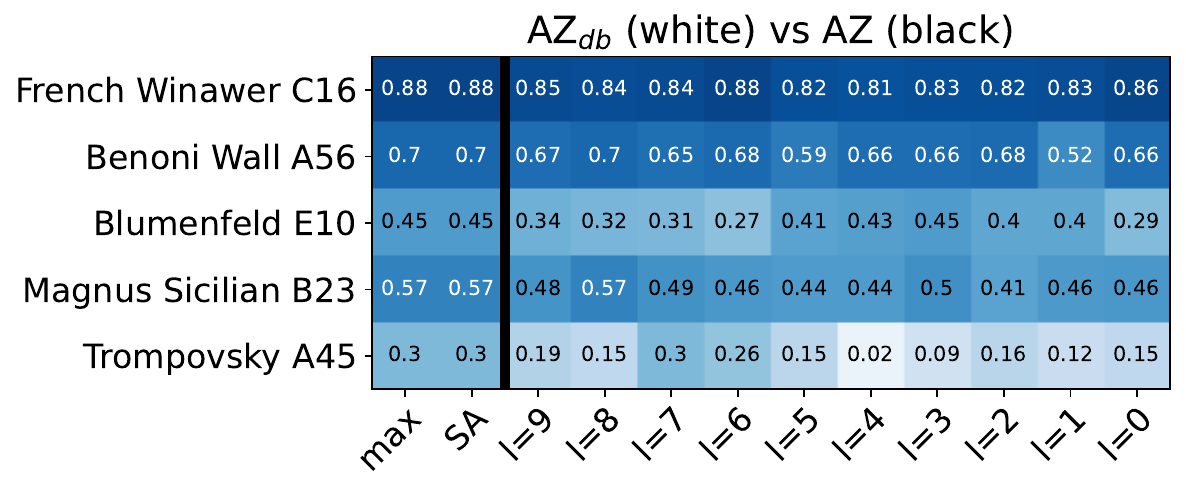}}
    \caption{Results from a subset of chess openings between \cz and AZ (scores averaged from 500 random seeds) where \cz played as White and AZ as Black. Results from all openings are presented in the Appendix.}
    \label{fig:cz_vs_az_400sims_matchups_subset}
\end{figure}

In \cref{fig:cz_vs_az_400sims_matchups_subset} we further inspect these results. We report the average score (a win is scored as $1$, draw is $0$ and loss is $-1$) of the players in \cz when playing as white in a subset of openings. The results on all the opening positions can be found in the Appendix (\cref{fig:cz_vs_az_400sims_matchups}). We can see that the performance of the players in \cz vary across openings. In \textit{French Winawer C16}, all the players perform roughly the same. However, in other openings, there are specific players that perform significantly better than others. For example, player 0 is strong in \textit{Benoni Wall A56}, but is weak in comparison to the other players in \textit{Blumenfeld E10}. While player 0 is the overall best player, they are not the best player in any specific opening. They are also not the best player when playing as white or black. \cref{table:matchup_stats} in the Appendix shows the average win rate of each player when playing as white and black. Player 0 won 68.27\% as white and 40.21\% as black. However, Player 2 had the highest win rate as white (69.86\%), and player 9 as black (40.38\%). 
These results suggest that different players specialize in different openings and that sub-additive planning can be used to select the better player in each opening. Further, \cref{fig:sub_additive_match_ups} in the Appendix shows how sub-additive planning performs as a function of the number of seeds.

\begin{figure}
\centering
    \includegraphics[width=\textwidth]{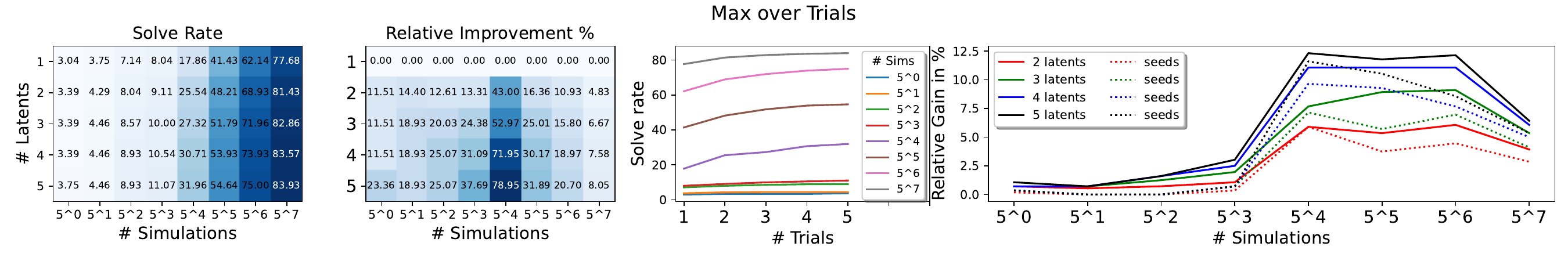}
    \includegraphics[width=\textwidth]{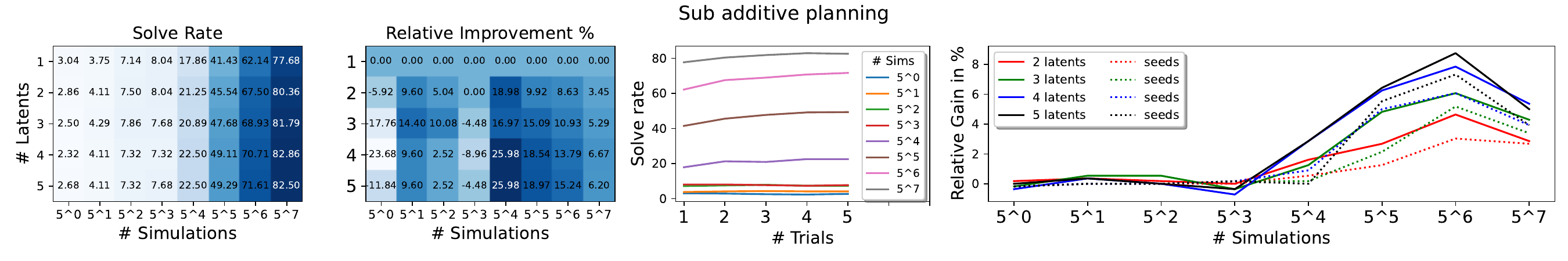}
    \vspace{-0.3cm}
    \caption{Scaling laws with \cz. Top: Max over trials, Bottom: Sub-additive planning. Left to right: (1) Solve rate in $\%$ on Lichess puzzles with a different number of simulations and latents. (2) Relative gains in $\%$ from increasing the number of latents for each simulation budget. (3) scaling with the number of latents for different simulation budgets. (4) Relative gains with a different number of trials --  latent in solid lines, seeds in dashed lines -- at different simulation budgets. }
    \label{fig:scaling}
\end{figure}

\paragraph{Ablative analysis of diversity bonuses in \cz.}
In the puzzle evaluation section, we observed that diversity bonuses emerge at the computational boundaries of \cz. In this section, we analyse what components of \cz are the most important for diversity bonuses and if diversity bonuses emerge at other compute budgets. We focus our evaluation on the Lichess data set and use the fast configuration for that. 
In \cref{fig:scaling} we study how \cz scales with different simulation budgets and team sizes. The top figure presents results for \textit{max-over-latents} and the bottom figure shows results for sub-additive planning based on LCB (\cref{eq:sub-additive}). Most importantly, we observe diversity bonuses for all compute budgets and team sizes. 
In the leftmost table, we present the absolute solve rate in $\%$. We can see that \cz's performance improves monotonically with the number of simulations and the number of trials, implying that larger teams solve more puzzles together. We can also see that on the third sub-figure, which presents the same data in a different manner. For each column in the first table (simulation budget) we draw a line that presents the solve rate in $\%$ as a function of the number of trials (x-axis). We can see that the diversity bonuses keep increasing as we increase  the number of trials for each simulation budget.

The second table from the right shows the relative gains computed in $\%$ \footnote{we divide each row by the first row (player $0$), subtract $1$ and multiply by $100$} of \cz from having more players in the team.  Interestingly, the highest relative gains are achieved when the number of simulations is $5^4=625,$ which is the simulation budget that is closest to the one we use in training (see the discussion section). 
On the rightmost sub-figure, we compare a diverse team of agents with a more homogeneous team. For the diverse team, we use different latents as before (solid lines) and for the homogeneous group, we use the best latent in the group (latent $0$) and allow it different trials of search (with different seeds, in dashed lines). We can see that across different group sizes, simulation budgets, and for both \textit{max-over-latents} and sub-additive planning, a diverse team outperforms the homogeneous one.

\begin{figure}
\centering
    \includegraphics[clip, trim={3cm 0 3cm 0}, width=0.7\linewidth]{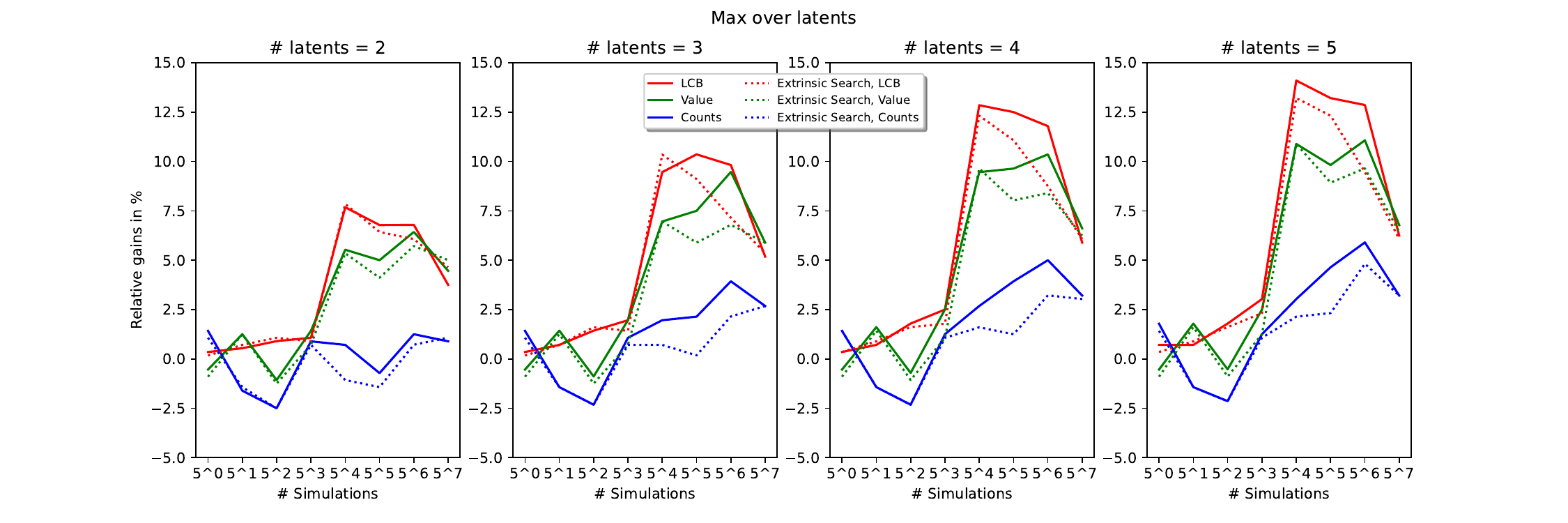}
    \includegraphics[clip, trim={3cm 0 3cm 0}, width=0.7\linewidth]{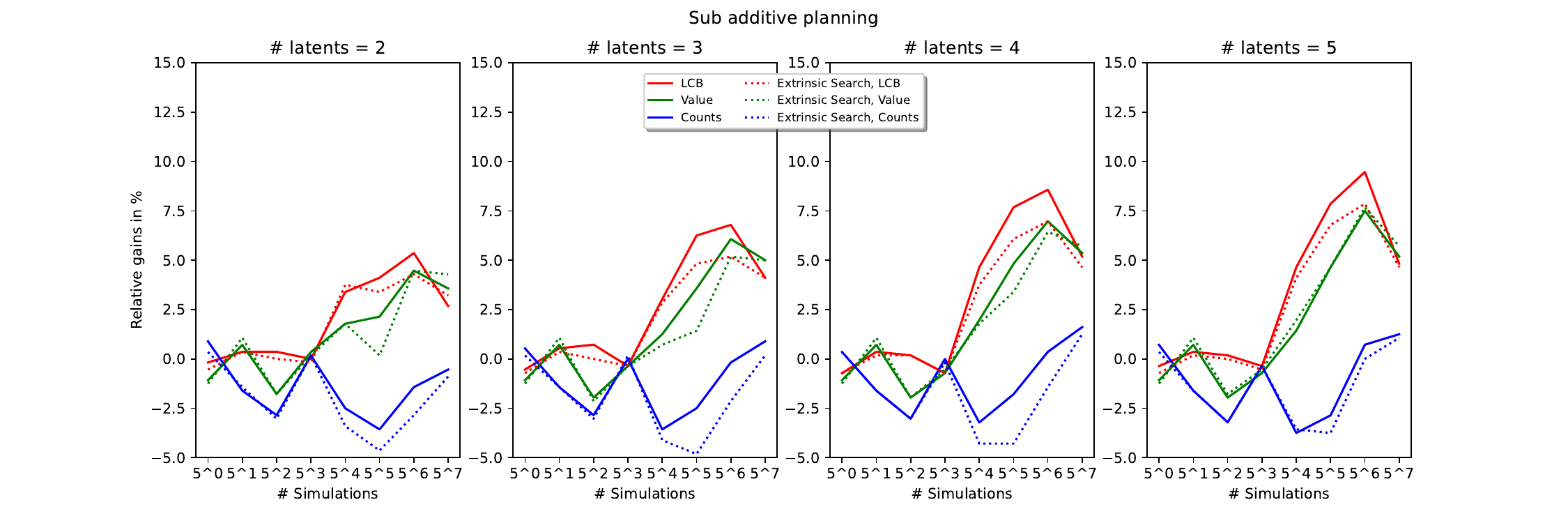}
    \caption{An ablation of action selection rules (color) and search type (solid/dashed) with \textit{max-over-latents} (top) and sub-additve planning (bottom). Sub figures correspond to the number of trials. }
    \label{fig:ablations_max}
\end{figure}

We further study diversity bonuses in a few variations of \cz and compare different sub-additive planning methods (counts, value, and LCB) in \cref{eq:sub-additive}. We also compare \cz with a variation that only uses the extrinsic reward and extrinsic value function at test time (dashed lines). Inspecting \cref{fig:ablations_max},  we can see that using the intrinsic reward and value function (solid lines) receives higher diversity bonuses across different action selection rules and simulation budgets. We also see higher diversity bonuses when selecting actions based on value and LCB rather than counts. 

\textbf{Discussion.} \cz benefited from diversity bonuses regardless of the amount of simulations it was given. However, it gained the most from diversity in solving puzzles when it was given 625 simulations (see \cref{fig:scaling}), and it discovered the most unique openings when it was given $400$ simulations. One possible explanation for these findings is the \emph{spinning top hypothesis} \cite{czarnecki2020real}, which states that the largest diversity of a population of players is observed at a medium level of play, and at the low and high levels of play there is less room for diversity. Indeed, Sanjaya et al.  \cite{sanjaya2022measuring} observed that in chess, human play exhibits the most diversity in the medium levels of play ($1600$ ELO) compared to the higher and lower levels. A different possible explanation is that \cz's diversity overfits to the number of simulations it uses for training ($400$), in the sense that it achieves higher diversity bonuses when it uses the same number of simulations it used for training compared to other simulation budgets. Note that AZ's and \cz's performance improves monotonically with the number of simulations and we do not suggest that the performance overfits to the number of simulations, only that the diversity does. To better understand this explanation, recall that the expected features $\psi$ in our diversity objective are estimated from the average games played by \cz, and since these games are played with $400$ simulations, it is possible that the diversity objective itself overfits to the behavior of \cz's policies with $400$ simulations. 

\section{Related work}
Chess has been a long-standing subject of AI research, and is often called the “Drosophila of AI” \cite{chess_dorosopilia}. The first chess program was written by Alan Turing on slips of paper, before the invention of the computer. Achieving human-level performance in chess was known as the “chess Turing test”, and computer scientists made significant progress towards this goal until Deep Blue finally beat the world chess champion in 1996. In his book “Deep Thinking” \cite{deepthinking} Kasparov tells his side of the story of his loss to Deep Blue. He describes the experience of strategizing against an opponent that is tireless and never makes mistakes, and the mistakes he made that led to his defeat. While some have expressed concerns that chess will become less popular as machines become stronger than humans, chess is now more popular than ever due to an increase in online play during the COVID-19 pandemic, and the success of the Netflix series “The Queen's Gambit” and streamers in bringing attention to the sport (\href{https://www.wuft.org/news/2023/02/17/chess-grows-more-popular-with-chess-com-experiencing-server-issues-due-to-high-user-volume/}{WUFT}).

In 2018, AZ made a significant impact on the world of chess by learning to play chess on its own, without any human knowledge. The games between AZ and SF drew a lot of attention from the chess community, on which Kasparov wrote that “Deep blue was the end, AlphaZero is a beginning”. In fact, almost all modern chess engines today have followed AZ's lead and use a neural network score function instead of a human-designed one. Professional chess players also changed their playing style after the AZ vs SF games were introduced \cite{gonzalez2022alphazero}, prototyped new variants of chess using AZ \cite{tomavsev2020assessing}, and investigated when and where human chess concepts are represented in AZ \cite{mcgrath2022acquisition}. AZ has also been extended beyond games to discover better algorithms to compress videos \cite{mandhane2022muzero}, for matrix multiplication \cite{fawzi2022discovering}, and similar techniques combining search and a neural network have been applied to other domains \cite{kemmerling2023beyond}.

Despite the strength of chess engines, it is uncertain whether we are near optimal play, as evidenced by the results of computer chess competitions like the \href{https://tcec-chess.com/}{TCEC}, where the strength of the top engines continue to increase with each season. Another piece of evidence is that modern chess engines tend to make blunders when constrained to use a limited amount of search. A famous example is the winning move in game eight of the Classical World Chess Championship 2004 between Peter Leko and Vladimir Kramnik (\href{https://en.chessbase.com/post/game-8-leko-wins-to-take-the-lead}{chessbase}, \href{https://www.youtube.com/watch?v=yGnpewUKP88&t=1s}{Agadmator's Chess Channel}): \emph{“Grand Master Vladimir Kramnik relied a bit too much on the engine in his Marshall Gambit preparation. The engine initially told Kramnik he was winning, but Leko let his engine calculate more deeply and realized that his mating attack would triumph over Kramnik's passed pawn”} \cite{10positions}.  While this example is from 2004 and nowadays chess engines can easily solve it, we can find similar cases on “chess twitter” on a regular basis. For example, GM Anish Giri explains that for similar reasons \emph{“top players use cloud engines and not the free stuff available online”} (\href{https://twitter.com/anishgiri/status/1638429747930595331?s=20}{Twitter}). 

In order to beat superhuman computers, chess players developed \href{https://en.wikipedia.org/wiki/Anti-computer_tactics}{anti-computer tactics}, that exploit blind spots in how computers reason about chess. Chess players have been using these tactics against chess programs in their early days, but successful examples became fewer as machines became better. One notable exception is the achievements of GM Hikaru Nakamura against modern chess engines. In 2008, Nakamura played against the top engine at the time, Rybka, and managed to close the position, offer sacrifices, and convince the engine to overpress and eventually lose (\href{https://www.chess.com/blog/SamCopeland/hikaru-nakamura-crushes-computer-in-blitz-chess-top-10-of-the-2000s-rybka-vs-nakamura-2008}{source}); more recently Nakamura used similar tactics against Mittens and secured a draw ( \href{https://youtu.be/EYB7NOSY-hI}{source}). 

Another example of the limitations of computer chess programs is that they cannot solve certain positions \cite{penrose,penrose2,10positions}. To address this, scientists have developed puzzle-specific chess programs like \href{https://www.chessprogramming.org/Crystal}{Crystal} -- a UCI chess engine by Joseph Ellis derived from Stockfish. Crystal focuses on better problem solving abilities, and includes fortress detection code. It further addresses some rarely occurring positional or tactical blindness due to too aggressive reductions and pruning (empirical comparison can be found in the Appendix). Other researchers developed specific methods for specific positions like fortresses \cite{guid2012detecting}. In a related study, GM Hedinn Steingrimsson collected a set of chess fortresses and perturbed them by adding or removing pieces \cite{steingrimsson2021chess}. He then analyzed the performance of different variants of Lc0 on these positions. As can be seen in Figure 3 in the study, after 1 million MCTS simulations, the best performing Lc0 network did not solve all the fortress positions they considered. In many cases, it did not find the correct solution since it did not explore it enough. Further investigations suggested that considering the engine’s top-n moves rather than the top-1 and guiding it through various stages of the positions improved the results. In other cases, it did find the correct solution, but did not understand the position. This is similar to what we report with vanilla AZ in \cref{table:train_on_test_results} (top row) and \cref{fig:puzzles_hard_p1} (AZ). Steingrimsson concluded that Lc0 struggles to generalise to puzzle positions. He showed that Lc0 can be improved by increasing exploration, and hypothesized that training diverse agents is a promising future direction.

In this work, we investigated whether creative problem-solving mechanisms identified in humans can make machines more creative and help them solve such positions. This was inspired by Kasparov's statement that “creativity has a human quality: it accepts the notion of failure” \cite{deepthinking}. We focused on diverse policy discovery, since diverse teams are known to be more creative and robust to failures.

Although diverse policy discovery has not been extensively studied in chess, there are some related mentions. For example, chess engine designers make different choices that can lead to different playing styles. Lc0 and AZ use MCTS, while SF uses alpha-beta pruning. Even the rules of chess have some references to diversity, such as the tension between the bishop and the knight \cite{demis}. Chess opening theory is also diverse, as many opening positions have multiple possible variations. As an example, different AZ seeds prefer different moves in the Ruy Lopez opening \cite{mcgrath2022acquisition}. In the past, chess clubs used to collaborate and play each other via \href{https://www.chess.com/article/view/correspondence-chess---a-histo}{correspondence chess}, which led to creative and impressive games. One notable mention is the “Kasparov versus the World” game \cite{kasparov2000kasparov}, where a team of thousands played against Kasparov and decided on each move by plurality vote. The team demonstrated strong chess on which Kasparov wrote that he had “never expended as much effort on any other game in his life”. In a similar vein, chess grandmasters often prepare for important games or matches with a team, composed of strong players with diverse styles and qualities (e.g., \href{https://www.youtube.com/watch?v=VEjEZuWatUg}{GM Magnus Carlsen's team}). GM Magnus Carlsen has also commented on the importance of finding ideas that are missed by other players in the \href{https://www.youtube.com/watch?v=0ZO28NtkwwQ
}{Lex Fridman Podcast}: \emph{“Now it’s all about finding ideas that are missed by agents, either that they are missed entirely or missed at low depth. These are used for surprising your opponent with lines in which you have more knowledge”}. There is also a significant body of work in chess on identifying and generating the playing style of specific human players \cite{mcilroy2020aligning} called Maia chess.  Maia chess is not optimized for strength, but demonstrates diversity by playing with different human styles and at various Elo levels, which is similar in spirit to the diversity of \cz. Lastly, GM Hedinn Steingrimsson hypothesised that diverse agents with a heterogeneous skill set might be able to solve more chess fortresses \cite{steingrimsson2021chess}.

In AI, the concept of diversity dates back to Turing's cultural search in “Intelligent Machinery” \cite{Turing}: \emph{“it is necessary for a man to be immersed in an environment of other men, whose techniques he absorbs, ..., he may then perhaps do little search on his own to make new discoveries which are passed to other men”}. Turing listed cultural search as one of three types of search in which machines can potentially learn. Since then, diversity had been studied in different sub topics. The \href{https://quality-diversity.github.io/}{Quality-Diversity web page} provides an excellent overview of the work done in the evolutionary learning community \cite{pugh2016quality,cully2017quality,mouret2015illuminating}. In RL, research had been conducted on intrinsic rewards for diversity \cite{gregor2016variational, eysenbach2018diversity}, on repulsive and attractive forces \cite{vassiliades2016scaling,flet2021adversarially,liu2017stein,zahavy2022discovering}. There had also been a significant amount of study on diversity in the multi-agent community \cite{balduzzi2019open,czarnecki2020real,vinyals2019alphastar,liu2022neupl}, and cooperative play \cite{carroll2019utility,NEURIPS2021_797134c3}.
We refer the reader to the Appendix for further discussion about these works. 

Learning diverse policies and combining them with sub additive planning is also strongly related to ensemble learning techniques \cite{kittler1998combining,hansen1990neural,polikar2006ensemble}.
There are three main classes of ensemble learning methods: bagging, stacking, and boosting.
In bagging, a single algorithm is used to train different models on random bootstrap samples of the same training data set. In stacking, the same training data set is used to train different machine learning models and/or algorithms. And in boosting, the machine learning models are trained in iterations, where subsequent models focus on training samples that previous models made mistakes on. The models are then combined with a pre-defined rule, a learned rule, or an adaptive rule depending on the method. 

One of the core ideas behind ensemble learning techniques is, that the members of the ensemble make diverse predictions, as they are trained on different data splits, or use different loss functions, learning algorithms or models, during training. These diversifying techniques rely on having access to an entire fixed training set, but in RL, on the other hand, the data that the agent is trained on is collected by the agent as it interacts with the environment. Therefore, in this work we have focused on methods that collect diverse training data sets using diversity techniques. Ensemble learning has also been studied in RL in the context of overestimation bias  \cite{van2016deep,anschel2017averaged,peer2021ensemble} and  exploration-exploitation \cite{osband2016deep}. However, none of these works have focused on diversity. 

Lastly, sub-additive planning had been investigated in AI as well as in other fields. In psychology, heuristics is the process by which humans use mental short cuts to arrive at decisions. Heuristics are simple strategies that humans, animals, organizations, and even machines use to quickly form judgments, make decisions, and find solutions to complex problems. Heuristics are also considered to be an evidence for bounded rationality in human decision making. In RL, a commonly used heuristic is General Policy Improvement (GPI) \cite{barreto2017successor}, which is similar to sub-additive planning based on value. However, in special situations GPI is sub optimal, and can be improved \cite{zahavy2020planning}. Similarly, our results suggest that the LCB and Gap methods performed better than value.

\section{Summary}
This paper investigated whether AI can benefit from diversity bonuses, inspired by how teams of diverse thinkers outperform homogenous teams in challenging tasks. We tested this hypothesis in the game of chess, the “drosophila of AI”, which poses computational challenges for AI systems due to the large number of possibilities that have to be calculated. We proposed \cz, a team of AlphaZero (AZ) agents that are modeled via a latent-conditioned architecture, and trained its players to play chess differently via behavioral diversity techniques and observed that they choose different pawn structures, castle at different times, and prefer different moves in the opening. 
In a chess tournament against AZ, we observed that different players in \cz specialize in different openings, and, that selecting a player from \cz via sub-additive planning leads to an improvement in Elo points.

In our research, we discovered that diversity bonuses are a consistent phenomenon in solving puzzles under varying computational resource constraints. The \cz algorithm exhibited superior performance in solving puzzles when it employed sub-additive planning and a max-over-latents approach. Moreover, its performance further improved as more diverse players were included. Notably, these performance advantages persisted even when the computational budget of a single search player was pushed to its computational rationality limit \cite{lewis2014computational} of 100M simulations (increasing the number of simulations beyond this threshold resulted in out-of-memory issues).

Some of the hardest chess puzzles we considered were created by Sir Roger Penrose. Penrose argued that human creativity is not entirely algorithmic and cannot be replicated by a sufficiently complex computer \cite{penrose1990emperor,penrose1994shadows}. By understanding why humans can solve puzzles that AI cannot, we may be able to uncover the non-algorithmic aspects of human thinking and gain a better understanding of human consciousness \cite{penrose}. While \cref{fig:puzzles_hard_p1} does show that AZ does not understand these positions, it also suggests that AZ can solve almost any puzzle using a simple computational process: learning from trial and error via RL by starting chess games from puzzle positions and playing against itself. In other words, \cref{fig:puzzles_hard_p1} demonstrates that the Penrose positions are computable in the sense presented in \cite{penrose1994shadows} and implied by the finiteness of chess, but also, that there are indeed gaps between human and machine thinking. Through careful analysis, we suggested that the challenge for AZ is that it does not see similar positions when it learns to play chess by playing against itself, and therefore, struggles to generalize to these unseen, OOD positions. In contrast, each policy in \cz is intrinsically motivated to experience different games, making it less likely that a specific position will be OOD for any of its policies. Indeed, in \cref{fig:leaderboard} we observed that some of the players in the \cz team were able to solve the Penrose positions without training on them, while other players, including AZ, were not. This suggests that incorporating human-like creativity and diversity into AZ can improve its ability to generalize. On the other hand, there were still many puzzle positions that \cz failed to generalize to as a team, suggesting that there is still a gap between human and machine thinking. We hope that our paper will inspire future research of these gaps.

\newpage


\appendix

\section{Additional related work}

\textbf{QD} optimization is a type of evolutionary algorithm that aims at generating large collections of diverse solutions that are all high-performing  \cite{pugh2016quality,cully2017quality}. It comprises two main families of approaches: MAP-Elites \cite{mouret2015illuminating} and novelty search with local competition \cite{lehman2011evolving}, which both distinguish and maintain policies that are different in the behavior space. 
The main difference is how the policy collection is implemented, either as a structured grid or an unstructured collection, respectively. Further references can be found on the \href{https://quality-diversity.github.io/}{QD webpage}. 

In \textbf{RL}, intrinsic rewards have been used for discovering diverse skills. The most common approach is to define diversity in terms of the discriminability of different trajectory-specific quantities and to use these ideas to maximize the Mutual information between states and skills \cite{gregor2016variational, eysenbach2018diversity} or eigenvectors of the graph Laplacian \cite{klissarov2023deep}. Other works implicitly induce diversity to learn policies that maximize the set robustness to the worst-possible reward \cite{kumar2020one,zahavy2021discovering}, and some add diversity as a regularizer when maximizing the extrinsic reward \cite{masood2019diversity}.

Parker-Holder et al. \cite{parker2020effective} measure diversity between behavioral embeddings of policies. The advantage is that the behavioral embeddings can be defined via a differential function directly on the parameters of the policy, which makes the diversity objective differentiable w.r.t to the diversity measure.  Other works use successor features \cite{barreto2017successor} to discover and control policies (see \cite{machado2023temporal} for a recent survey). In this line of work, there is a sequence of rounds, and a new policy is discovered in each round by maximizing a stationary reward signal (the objective for each new policy is a standard RL problem). The rewards can be random vectors, one hot vectors, or an output of an algorithm (for example, an algorithm can be designed to increase the diversity of the set). 

Repulsive and attractive forces have also been studied in related work. Vassiliades et al.  \cite{vassiliades2016scaling} suggested to use Voronoi tessellation to partition the feature space of the MAP-Elite algorithm to regions of equal size and Liu et al. \cite{liu2017stein} proposed a Stein Variational Policy Gradient with repulsive and attractive components. Flet-Berliac et al.  \cite{flet2021adversarially} introduced an adversary that mimics the actor in an actor-critic agent and then added a repulsive force between the agent and the adversary. 

There has also been a significant amount of work in \textbf{multi-agent} learning on discovering diverse policies. A common approach in this literature is to encourage response diversity via matchmaking to increase the space of strategies \cite{balduzzi2019open}, which is particularly useful in non-transitive games. Sanjaya et al. \cite{sanjaya2022measuring} quantified
the non-transitivity in chess through real-world data from human players over one billion chess games from Lichess and FICS. Their findings suggest that the strategy space occupied by real-world chess strategies demonstrates a spinning top geometry \cite{czarnecki2020real}. These findings are correlated with our observation that there is more diversity in \cz play style when it uses a medium number of simulations ($400$) and less when it uses a high or low number of simulations.  

Response diversity techniques are usually combined with population-based training \cite{jaderberg2017population}, which represents a collection of players with a separate set of parameters for each player. One important example is the Alpha League, which led to significantly improved AlphaStar's performance in Starcraft 2 \cite{vinyals2019alphastar}. More recently, response diversity has also been used with a latent conditioned architecture \cite{liu2022neupl,pmlr-v162-liu22h}, closer to our setup. 

Another important use case for diversity in multi-agent learning is to first discover diverse high-quality policies and then train the best response policy that exploits all of the discovered policies \cite{pmlr-v139-lupu21a}. In another example, Liu et al. \cite{liu2017stein} considered diversity in Markov games and proposed a state-occupancy-based diversity objective via the shared state-action occupancy of all the players in the game.

\section{Response diversity}
\label{sec:matchmaking}
\subsection{Technical details}

We encourage response diversity in a league of players $\setpolicies$ by training on matchups sampled from a probability function $\Prob(i, j, t^i)$. Intuitively, the matchmaking works by first uniformly sampling player $i$ from $\Pi$. It then uniformly samples an indicator $t^i$ to denote whether player $i$ is playing as black or white, and the oponent player $j$ is sampled conditionally as:
\begin{equation}
    \label{eq:matchmaking}
    \Prob(i, j, t^i) \eqdef \Prob(j | i, t^i) \Prob_\text{unif}(i) \Prob_\text{unif}(t^i)
\end{equation}
Note that chess is not a symmetric game. There is \emph{"a general consensus among players and theorists that the player who makes the first move (white) has an inherent advantage"} (\href{https://en.wikipedia.org/wiki/First-move_advantage_in_chess}{Wikipedia}). Hence we condition the sampling of $j$ on both $i$ and $t^i$.

\begin{figure}[t]
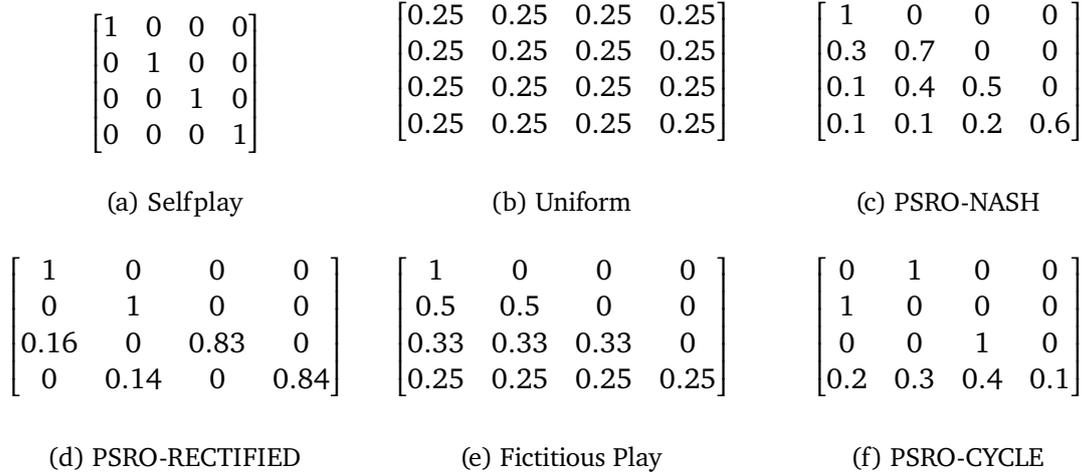

    \centering
    \captionsetup[subfigure]{justification=centering}
    \begin{subfigure}[b]{0.3\textwidth}
    \[
        \begin{bmatrix}
            1 & 0 & 0 & 0 \\
            0 & 1 & 0 & 0 \\
            0 & 0 & 1 & 0 \\
            0 & 0 & 0 & 1
        \end{bmatrix}
    \]
        \caption{Selfplay}
    \end{subfigure}
    \begin{subfigure}[b]{0.3\textwidth}
    \[
        \begin{bmatrix}
            0.25 & 0.25 & 0.25 & 0.25 \\
            0.25 & 0.25 & 0.25 & 0.25 \\
            0.25 & 0.25 & 0.25 & 0.25 \\
            0.25 & 0.25 & 0.25 & 0.25
        \end{bmatrix}
    \]
        \caption{Uniform}
    \end{subfigure}
    \begin{subfigure}[b]{0.3\textwidth}
    \[
        \begin{bmatrix}
            1 & 0 & 0 & 0 \\
            0.3 & 0.7 & 0 & 0 \\
            0.1 & 0.4 & 0.5 & 0 \\
            0.1 & 0.1 & 0.2 & 0.6 \\
        \end{bmatrix}
    \]
        \caption{PSRO-NASH}
    \end{subfigure}
    \begin{subfigure}[b]{0.3\textwidth}
    \[
        \begin{bmatrix}
            1 & 0 & 0 & 0 \\
            0 & 1 & 0 & 0 \\
            0.16 & 0 & 0.83 & 0 \\
            0 & 0.14 & 0 & 0.84 \\
        \end{bmatrix}
    \]
        \caption{PSRO-RECTIFIED}
    \end{subfigure}
    \begin{subfigure}[b]{0.3\textwidth}
    \[
        \begin{bmatrix}
            1 & 0 & 0 & 0 \\
            0.5 & 0.5 & 0 & 0 \\
            0.33 & 0.33 & 0.33 & 0 \\
            0.25 & 0.25 & 0.25 & 0.25 \\
        \end{bmatrix}
    \]
        \caption{Fictitious Play}
    \end{subfigure}
    \begin{subfigure}[b]{0.3\textwidth}
    \[
        \begin{bmatrix}
            0 & 1 & 0 & 0 \\
            1 & 0 & 0 & 0 \\
            0 & 0 & 1 & 0 \\
            0.2 & 0.3 & 0.4 & 0.1 \\
        \end{bmatrix}
    \]
        \caption{PSRO-CYCLE}
    \end{subfigure}
    \caption{Matchmaking Algorithms: (a) Selfplay: Each player only plays against itself. (b) Uniform: Each player is sampled uniformly at random. (c) PSRO-NASH: Player $j$ is sampled from a Nash mixture policy over the current population (up to $i$) that is gradually updated over the course of training. (d) PSRO-RECTIFIED: Similar to PSRO-NASH, except player $j$ is sampled from a ``rectified'' Nash mixture policy. (e) PSRO-UNIFORM: Player $j$ is sampled from the uniform mixture over the current population of players. (f) PSRO-CYCLE: The first $n-1$ players form a strategic cycle and the last player responds to Nash equilibrium over the entire league. }
    \label{fig:interaction_graphs}
\end{figure}

\begin{figure}[h]
    \centering
    \captionsetup[subfigure]{justification=centering}
    \begin{subfigure}[b]{0.2\textwidth}
    \includegraphics[width=1.0\linewidth]{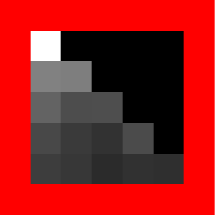}
        \caption{Player One (White)}
    \end{subfigure}
    \begin{subfigure}[b]{0.2\textwidth}
    \includegraphics[width=1.0\linewidth]{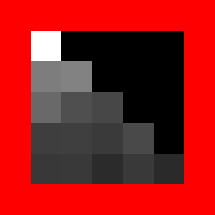}
        \caption{Player Two (Black)}
    \end{subfigure}
    \caption{Empirical PSRO-NASH Interaction Graphs for a league with $5$ players. Each grid represents a matchup between exploiter player i (row) and exploitee player j (column), lighter color implies higher probability (white is probability 1 and black is 0).  Since we only match an opponent $j$ with player $i$ for $j<=i$, the interaction graph is lower diagonal (but includes the diagonal and therefore allows self-play). We can see that most players get to play self-play games (have probability mass on the diagonal) but also get to play against other players.  
}
    \label{fig:real_psro_interaction_graphs}
\end{figure}

Several population learning frameworks can be expressed this way. In practice, $\Prob(j | i, t^i)$ is represented as the adjacency matrix of an interaction graph $\Sigma \in \mathbb{R}^{N \times N} \eqdef \{\sigma^i\}_{i=1}^{N}$, where each row $\sigma^i$ corresponds to a mixture policy over $\setpolicies$, and the $j^{\mathrm{th}}$ element is the probability of $i$ playing against $j$. For example, in Policy Space Response Oracle \citep[PSRO;][]{lanctot2017unified}, each agent is computed as the approximate best response to the Nash equilibrium over players in the current population for some fixed iteration order. Specifically, we consider the following matchmaking algorithms, also illustrated in  \cref{fig:interaction_graphs}:
\begin{itemize}
    \item \textbf{Selfplay} each player only plays against itself, \ie, $j = i$
    \item \textbf{Uniform} player $j$ is sampled uniformly from $\setpolicies$, \ie, $\Prob(j | i, t^i) = \Prob_\text{unif}(j)$
    \item \textbf{PSRO-NASH} player $j$ is sampled from a mixture policy $\sigma^{i}_{b}$ or a mixture policy $\sigma^{i}_{w}$, which are the Nash equilibria over $\{\pi_{k}\}_{k=0}^{i}$ playing as (b)lack and (w)hite respectively. For a real example of interaction graphs for a PSRO-NASH league, see \cref{fig:real_psro_interaction_graphs}.
    \item \textbf{PSRO-RECTIFIED} is the same as \textbf{PSRO-NASH}, except we respond to the \emph{``rectified''} Nash as per \citep{balduzzi2019open}, \ie, each player is trained against the Nash weighted mixture of players that it beats or ties, in order to discover strategic niches and enlarge the empirical gamescape.
    \item \textbf{Fictitious Play} or PSRO-UNIFORM or \citep{brown1951iterative, leslie2006generalised}: player $i$ responds to the uniform mixture over the current population of players. This has been shown to perform well empirically.
    \item \textbf{PSRO-CYCLE} the first $n-1$ players forms a strategic cycle and the last player responds to the Nash equilibrium over the entire league.
\end{itemize}

In contrast to the typical approach where a policy population is grown by iteratively training new policies against snapshots of existing policies, we train all players in the \cz league concurrently.

For PSRO variants, we compute separate interaction graphs $\Sigma_{b}$ and $\Sigma_{w}$ for playing as black and white respectively. During training, we track the outcome of matchups in a payoff matrix and compute $\Sigma_{w}$ and $\Sigma_{b}$ for the row (white) and column (black) players respectively. We also differ slightly from the standard PSRO setup and include the current player in the population it is responding to. This means player 0 is not treated as a fixed policy and essentially engages in standard AZ self-play. This works well empirically.

\textbf{Practical Considerations.}
For PSRO variants where each player is potentially responding to a different mixture policy or matched against specific opponents to discover strategic niches, it is necessary to filter the experience generated from each matchup so that only $i$'s experience (the exploiter) is used to update the agent, and not $j$'s experience (the exploitee). This is because updating from both players' experiences will effectively cancel out the diversity effect we are seeking when players are trained concurrently.

\textbf{Information Hiding.} For standard AZ self-play, MCTS uses the same agent to play both sides of the game during search. This is no longer true for \cz as each player in the league is conditioned on a latent variable $l$, and care needs to be taken to restrict the searching player from using privileged information about their opponent's identity. Therefore during training, we ensure that each player can only plan with their own policy and values.

\clearpage

\subsection{Matchmaker comparisons}

\begin{figure}[h]
\centering
    \includegraphics[width=0.9\linewidth]{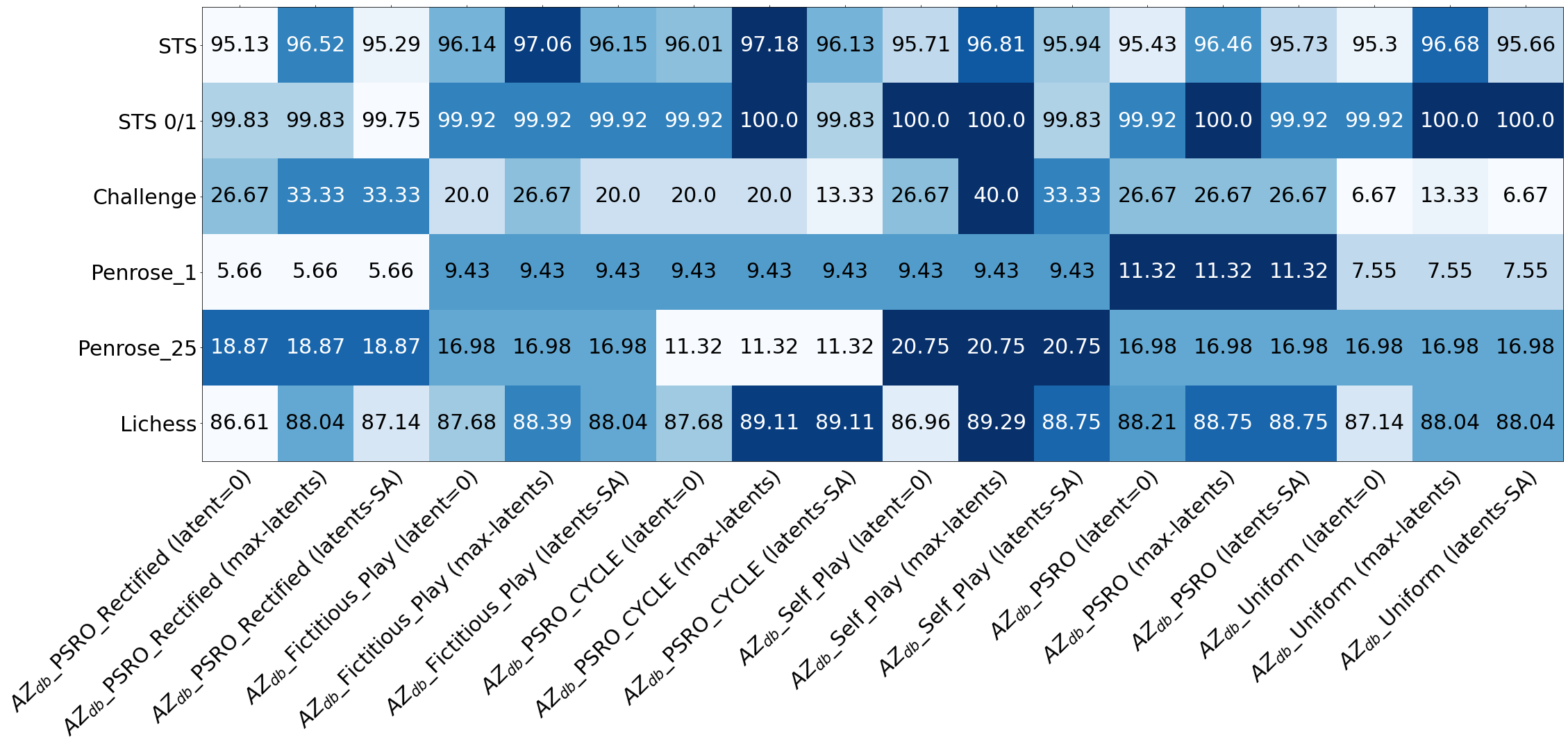}
    \caption{No diversity, without privileged information. }
    \label{fig:league1}
\end{figure}

\begin{figure}[h]
\centering
    \includegraphics[width=0.9\linewidth]{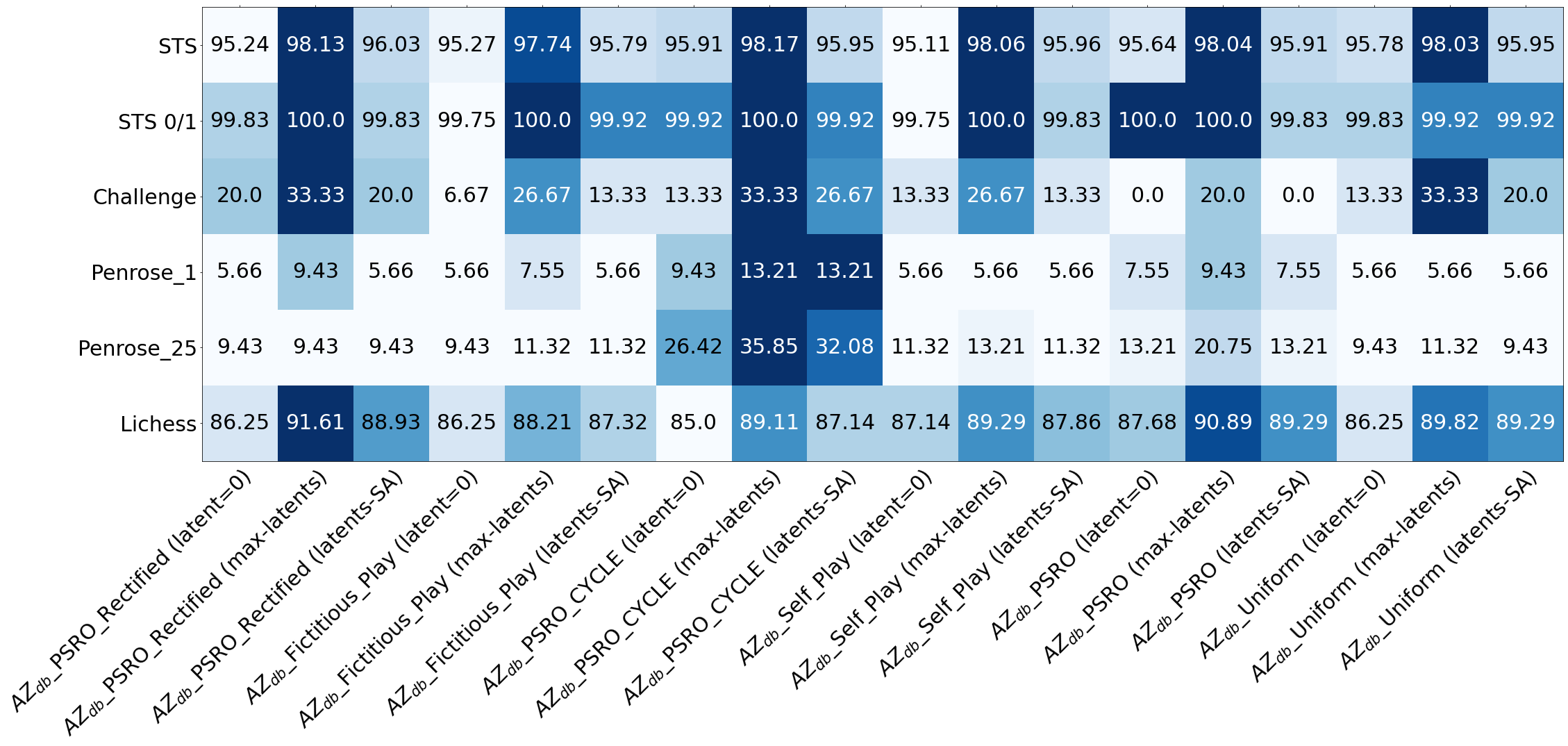}
    \caption{With diversity, without privileged information.}
    \label{fig:league2}
\end{figure}

\begin{figure}[h]
\centering
    \includegraphics[width=0.9\linewidth]{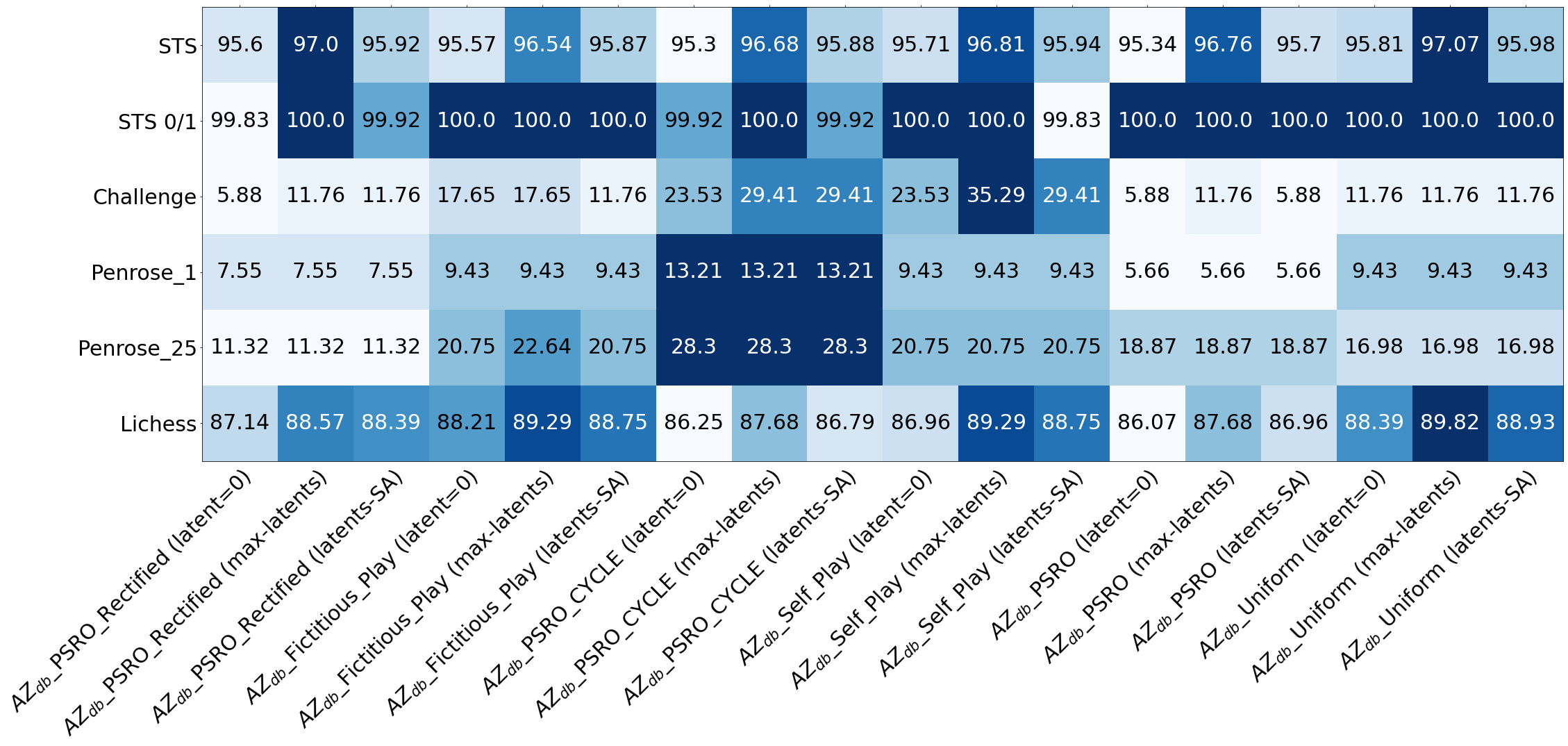}
    \caption{No diversity, with privileged information.}
    \label{fig:league3}
\end{figure}

\begin{figure}[h]
\centering
    \includegraphics[width=0.9\linewidth]{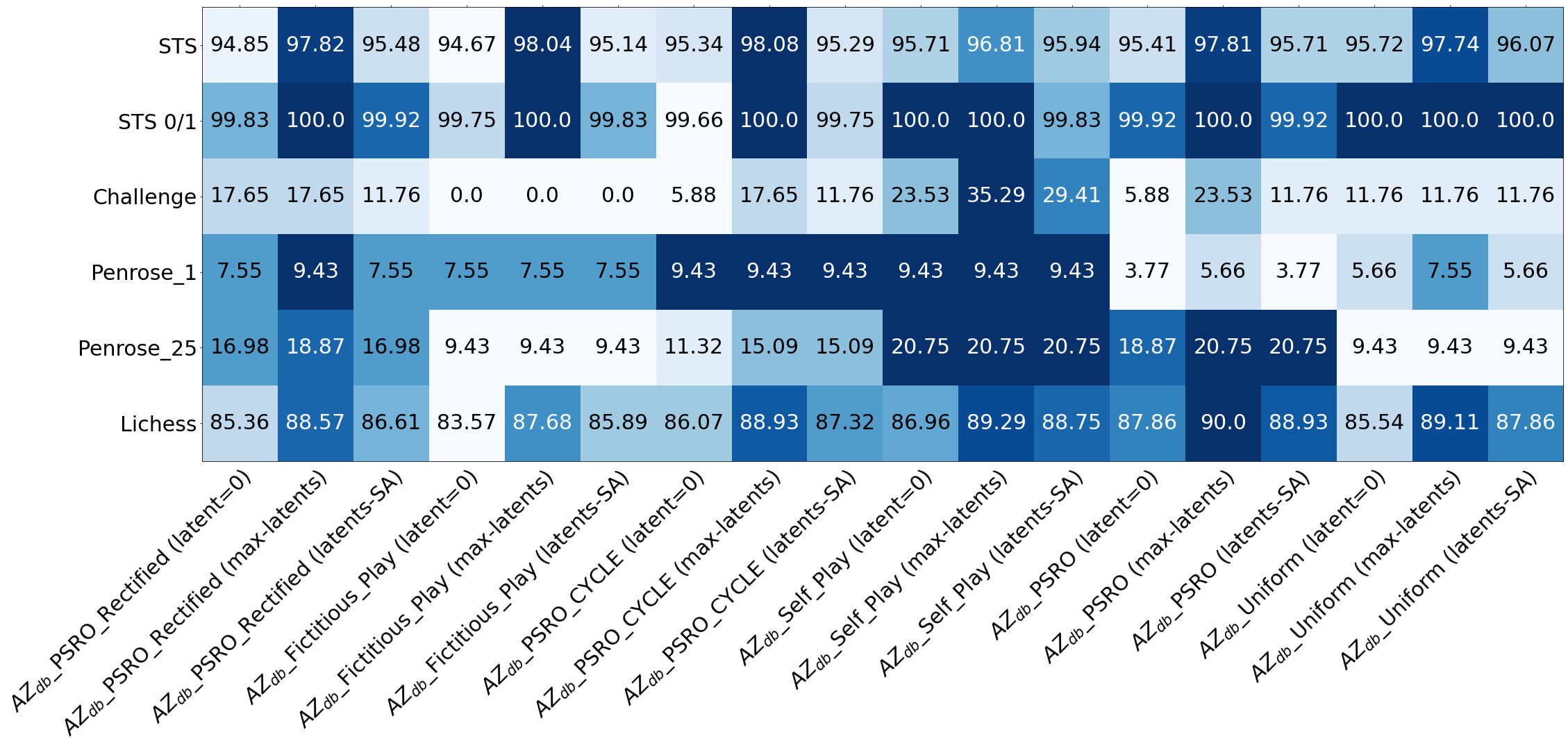}
    \caption{With diversity, with privileged information.}
    \label{fig:league4}
\end{figure}

We now compare the six matchmaking mechanisms proposed above in four experiments (\cref{fig:league1}-\ref{fig:league4}). In each experiment, we trained six agents using the fast configuration. These experiments differ in two aspects: we trained the agents with and without the diversity intrinsic reward (when there is no intrinsic reward, there is only response diversity), and with and without "information hiding", \ie, hiding the privileged information about the opponent during planning. The columns correspond to different matchmakers, grouped in triplets, where for each matchmaker we present the results for latent $0$, max-over-latents and sub additive planning (left-to-right).

The rows correspond to the data sets in the main paper, where we additionally include a version of STS with a $0/1$ loss (the agent gets a score of $1$ if it finds any of the propose solutions, regardless of their internal scores). In the Penrose set, we include two thresholds: $0.25$ as in the main paper and $0.1$. 

Inspecting the results, we could not identify a significant advantage to any of the match making mechanisms. The different matchmakers perform roughly the same, in particular, on the larger and more statistically significant data sets (STS, Lichess). Furthermore, different matchmakers were better than others in different configurations. We could not make an informed decision regarding which matchmaker to use given these results. Nevertheless, we believe that presenting these results here may be of benefit for other researchers interested in match making ideas, in particular in chess and other transitive games.

\clearpage

\clearpage

\clearpage

\section{Diversity bonuses in AZ}
\label{sec:db}
\begin{figure}[h]
\centering
    \includegraphics[width=0.4\linewidth]{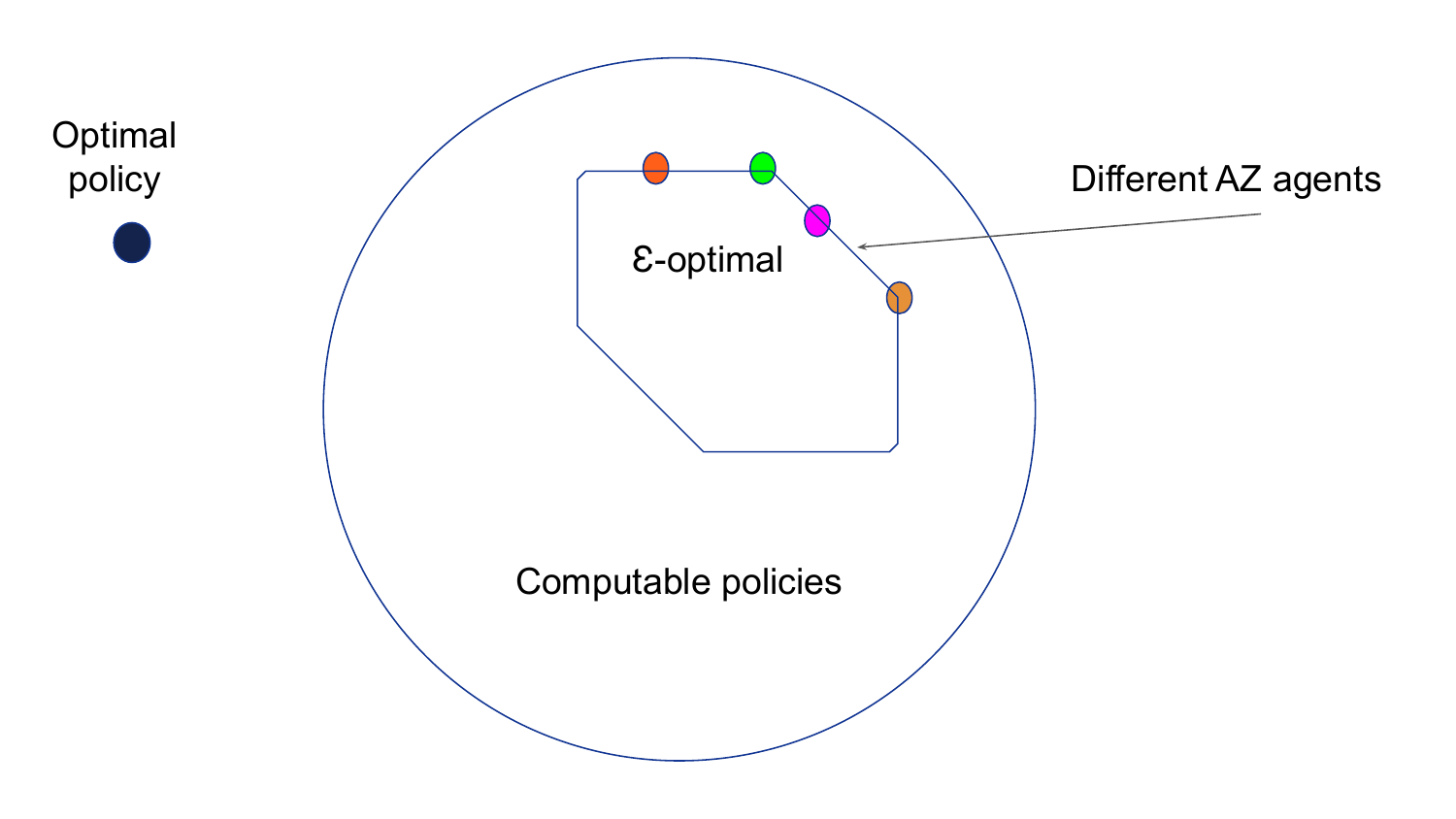}
    \includegraphics[clip, trim=0 0 0 2.5cm,width=0.45\linewidth]{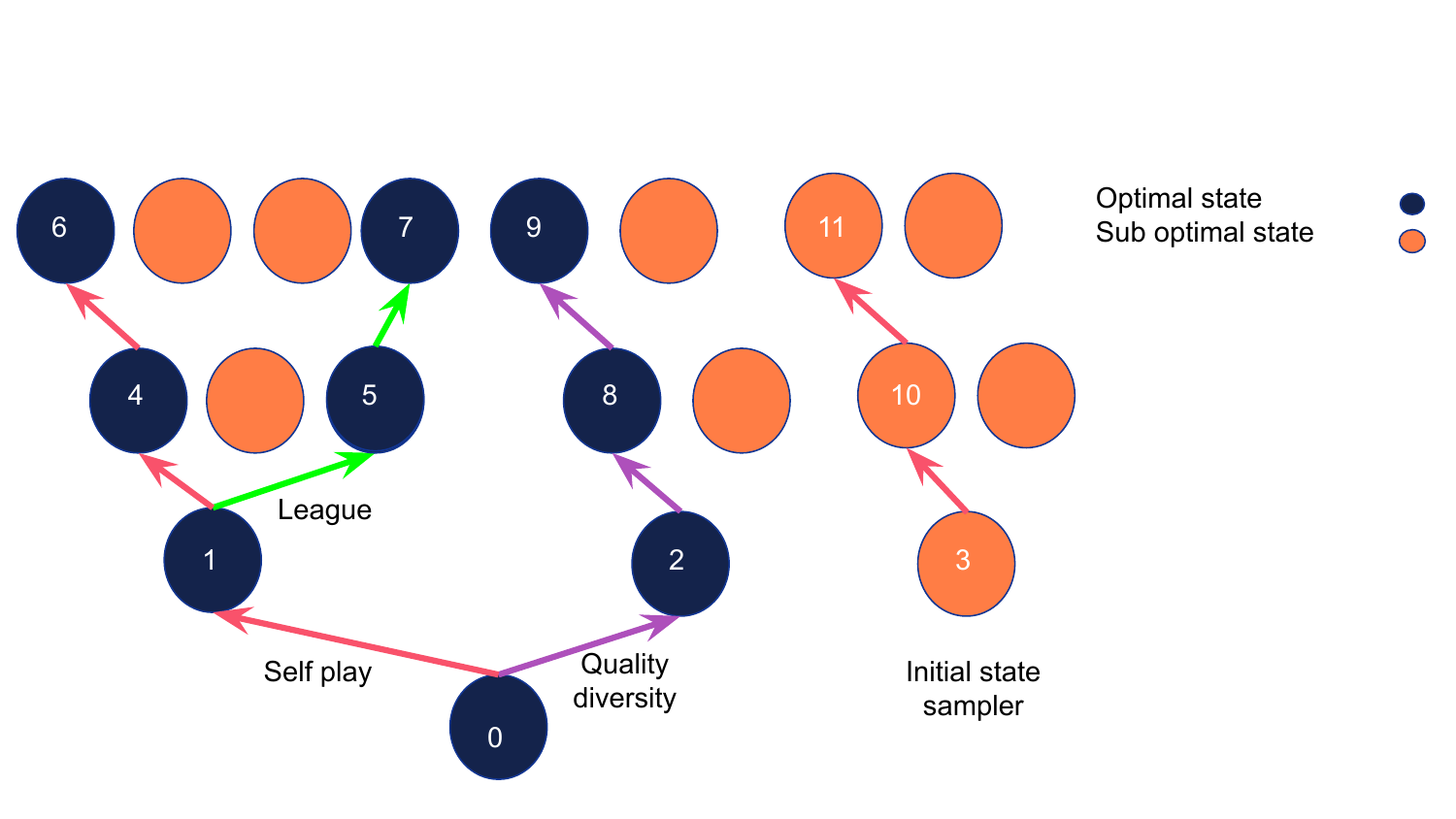}
    \caption{Left: an illustration of the space of chess agents. The optimal policy is not computeable; within the computeable policies there are different $\epsilon$-optimal policies. Right: An illustration of a state space. Blue colors represent optimal states, \ie, states that can be reached with an optimal policy. Red arrows (starting from state $0$) represent one possible optimal trajectory, whereas green and purple arrows represent alternative optimal trajectories. State $3$ is a sub-optimal state that would not be visited by an optimal policy. }
    \label{fig:traj}
\end{figure}

It might surprise some readers that RL can benefit from creative-problem solving mechanisms. In principle, RL can solve any problem via trial and error. However, RL methods cannot find truly optimal policies in most problems of practical interest and hence find a particular (hopefully) near-optimal solution whose choice is determined by the nature of the agent's bounds (memory, compute, training experience), as depicted in \cref{fig:traj} (left). This singular sub-optimal choice can potentially generalise poorly when faced with situations it did not encounter during its training. In other words, a single policy will face Out of Distribution (OOD) situations w.r.t to its self-play distribution (the games that the agent plays against itself and gets to train on). Incorporating multiple policies with different self-play distributions (diverse $d_\pi$) extends the self-play distribution of \cz as a team, and we can therefore expect that some players will generalize better than others in some situations.

Another potential issue is that RL is designed to find an optimal policy from an initial state distribution. This is fine if one only cares about playing chess from the initial board, but people are also interested in knowing what the optimal move is from a variety of positions, potentially outside of the path of the optimal policy. Even with advanced exploration techniques, this fundamental problem remains: exploration is designed to explore the state space until the agent is confident enough about the optimal policy, and at this point, it becomes less likely for it to encounter sub-optimal states. Another issue is that the policy is often deterministic, and as such, will play the same move from each position, ignoring other potentially optimal moves. 
Modern chess engines, however, are required to do much more than play perfect chess from the initial board: they are being asked to recommend moves from sub-optimal positions, analyze games, and more. 

In \cref{fig:traj} (right), we depict one hypothetical optimal trajectory (assuming access to an optimal policy) starting from the initial board (state $0$) and continuing to states $1,4$ and $6$ with red arrows (self-play). 
Having a diverse team of agents can help in alleviating this issue and improving our data distribution in the following manner. In green color, we depict a different optimal trajectory in which the opponent selects a different optimal action in state $1$ leading the game to states $5$ $\&$ $7$. The agent might miss such a trajectory when training with a self-play league and using other matchmakers can help to explore it. Alternatively, an agent may choose to play a different optimal action in state $0$ leading it to visit $2,8$ $\&$ $9$ by optimizing quality diversity (purple). 



\clearpage
\subsection{Evaluation of external chess engines in puzzle}

The goal of this work is to study diversity bonuses in AZ; these techniques are not specific to chess and can be applied to other domains. Thus, our focus has been on measuring the relative gains from diversity and not on comparing or making any claims about the strength of a particular chess engine compared to others.  

Nevertheless we believe it is informative to present the solve rate of reference chess engines, as it provides further evidence of the challenges in solving these puzzles. We evaluated Stockfish 8 \& 15, and Crystal 4.1 with the following configurations:
Level = 20, Number of threads = 16, Hash table size = 1024 MB, Search time / move = 20 min (long time control) and 2 min (short time control). The search time of the engines is equivalent to the search time of AZ in the respective figures.

\begin{figure}[t!]
\centering
    \includegraphics[width=\linewidth]{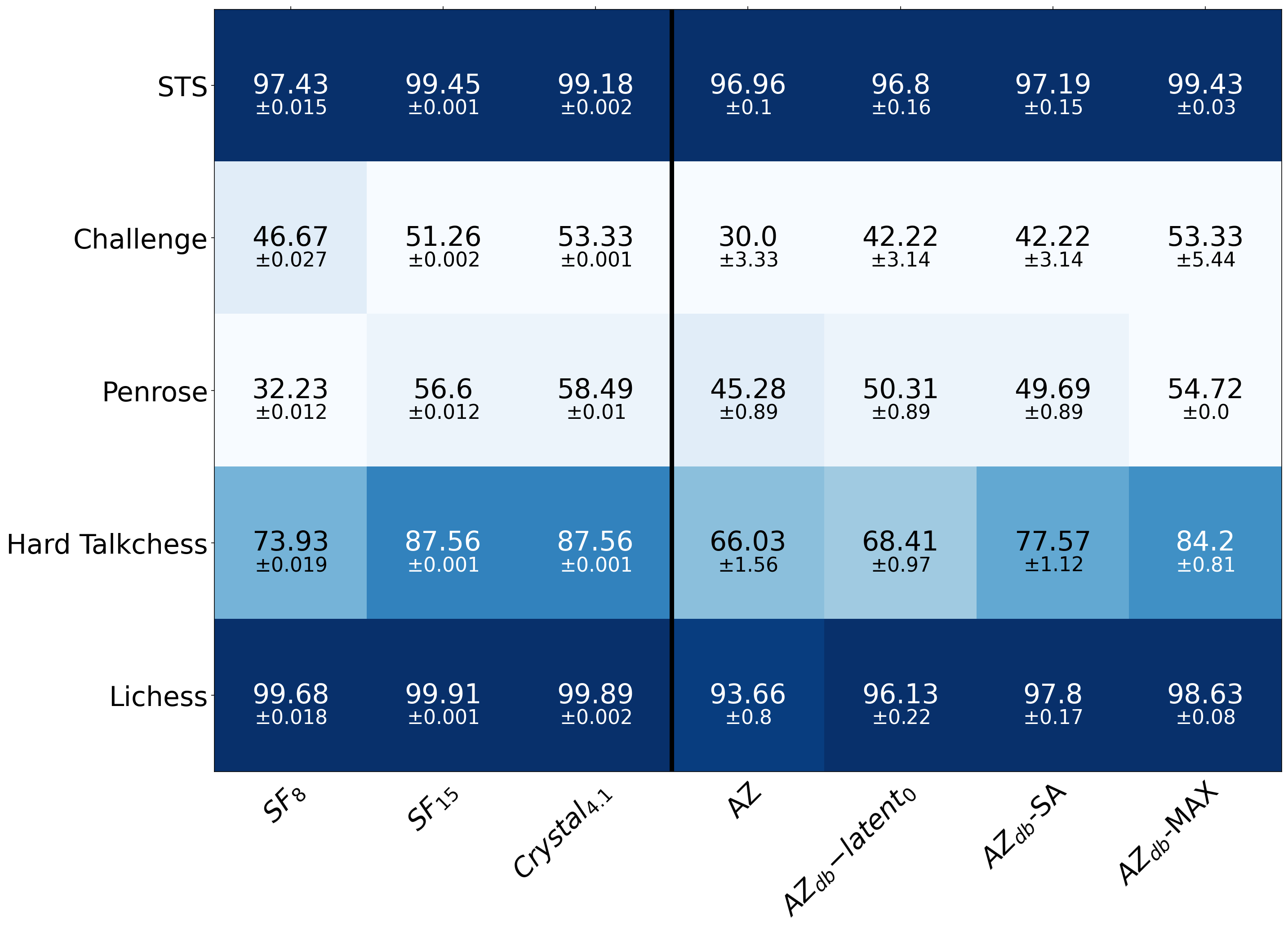}
    \caption{The solve rate of Stockfish 8, 15 \& Crystal 4.1 on different puzzle sets with a search-time of 20 min / move, along with those of AZ and \cz with sub-additive planning and max over latents. AZ and \cz were trained with the full configuration and used 100M simulations. Results are averaged over 3 evaluation seeds. This figure extends \cref{fig:leaderboard} (Left) by including results from chess engines.}
    \label{fig:leaderboard_engines}
\end{figure}

\begin{figure}[t!]
\centering
    \includegraphics[width=\linewidth]{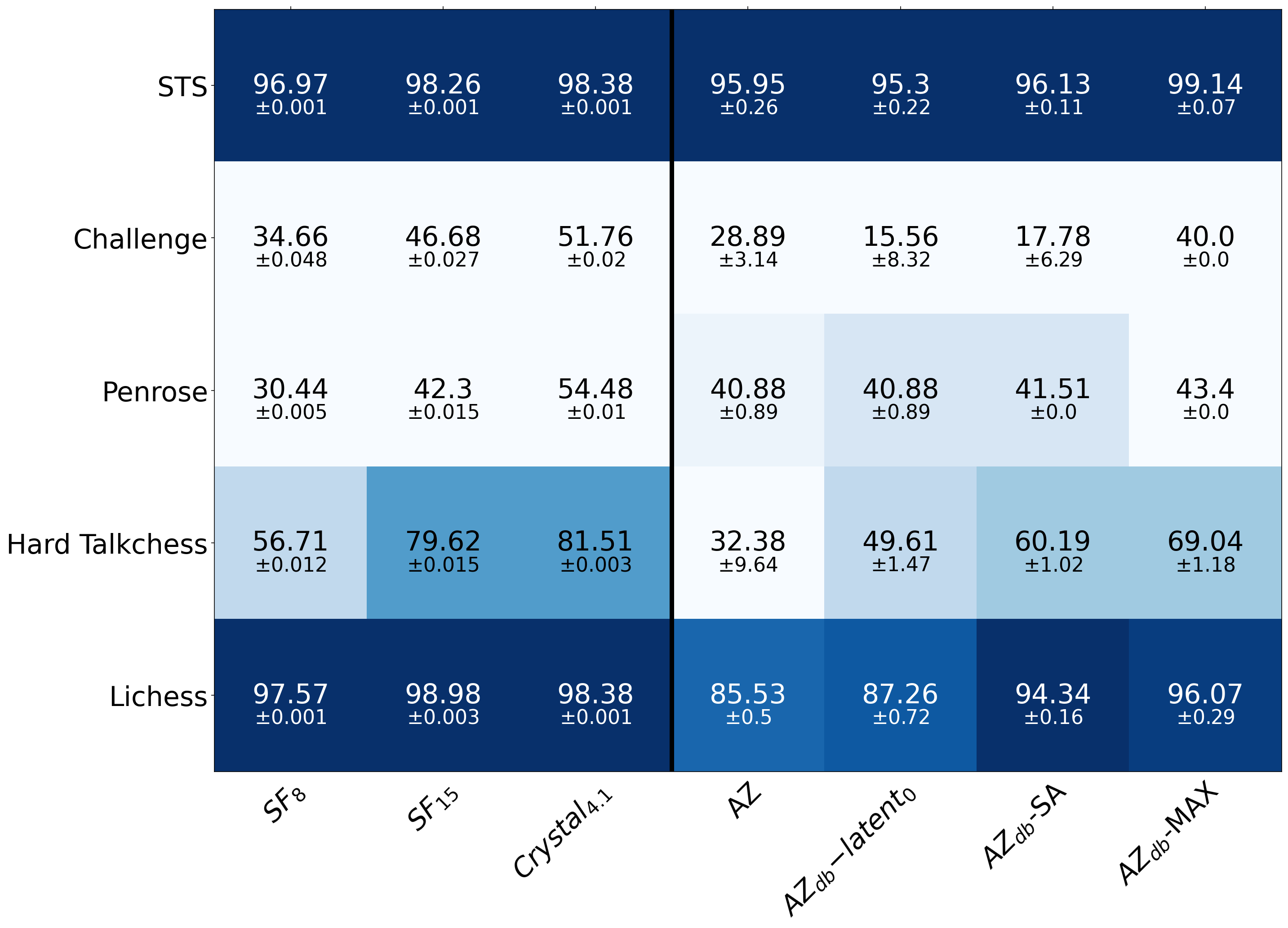}
    \caption{The solve rate of Stockfish 8, 15 \& Crystal 4.1 on different puzzle sets with a search-time of 2 min / move, along with those of AZ and \cz with sub-additive planning and max over latents. AZ and \cz were trained with the full configuration and used 1M simulations. Results are averaged over 3 evaluation seeds.}
    \label{fig:short_time_control_leaderboard_engines}
\end{figure}

The solve rates for long and short time controls across different puzzle  sets are reported in \cref{fig:leaderboard_engines} and \ref{fig:short_time_control_leaderboard_engines}.

\textit{Long time control puzzle evaluation (\cref{fig:leaderboard_engines}):} Stockfish 15 and Crystal solves more than 99\% of the puzzles in STS and Lichess sets. This is perhaps not surprising, as the the solutions for those puzzle sets were verified with Stockfish. Stockfish 8 is competitive with the other two engines on Lichess puzzle set but is a little worse in STS puzzle set. In the Challenge, Penrose and Hard Talkchess puzzle sets, the engines are not as strong as they were in STS and Lichess puzzle sets showing that these are indeed challenging chess positions. Crystal is the stronger among the three chess engines in Challenge and Penrose puzzle sets. Stockfish 15 and Crystal are jointly the best engines in the Hard Talkchess set. Stockfish 8 is significantly worse than the other two engines in Challenge, Penrose and Hard Talkchess sets. Also it is interesting to see that in these three puzzle sets \cz is stronger than Stockfish 8. 

\textit{Short time control puzzle evaluation (\cref{fig:short_time_control_leaderboard_engines}):} With short time controls, the three chess engines have lower solve rate across all puzzles in comparison to long time controls. The drop in performance is substantial (between 5-10\%) in Challenge, Penrose and Hard Talkchess puzzle sets while in STS and Lichess the drop in solve rate is only 1-2\%. This again shows that these puzzle sets have challenging chess positions and requires more time to identify the correct solution. 

\clearpage
\subsection{Visualizing the piece occupancies}


\begin{figure}[h]
\centering
    \includegraphics[width=\linewidth]{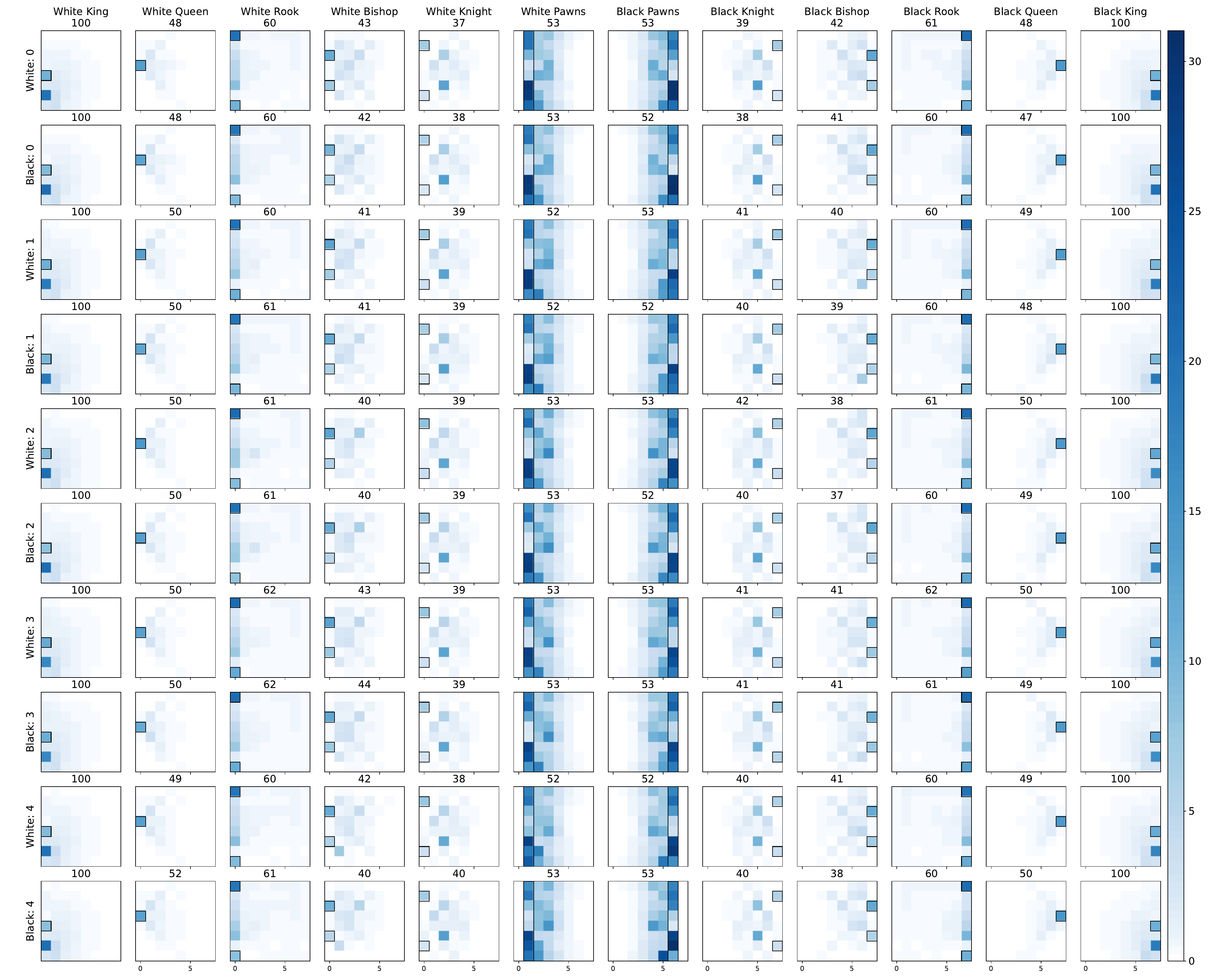}
    \caption{Piece occupancies of \cz. Columns correspond to pieces: leftmost five as white and rightmost as black. Rows correspond to players playing as white (odd rows) and black (even rows). Numbers above sub figures indicate the sum of the occupancy over the board divided by the number of pieces (eg 1 for king, 8 for pawns) and indicate how long does a piece stays on the board on average. Black rectangles highlight the starting position of each piece. Color units (in the colorbars) correspond to occupancy in $\%$ multiplied by the average number of moves in a game (65). For orientation, the chess coordinate system is displayed in Figure 3 in the main paper, top-left. }
    \label{fig:sfs}
\end{figure}

\clearpage
\begin{figure}[h]
\centering
    \includegraphics[width=\linewidth]{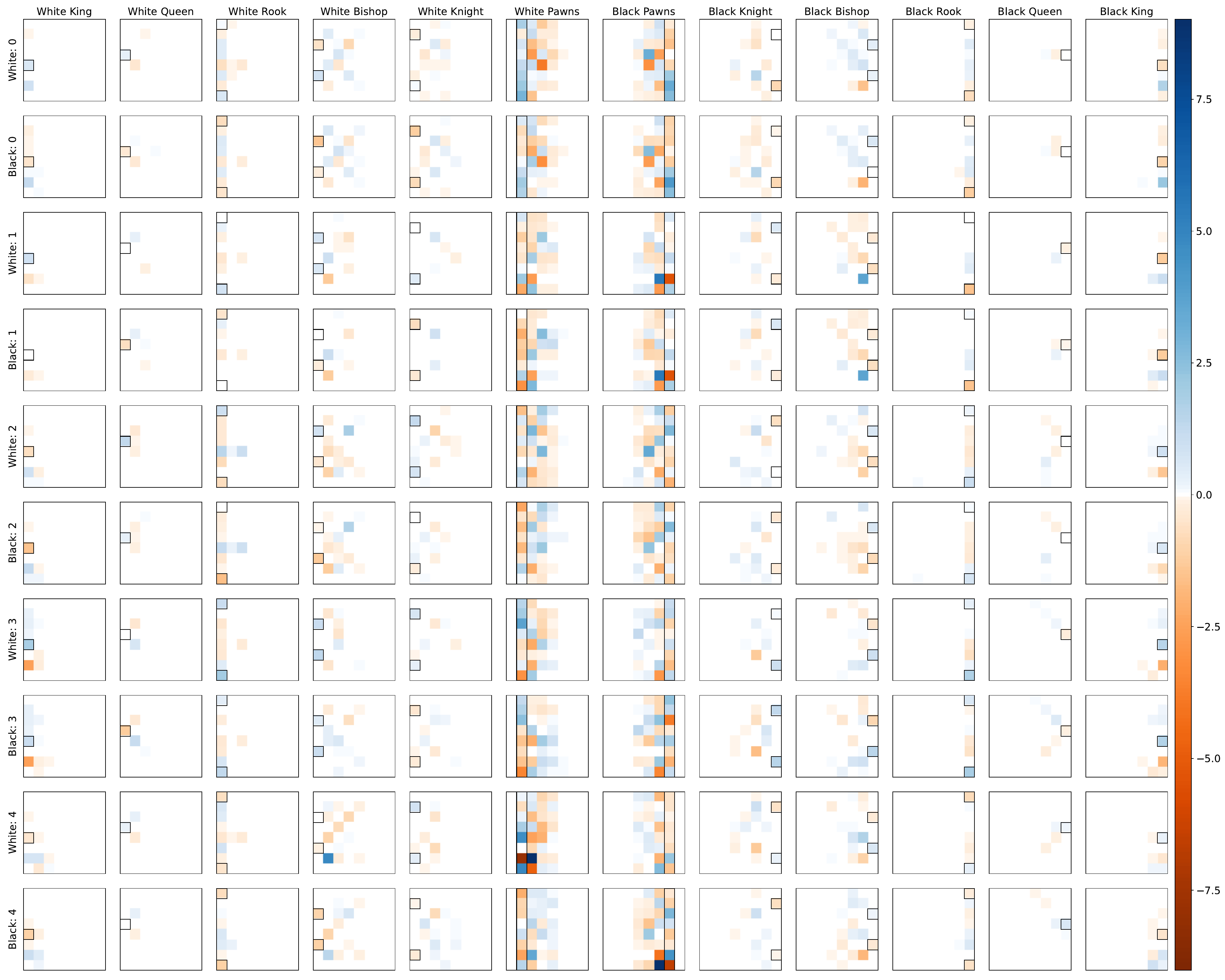}
    \caption{Mean-subtracted piece occupancies of \cz: $\tilde \psi ^i = \psi ^i - \frac{1}{\nplayers}\sum \psi^i$. Columns correspond to pieces: leftmost five as white and rightmost as black. Rows correspond to players playing as white (odd rows) and black (even rows). Blue color implies that a specific player has higher occupancy than the average, and red color implies lower occupancy. Black rectangles highlight the starting position of each piece. Color units (in the colorbars) correspond to occupancy in $\%$ multiplied by the average number of moves in a game (65). For orientation, the chess coordinate system is displayed in Figure 3 in the main paper, top-left.}
    \label{fig:sfs_diff}
\end{figure}

\clearpage
\subsection{Details on self-play from puzzle positions.}
\label{subsec:details_train_on_test}
Table 1 in the main paper presents the performance of AZ on the challenging puzzles sets (Challenge and Penrose) when we allow AZ to begin self-play games from puzzle positions. Here we further explain the details for each one of the variations in the table.

\textit{Self-play from puzzle positions}: This agent uses a \textit{start-position sampler} that samples an initial board position from the following distribution: with $\Prob = 0.5$ it uses the standard initial chess board and with $\Prob = 0.5$ it uniformly samples a position from the Challenge and Penrose Sets. Note that we do not give the solutions to the puzzles to the agent in any form. AZ is allowed to train on self-play games that it plays from these positions. 

\textit{+ intermediate positions}: While the previous agent used the first position from each puzzle in the \textit{start-position sampler}, this agent also samples intermediate positions from each puzzle in the Challenge set. Since those puzzles have a unique multi-step solution, adding intermediate positions allows the agent to play on more advanced steps of a puzzle and can help the agent with exploration.

\textit{+ exploration}: This agent additionally uses an exploration mechanism to sample actions during training. The agent samples actions from a softmax distribution ($\tau=10$; identical setup for exploration is used in AZ \citep{silver2018general}) constructed over the MCTS visit-counts $N^0(s, a)$ for the first $15$ steps of an episode. After the first $15$ steps, the agent continues to sample actions greedily from its visit-counts.

\textit{+ half-move clock}: The half-move clock in chess counts how many moves have been played since the last move where a pawn advanced or a piece was captured. When this counter reaches $100$ the game ends in a draw, which is known as the $50$-move draw rule. The closer this counter is to $100$, the more likely it is that a position will end up in a draw. Whenever we introduce positions in the start position matchmaker, we introduce them with the {\em half-move clock} set to $0$. However, in this experiment, we also included puzzle positions with half-move clocks set from $0$ to $80$, in increments of $20$. This data augmentation should make it easier for the agent to learn that some positions are leading to a draw.

\textit{Train/test splits.} In the middle panel of Table 1, we conducted two more experiments to understand the generalization ability of AZ. The first experiment focused on in distribution generalization, \ie,  AZ used a start-position sampler that consisted of $36$ positions from the Penrose set and was evaluated on the remaining $17$ held-out positions. This split implies that AZ had access to some of the variations of each Penrose position in its training set and the others in its test set (see Tables \ref{tbl:penrose1_1}-\ref{tbl:penrose1_8}). Inspecting the results, we can see that the agent achieves 88.24\% (see (3) in Table 1 in the main paper) suggesting that it generalizes well in distribution (across variations of the same position). The second experiment focused on OOD generalization, where AZ used all the puzzles from the Penrose set except the original (position 2) Penrose puzzle position and its variations (training set = $36$ puzzles; test set = $17$ puzzles). This agent achieved a performance of 11.76\% on the test set (see (4) in Table 1 in the main paper). 

\clearpage
\subsection{Additional results on AZ's understanding of puzzles}

\begin{figure}[b!]
\centering
    \includegraphics[width=0.6\linewidth]{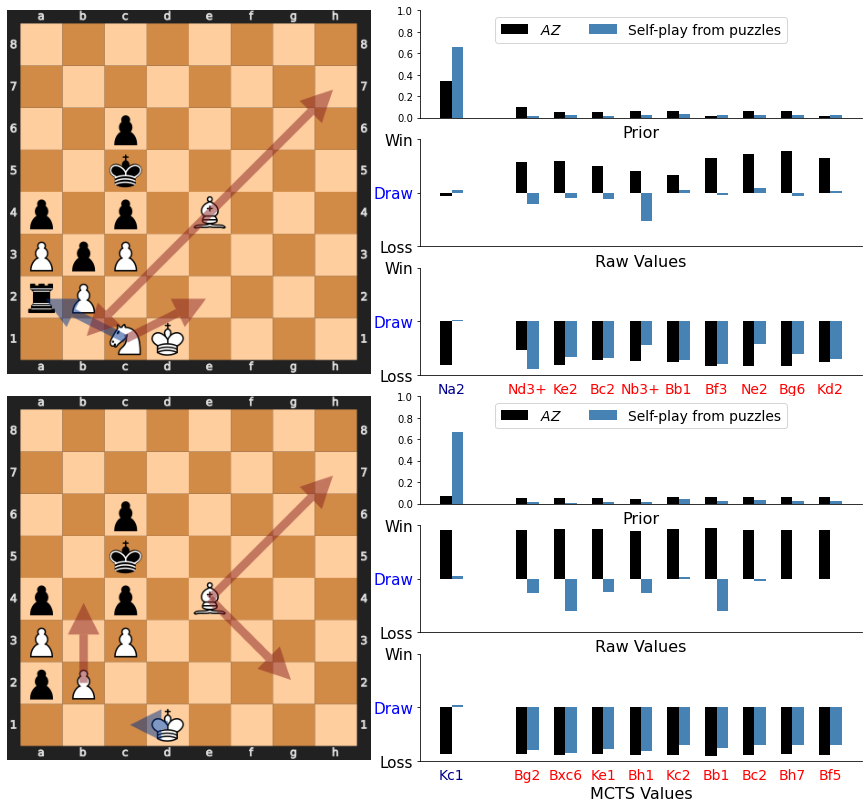}
    \caption{{\em Left:} Two puzzle positions from the Challenge set. {\em Right:} Visualizations of prior probability from a subset of legal actions, and its corresponding raw value and MCTS value estimates for AZ and AZ trained on puzzle sets. Correct and Incorrect moves are colored in Blue and Red respectively.}
    \label{fig:puzzles_hard_p2}
\end{figure}

In the main paper, we inspected AZ's understanding of the Penrose positions. We now complement this analysis with a position from the Challenge set in \cref{fig:puzzles_hard_p2}. 

The first and second rows in \cref{fig:puzzles_hard_p2} shows the first and third steps respectively of the Challenge set puzzle. Inspecting the first step we can see that AZ has a higher probability for selecting the correct action Nxa2 and also an accurate raw value estimate for this move (draw). However, the raw value estimates for the other actions (which are not part of the solution) suggest that the agent can win if it played any of them, and after MCTS, AZ predicts that all actions are equally likely to lose the game. 

In the third step of this puzzle (second row), it is white's turn to play, black has the possibility of promoting its a2 pawn in the next move, and thus, it is important to pick the correct move here. We indicate some of the legal moves with arrows, where red arrows indicate wrong moves that lead to a loss and the blue arrow shows the unique, correct move (Kc1) that leads to a draw (if black promotes the a2 pawn to a queen, white responds with Bishop b1 that locks the position). 

AZ again selects the wrong move, it has low prior probability for all the legal actions available from this position, and more importantly, its raw value estimates predict that it is a win for all of those legal moves! When AZ trained on puzzles with self-play, it correctly predicts the action (in terms of prior probability) and also understands the position correctly (measured in terms of raw and MCTS values).

\clearpage
\subsection{Sub-additive planning in chess matchups}

\begin{figure}[h!]
    \centering
    \includegraphics[width=0.5\linewidth]{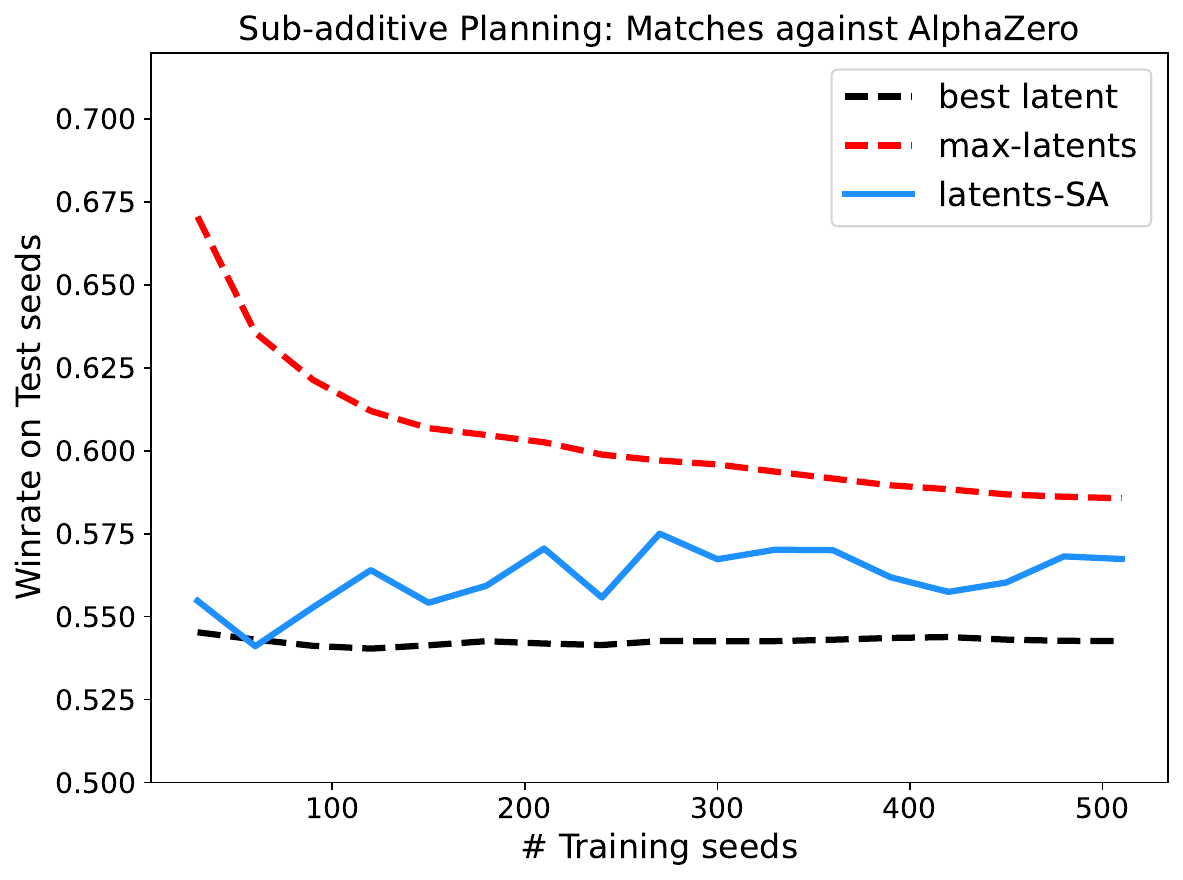}
    \caption{Sub-additive planning in chess matchups against AZ. Reports the average winrate of \cz against AZ, generated from $500$ random seeds. Also shows the winrate of the best latent (in Black) and \textit{max-over-latents} (in Red) across those matchups.}
    \label{fig:sub_additive_match_ups}
\end{figure}

In this ablation experiment, we looked at the performance of sub-additivity as a function of the number of seeds in matchups. We looked at the averaged scores from matchups across opening positions generated using a certain number of random seeds and, from those matchups, defined a look-up table that mapped from an opening position and \cz's player color to the player from the team with the highest score against AZ. This look-up table was used to select a player on hold-out matchups from a distinct set of $30$ seeds. This answers the question of whether such a sub-additive approach can generalize to a hold-out set of matchups and more broadly answers the question of whether different players specialize to specific opening positions. The results from this study are presented in \cref{fig:sub_additive_match_ups}, where the average win rate (scores in $[-1, 1]$ are transformed to $[0, 1]$) against AZ as a function of \textit{training seeds}, \ie, the number of matchups that were used to define the look-up table for sub-additive player selection. The figure also shows in dashed horizontal lines the average win rate of the best latent (black) and \textit{max-over-latents} (red). From the plot, we see that when players are selected sub-additively using the look-up table (constructed from $90$ seeds), the win rate of \cz improves to $0.598$ on the hold-out set of matchups (from $30$ random seeds). This finding is similar to the one that was previously described in Figure 1 in the main paper (Right): sub-additive planning with players in \cz improves performance also in matches, while the previous experiments demonstrated this only in puzzles. The figure also shows that sub-additive planning in matchups improves as a function of training seeds (that are used to construct the look-up table). 

\begin{figure}[b!]
    \centering
    \includegraphics[width=0.5\linewidth]{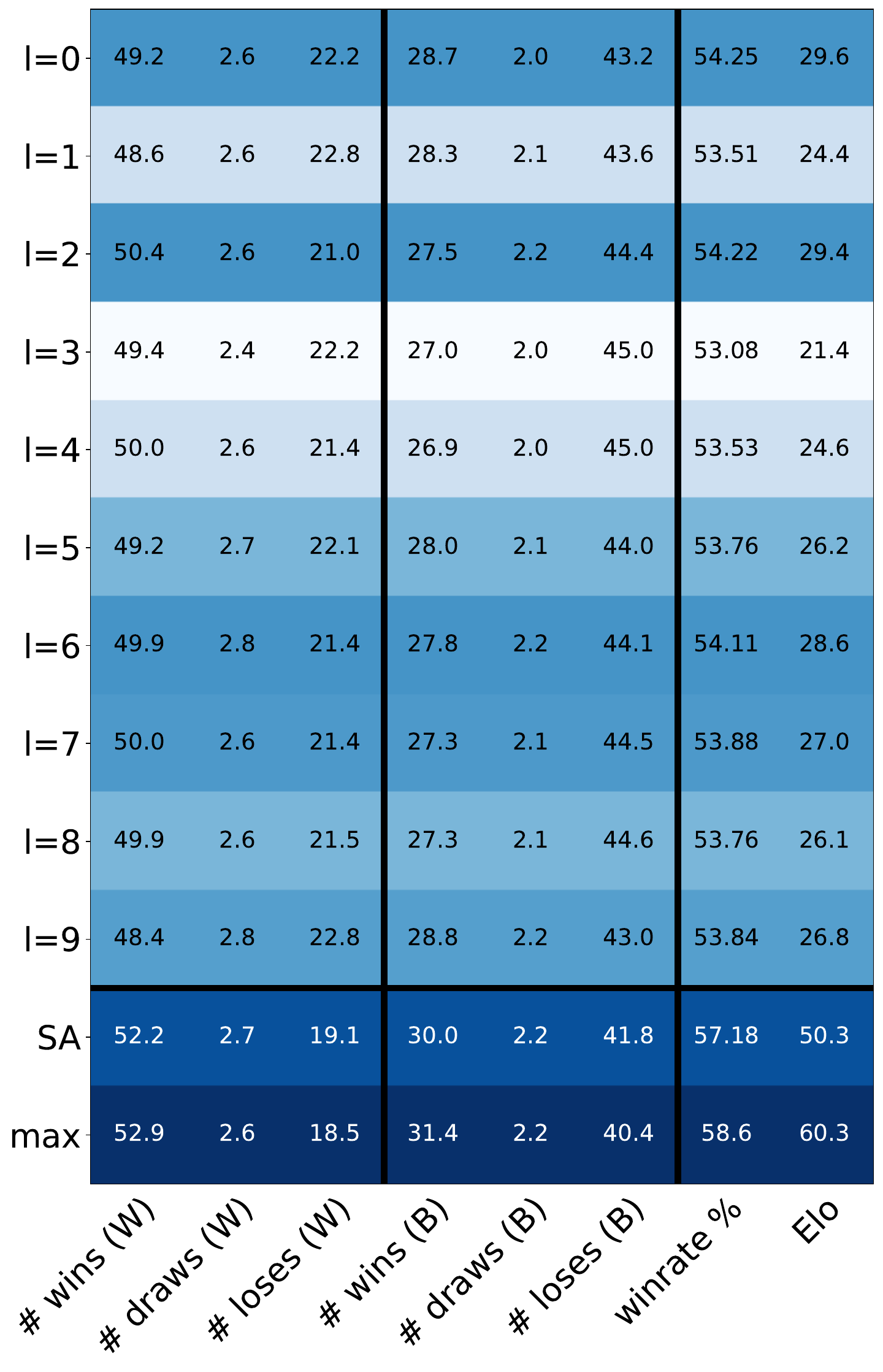}
    \caption{Summary of chess matchups between \cz and AZ. The first ten rows reports the average of the match outcomes (i.e., wins, draws and loses) for each player in the team. The last two rows reports the same for sub-additive planning and \textit{max-over-latents}. In addition to game outcomes, we also report the average winrate and relative Elo improvement over AZ.}
    \label{fig:matchup_summary}
\end{figure}

For each opening and seed, we calculated the average score of each player across all other seeds (leaving one seed out) and defined a mapping from an opening to a player with the highest score. We then evaluated this mapping on the held-out seed of matchup. This process is repeated for each seed that is held-out for evaluation and the scores obtained from the evaluation are averaged and reported as the sub-additive planning scores in \cref{fig:matchup_summary}.

\clearpage

\clearpage
\begin{figure}[h]
\centering
    \subfloat{{\includegraphics[width=0.38\linewidth]{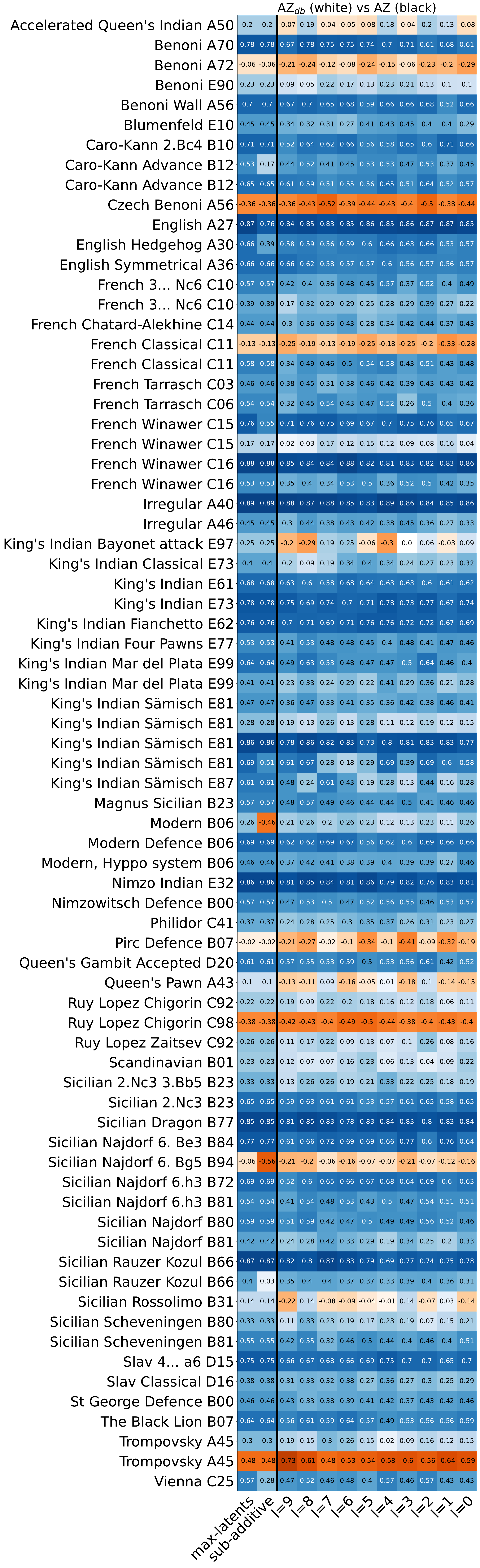} }}%
    \subfloat{{\includegraphics[width=0.386\linewidth]{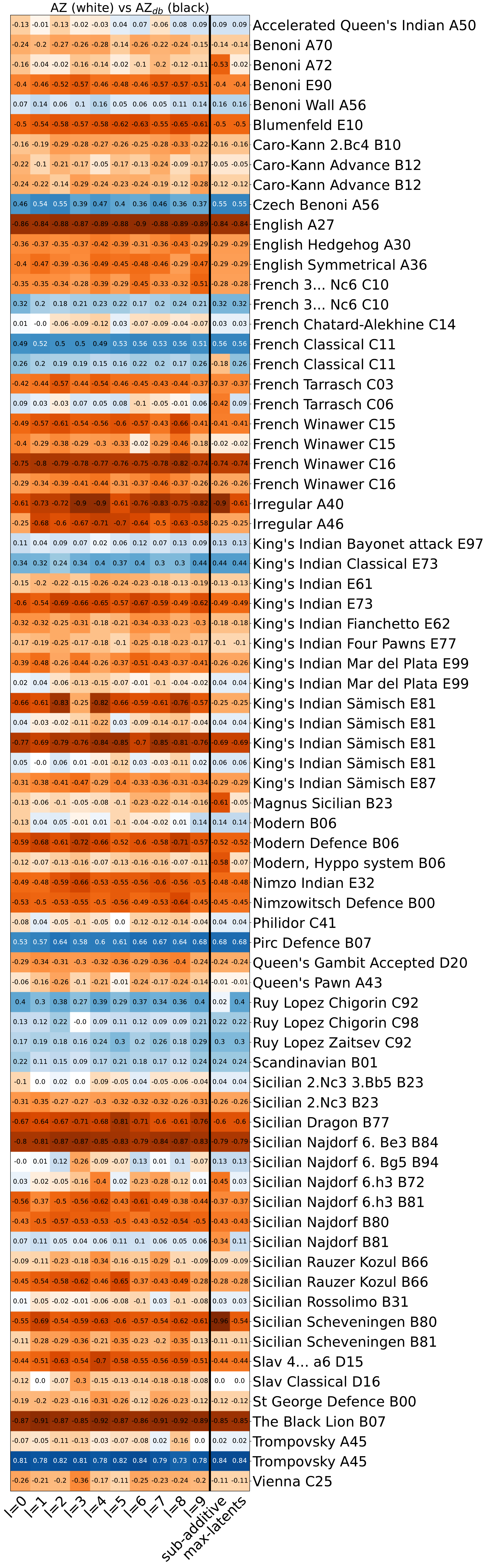} }}%
    \caption{Results from different chess openings between \cz and AZ (scores averaged from 30 random seeds). In the figure on the \textit{left}, \cz played as White and AZ as Black; the colors were flipped for the figure in \textit{right}.}
    \label{fig:cz_vs_az_400sims_matchups}
\end{figure}
\clearpage

\begin{table}[h!]
\caption{Average winrates of each player in \cz when playing as White and as Black against AZ.}
\centering
    \begin{tabular}{p{0.5\linewidth} | p{0.1\linewidth} || p{0.1\linewidth} } 
     \specialrule{.2em}{.1em}{.1em} 
      \hspace{3cm} \cz team & as White & as Black \\
     \specialrule{.2em}{.1em}{.1em} 
     \hspace{3cm} Player $0$ & 68.27\% & 40.21\% \\
     \hspace{3cm} Player $1$ & 67.39\% & 39.63\% \\
     \hspace{3cm} Player $2$ & \textbf{69.86\%} & 38.57\% \\
     \hspace{3cm} Player $3$ & 68.31\% & 37.83\% \\
     \hspace{3cm} Player $4$ & 69.26\% & 37.79\% \\
     \hspace{3cm} Player $5$ & 68.32\% & 38.19\% \\
     \hspace{3cm} Player $6$ & 69.23\% & 38.98\% \\
     \hspace{3cm} Player $7$ & 69.36\% & 38.38\% \\
     \hspace{3cm} Player $8$ & 69.24\% & 38.26\% \\
     \hspace{3cm} Player $9$ & 67.30\% & \textbf{40.38\%} \\
     \specialrule{.2em}{.1em}{.1em} 
    \end{tabular}
\label{table:matchup_stats}
\end{table}

\clearpage

\section{Puzzle sets}

The \textit{Lichess} data set is composed of $560$ puzzles that we selected from a larger data set provided by  \href{https://database.lichess.org/#puzzles}{lichess.org}. The complete data set includes $3M$ puzzles collected from $300M$ analysed games from the Lichess database and re-analyzed with SF. We selected $560$ puzzles from them by creating a subset of the $20K$ hardest puzzles (rated by "ELO" score) and then further selecting from them the positions that were unsolved by SF15 with a low number of simulations ($10k$). Each puzzle in the set requires multiple steps; the agent receives a score of $1$ if it predicts the correct unique action in all the steps and $0$ otherwise.

The\textit{ Strategic Test Suite (\href{https://sites.google.com/site/strategictestsuite/about?authuser=0}{STS})} is a data set of $1088$ single-step, multi-choice chess puzzles, designed to evaluate chess engine's long term understanding of strategical and positional concepts. The suite was created by Dann Corbit and Swaminathan Natarajan, but we adapt the more recent \href{https://sites.google.com/site/strategictestsuite/about?authuser=0}{STS-Rating (v6)} by Ferdinand Mosca. Actions receive scores $\in[0,1000],$ we normalize the score for each action by the score of the best action ($1000$), so the score in each puzzle is $\in[0,1]$.

The next two data sets were collected and created by our team. They include puzzles from different sources  and we refer the reader to the tables in the end of the Appendix for more details.

The \textit{Challenge set} includes $15$ multi-choice \& step puzzles that are designed to be challenging to modern chess engines. A score of one is given if the agent predicts the correct unique action in each step of the puzzle and $0$ otherwise. We deal with a few cases of ambiguous solutions via a simple concept of "or lists", if the agent finds one of the possible unique multi-step solutions to the puzzle it gets a score of $1$ in this puzzle. Sometimes the opponent has different ways to respond to the agent's moves, in this case, we apply an "and list" and require the agent to find the unique multi-step solution on all the variations. These positions were collected from different sources including  \citep{10positions} and \citep{steingrimsson2021chess}. 

The \textit{Penrose set} includes $53$ challenging single-step puzzles, where the goal is to predict the value of the position (in $[0, 0.5, 1]$). A score of one is given if the value prediction (after a search) is within a threshold (of $0.25$ or $0.1$) from the correct value and $0$ otherwise. The set includes three sources of puzzles: (1) The Penrose position(s): a chess puzzle proposed by Sir Roger Penrose to \emph{"learn more about the uniqueness of the human brain"} \citep{penrose}, can be seen in \cref{tbl:penrose1_4}, top. We also include a second position by Penrose with a similar idea that we found  \href{https://www.youtube.com/watch?v=xGbgDf4HCHU}{here} (\cref{tbl:penrose1_1}, top). (2) A set of variations of the two Penrose positions as well as other puzzles created by our team (Tables \ref{tbl:penrose1_1}-\ref{tbl:penrose1_8}). (3) Fortress positions from \citep{guid2012detecting}, selected from \citep{dvoretsky2010dvoretsky} and blocked positions composed by us (Tables \ref{tbl:blocked1}-\ref{tbl:blocked6}). We also include variations for some of these positions by adding or removing pieces from them, similar to \citep{steingrimsson2021chess}. Together with the Lichess data set, these puzzle sets present a diverse set of challenges for chess engines.

The following tables present positions from the Penrose and Challenge sets respectively. The tables include the starting position of each puzzle, the FEN, the solution, the player to play and comments on each position. The comment provides context on the position and sometimes explain why it may be hard for engines to correctly evaluate. This analysis was performed using a web version of SF, and we emphasize that it is not using SF at its full strength. The goal of these comments is to provide context on why the position is hard to chess programs and we use SF due to its popularity.

\clearpage

\begin{table}[h!]
  \centering
  \begin{tabular}{ | c | m{1.5cm} | m{1.5cm} | m{6cm} |}
    \hline
    Board & Solution & To Play & Comment \\ \hline
    \begin{minipage}{0.3\textwidth}
      \captionsetup{justification=centering}
      \caption*{\tiny{knb5/1b1p4/p1pPb3/P1P5/6p1/6Pp/7P/K7 w - -}}
      \includegraphics[width=0.9\linewidth,]{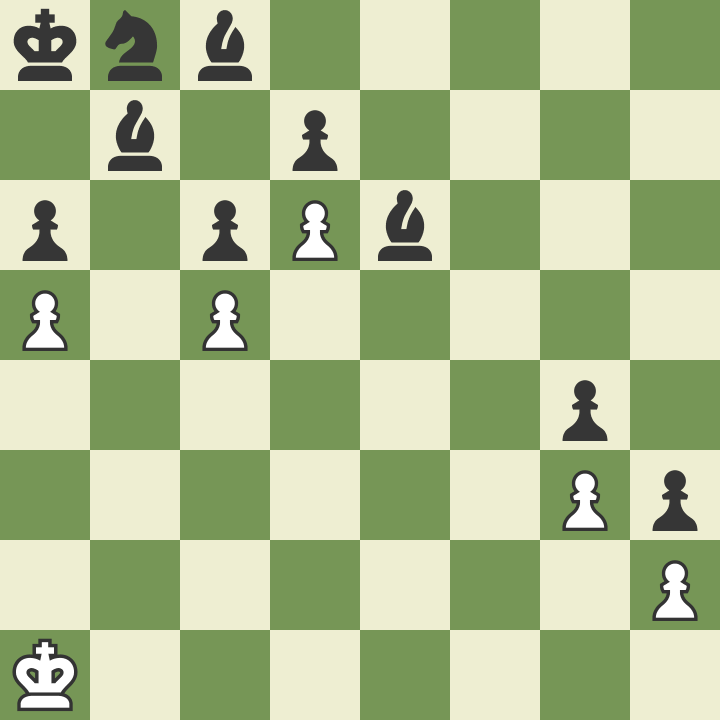}
    \end{minipage}
    & Draw & White &
    White can draw in several ways, for example, by repeating Kb2-a1 to draw via the three-fold repetition rule or the 50 move rule. There is no way for white or black to make progress.  This position would be even be drawn is white picks up the pawns on g4 and h3. The reason is that black can sacrifice the bishop for one of the pawns. The resulting position is a draw because it’s a fortress.
    \\ \hline
    
    \begin{minipage}{.3\textwidth}
      \captionsetup{justification=centering}
      \caption*{\tiny{knb4Q/1b1p4/p1pP2K1/P1P5/8/8/8/8 b - -}}
      \includegraphics[width=0.9\linewidth,]{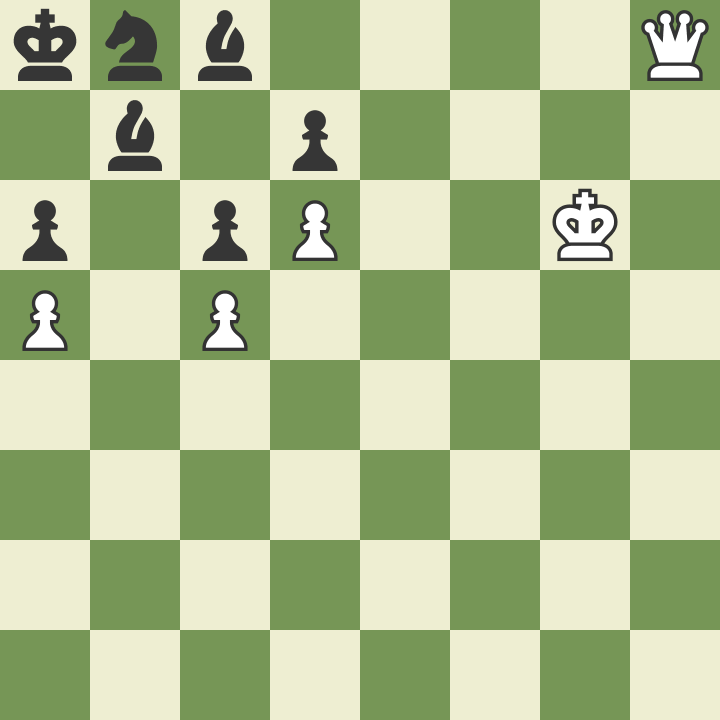}
    \end{minipage}
    
    & 
    Draw
    &
    Black
    &
    A position that arises from a calculation sequence in the baseline puzzle. This position is a fortress, and white cannot make further progress.

    \\ \hline
    
    \begin{minipage}{.3\textwidth}
      \captionsetup{justification=centering}
      \caption*{\tiny{knb5/1b1p1b2/p1pPb3/P1P5/6p1/6Pp/7P/K7 w - -}
      \includegraphics[width=0.9\linewidth,]{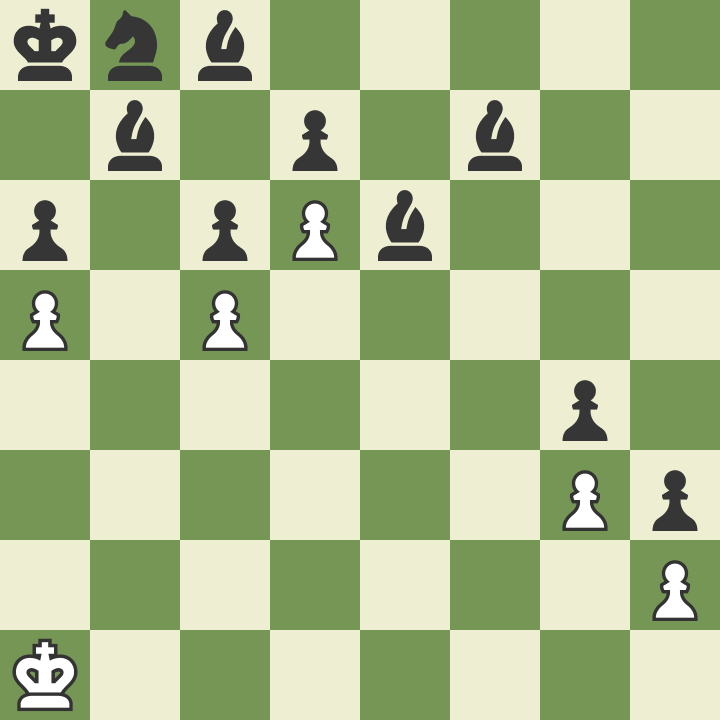}
}
    \end{minipage}
    
    & 
    Draw
    &
    White
    &
    This puzzle is even more difficult for SF, as there is an additional piece meaning more possible moves. However, for humans, it’s clear that the fundamental aspects are the same.
    \\ \hline
    
    \begin{minipage}{.3\textwidth}
      \captionsetup{justification=centering}
      \caption*{\tiny{knb5/pb1p4/p1pPb3/P1P5/6p1/6Pp/7P/K7 w - -}
      \includegraphics[width=0.9\linewidth,]{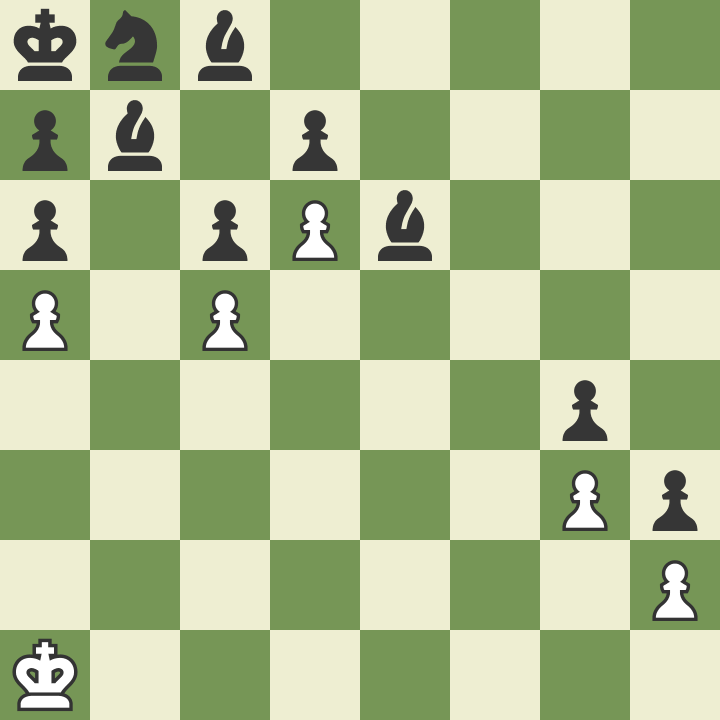}
}
    \end{minipage}
    
    & 
    Draw
    &
    White
    &
    Interestingly enough, this position is easier for Stockfish to evaluate. This is likely because the  pieces in the left top of the board have no moves, and SF recognizes that stalemate (no moves left, resulting in a draw) is a likely outcome.
    \\ \hline
  \end{tabular}
    \captionsetup{justification=centering}
  \caption{Penrose set positions, Penrose variation 1, positions 1-4}\label{tbl:penrose1_1}
\end{table}\clearpage
\begin{table}[h!]
  \centering
  \begin{tabular}{ | c | m{1.5cm} | m{1.5cm} | m{6cm} |}
    \hline
    Board & Solution & To Play & Comment \\ \hline 
    \begin{minipage}{0.3\textwidth}
    \captionsetup{justification=centering}
    \caption*{\tiny{knb5/1b1p4/p1pPp3/P1P2p2/3K2p1/6Pp/7P/8 w - -}}
      \includegraphics[width=0.9\linewidth,]{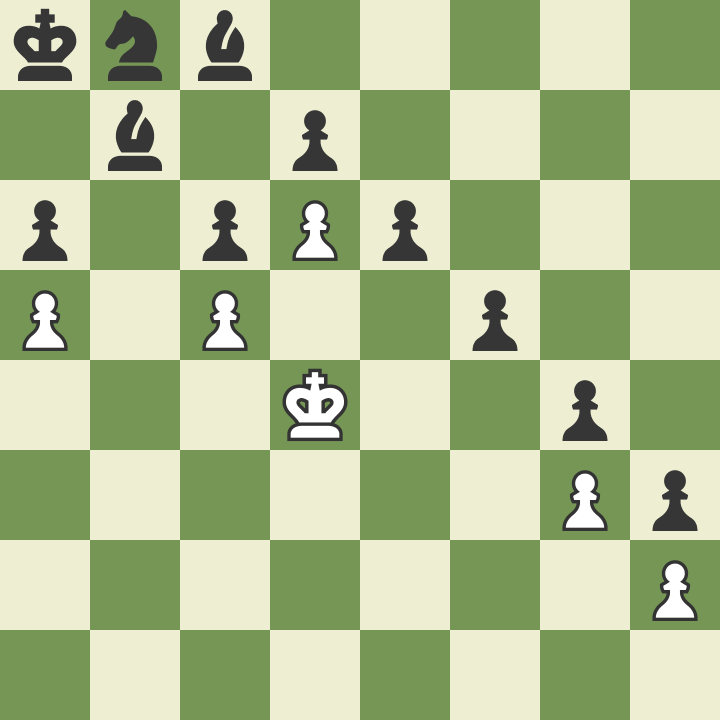}
    \end{minipage}
    & Draw & White &
    This also seems to be an easier position for SF to understand, likely because it’s easier to calculate the relevant variations (contrary to the above example which requires some ‘pattern’ recognition).
    \\ \hline
    \begin{minipage}{0.3\textwidth}
    \captionsetup{justification=centering}
    \caption*{\tiny{knb5/1b1p2p1/p1pP2p1/P1P3p1/6p1/6Pp/7P/K7 w - -}}
      \includegraphics[width=0.9\linewidth,]{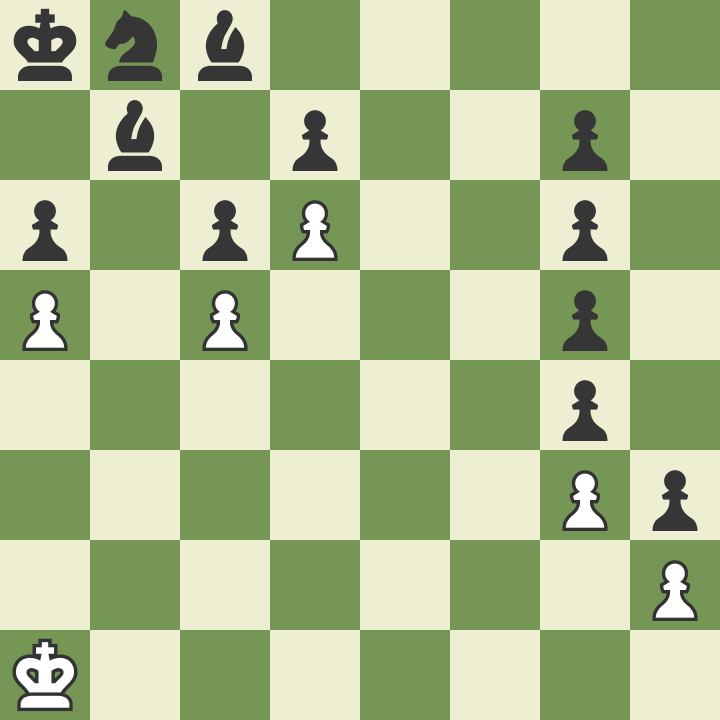}
    \end{minipage}
    & Draw & White &
    Like the position above, this seems to be an easier position for SF to understand. It’s a combination of the two reasons above – fewer variations to calculate, and white has no way to infiltrate black’s position.
    \\ \hline
    \begin{minipage}{0.3\textwidth}
    \captionsetup{justification=centering}
    \caption*{\tiny{knb5/rb1p4/p1pPb3/P1P5/6p1/6Pp/7P/K7 w - -}}
      \includegraphics[width=0.9\linewidth,]{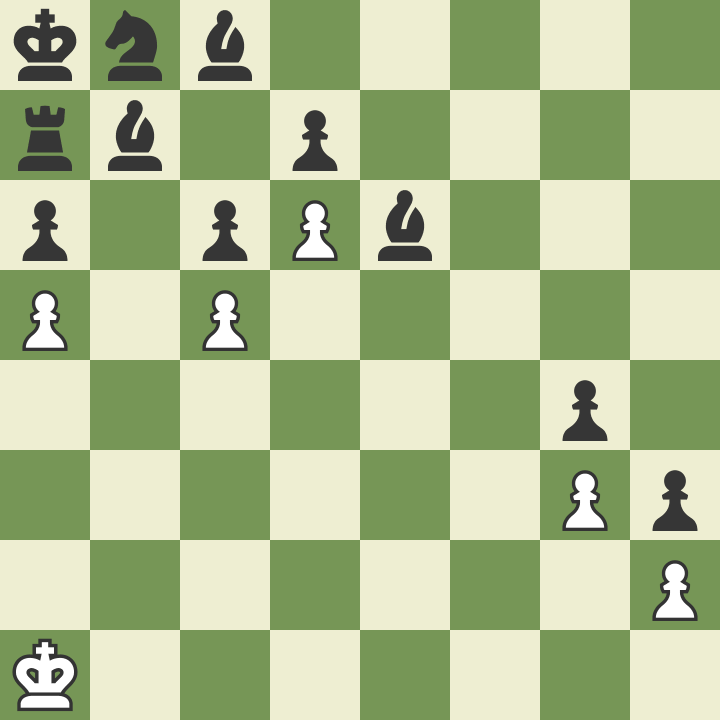}
    \end{minipage}
    &
    Draw
    &
    White
    &
    Similar to the position with the added pawn on a7 (three rows up), this position seems to be easier for SF despite the large material increase. 
    \\ \hline
    \begin{minipage}{0.3\textwidth}
    \captionsetup{justification=centering}
    \caption*{\tiny{knb5/1b1p4/pPpPb3/P1P5/6p1/6Pp/7P/K7 w - -}}
      \includegraphics[width=0.9\linewidth,]{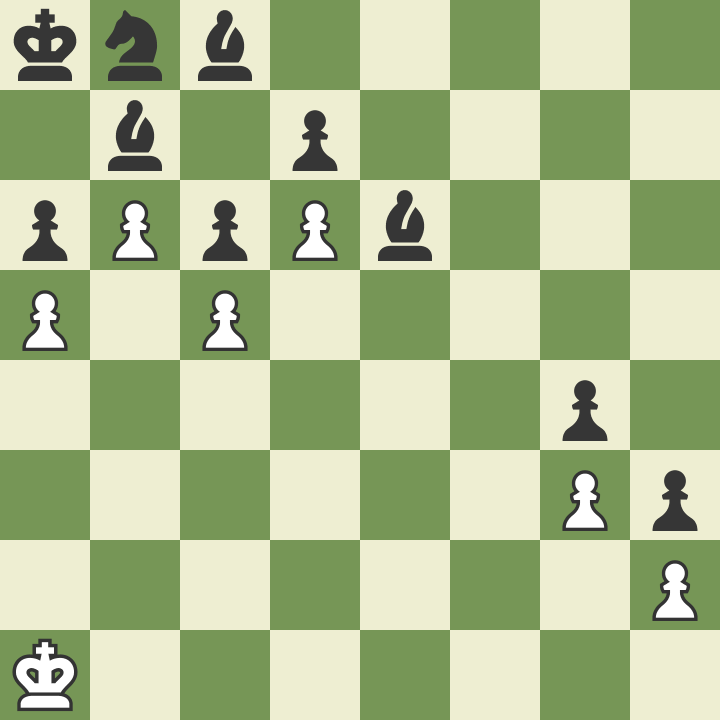}
    \end{minipage}
    & Draw & White &
    This seems to be an easier position to understand (than the original puzzle), suggesting that perhaps stalemates are likely an easier concept to grasp than fortresses. 
  \\ \hline
  \end{tabular}
  \captionsetup{justification=centering}
  \caption{Penrose set positions, Penrose variation 1, positions 5-8}\label{tbl:penrose1_2}
\end{table}\clearpage
\begin{table}[h!]
  \centering
  \begin{tabular}{ | c | m{1.5cm} | m{1.5cm} | m{6cm} |}
    \hline
    Board & Solution & To Play & Comment \\ \hline
    \begin{minipage}{0.3\textwidth}
    \captionsetup{justification=centering}
    \caption*{\tiny{1nb5/kb1p4/pNpPb3/P1P5/6p1/6Pp/7P/K7 w - -}}
      \includegraphics[width=0.9\linewidth,]{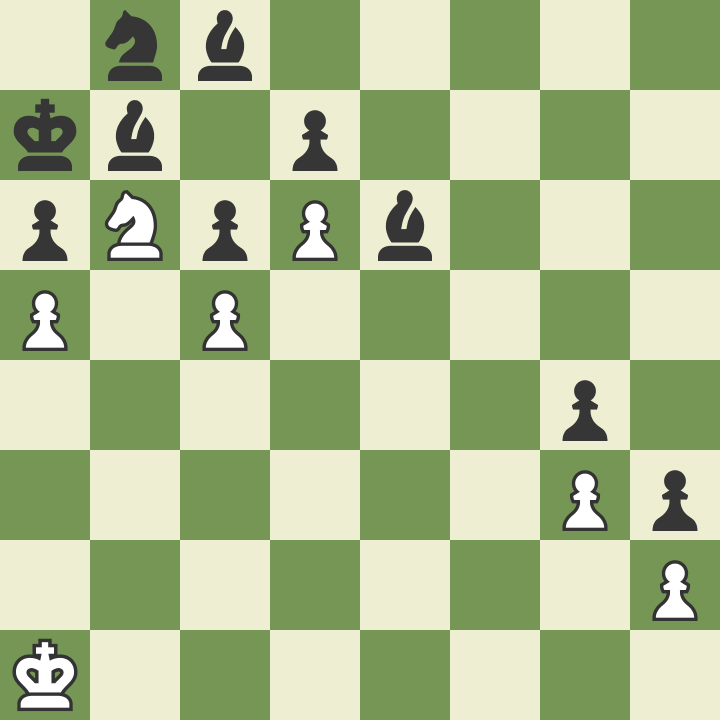}
    \end{minipage}
    & Win & White &
    Knight a4 keeps the position locked and allows the white king to fight with the knight against the black bishop on e6 and pick up the pawns.
    \\ \hline
    \begin{minipage}{0.3\textwidth}
    \captionsetup{justification=centering}
    \caption*{\tiny{1nb5/kb1p4/pNpPb3/P1P5/5Kp1/6Pp/7P/8 w - -}}
      \includegraphics[width=0.9\linewidth,]{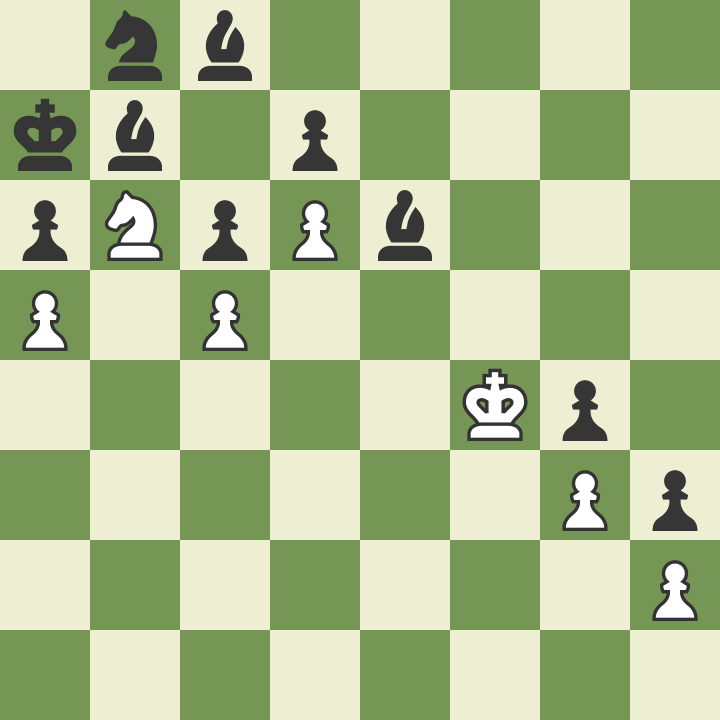}
    \end{minipage}
    & Win & White & Here white has a concrete way to win: 1. Kg5 Bb3 2. Kxg4. After that, white can carefully advance the g-pawn, and promote the g-pawn and/or pick up the h-pawn. The knight on b6 is important for two reasons: [1] it prevents black from having a fortress and [2] allows white to control when black can play Ka7-Ka8. SF misevaluates this as a draw.
    \\ \hline
    \begin{minipage}{0.3\textwidth}
    \captionsetup{justification=centering}
    \caption*{\tiny{knb5/1b1p4/p1pP4/P1P5/6p1/6Pp/7P/K7 w - -}}
      \includegraphics[width=0.9\linewidth,]{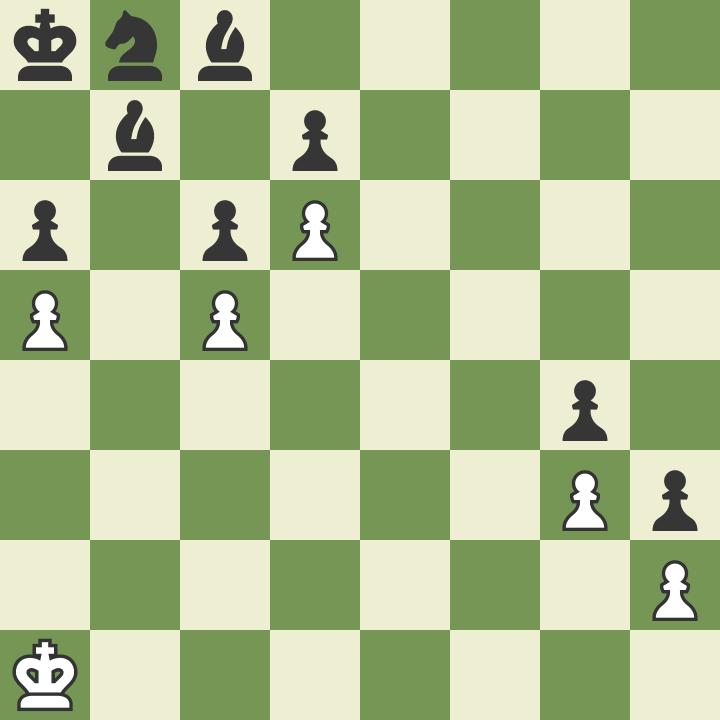}
    \end{minipage}
    & Win & White &
    This position seems to be easy for SF to find the correct move. White can win by walking the king over to the right side on the board, picking up both pawns and then using the two queens to mate black (or using the pieces to take black’s pieces and the pawns on d7 and c6).
    \\ \hline
    \begin{minipage}{0.3\textwidth}
    \captionsetup{justification=centering}
    \caption*{\tiny{kn6/1b1p4/p1pPb3/P1P5/6p1/6Pp/7P/K7 w - -}}
      \includegraphics[width=0.9\linewidth,]{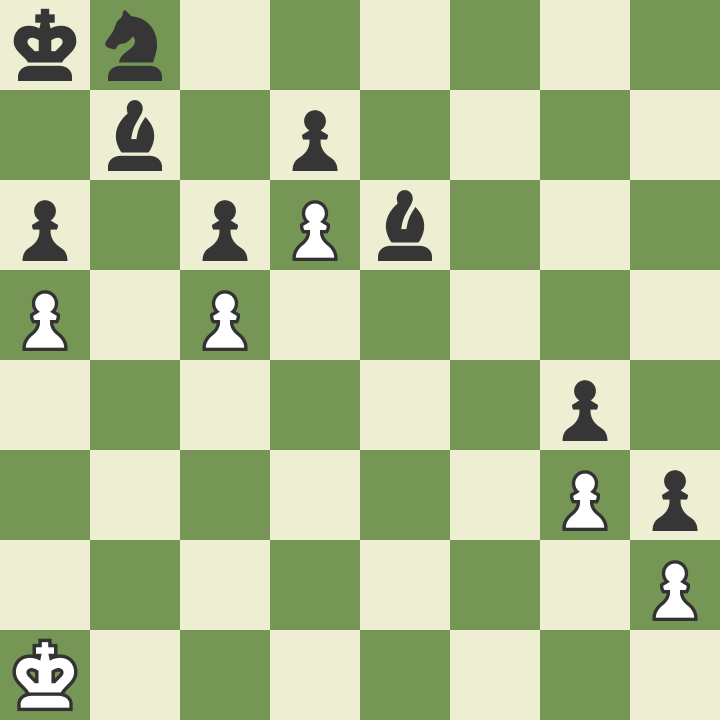}
    \end{minipage}
    & Lose & White &
    The black king can escape by moving the b7-bishop to a8. From there its an easy win. 
    \\ \hline
  \end{tabular}
  \captionsetup{justification=centering}
  \caption{Penrose set positions, Penrose variation 1, positions 9-12}\label{tbl:penrose1_3}
\end{table}\clearpage

\begin{table}[h!]
  \centering
  \begin{tabular}{ | c | m{1.5cm} | m{1.5cm} | m{6cm} |}
    \hline
    Board & Solution & To Play & Comment \\ \hline
    \begin{minipage}{0.3\textwidth}
    \captionsetup{justification=centering}
    \caption*{\tiny{8/p7/kpP5/qrp1b3/rpP2b2/pP4b1/P3K3/8 w - -}}
      \includegraphics[width=0.9\linewidth,]{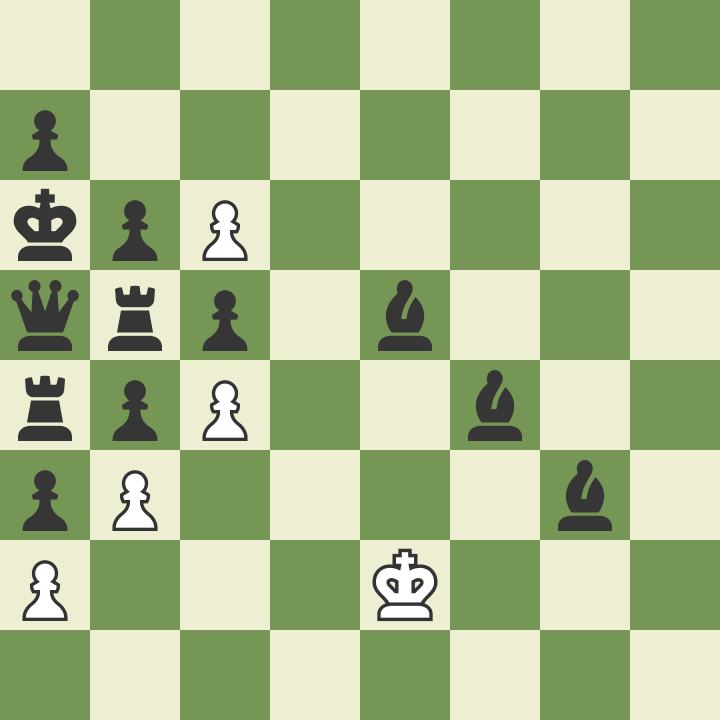}
    \end{minipage}
    & Draw & White &
    The Penrose position. As white has control over the white squares, and black has control over the black squares, neither side can make any progress to win the game. As such, this position is a draw.
    \\ \hline
    \begin{minipage}{0.3\textwidth}
    \captionsetup{justification=centering}
    \caption*{\tiny{8/p7/kpP2b2/qrp1b3/rpP2b2/pP4b1/P3K3/8 w - -}}
      \includegraphics[width=0.9\linewidth,]{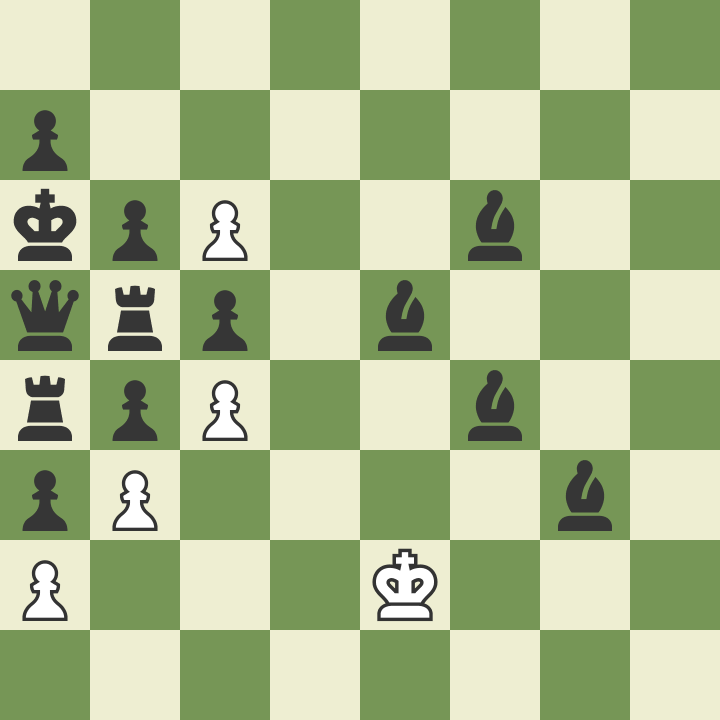}
    \end{minipage}
    & Draw & White &
    Compared to the original penrose position, we have further increased the number of bishops leading to a larger material advantage for black. However, the additional pieces do not help black – black cannot use them to make further progress. As such, the positions is still a draw.
    \\ \hline
    \begin{minipage}{0.3\textwidth}
    \captionsetup{justification=centering}
    \caption*{\tiny{5b2/p5b1/kpP2b1b/qrp1b3/rpP2b2/pP4b1/P3K3/8 w - -}}
      \includegraphics[width=0.9\linewidth,]{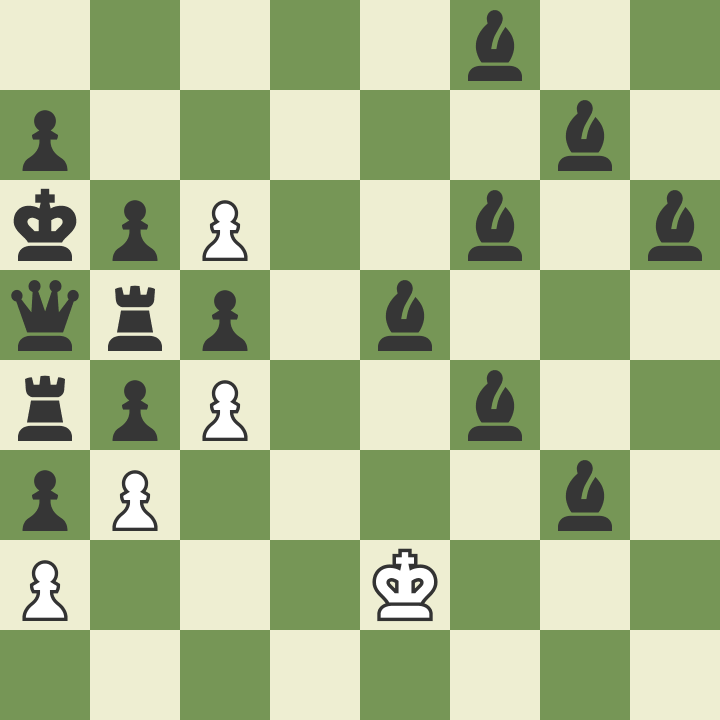}
    \end{minipage}
    & Draw & White &
    In this position, there are even more bishops compared to the position above. However, the position is still a draw. This position is more difficult to evaluated with calculation alone, as there are more possible move sequences. However, if the engine understands that neither side can make any progress towards winning, then it can correctly evaluate the position.
    \\ \hline
    \begin{minipage}{0.3\textwidth}
    \captionsetup{justification=centering}
    \caption*{\tiny{8/p7/kpP5/qrp1b3/rpP5/pP6/P3K3/8 w - -}}
      \includegraphics[width=0.9\linewidth,]{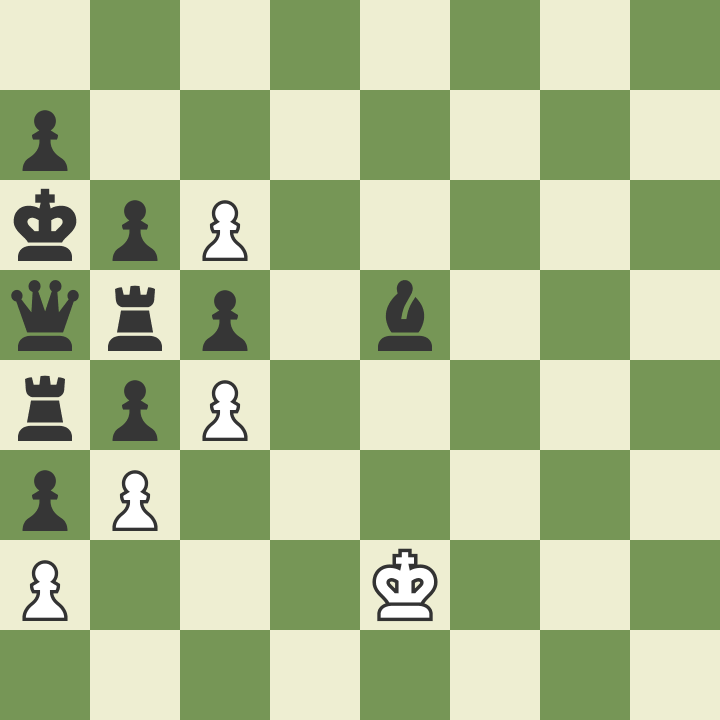}
    \end{minipage}
    & Draw & White &
    Similar to the original position, however black only has one bishop. SF recognizes that this position is a draw, despite the fact that black has large material advantage.
    \\ \hline
  \end{tabular}
  \captionsetup{justification=centering}
  \caption{Penrose set positions, Penrose variation 2, positions 1-4}\label{tbl:penrose1_4}
\end{table}\clearpage

\begin{table}[h!]
  \centering
  \begin{tabular}{ | c | m{1.5cm} | m{1.5cm} | m{6cm} |}
    \hline
    Board & Solution & To Play & Comment \\ \hline
    \begin{minipage}{0.3\textwidth}
    \captionsetup{justification=centering}
    \caption*{\tiny{8/p7/kpP5/qrp1b3/rpP2b2/pP6/P3K3/8 w - -}}
      \includegraphics[width=0.9\linewidth,]{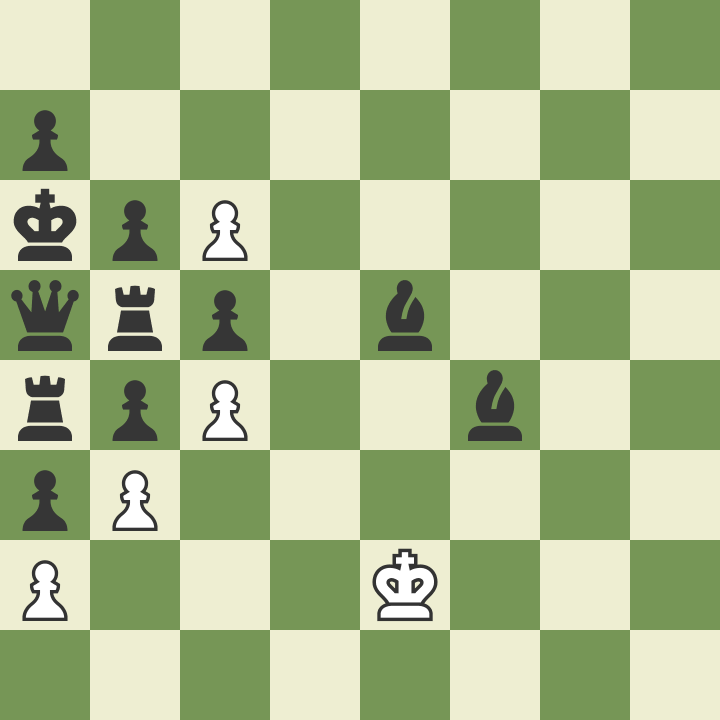}
    \end{minipage}
    & Draw & White &
    Similar to the original position, however black has two bishops. However, SF gets this wrong. 
    \\ \hline
    \begin{minipage}{0.3\textwidth}
    \captionsetup{justification=centering}
    \caption*{\tiny{8/p7/kpP2p2/qrp1bp2/rpP2p2/pP6/P3K3/8 w - -}}
      \includegraphics[width=0.9\linewidth,]{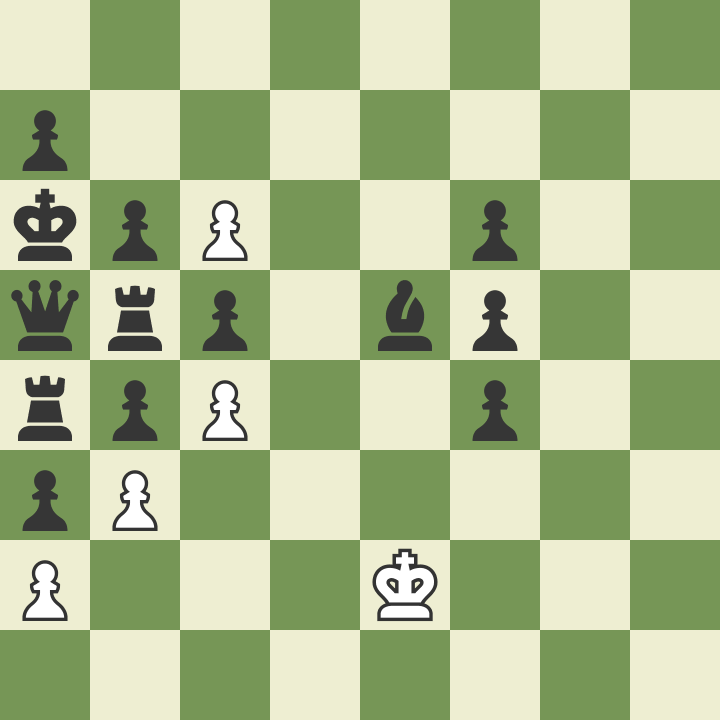}
    \end{minipage}
    & Draw & White &
    SF understands that this is a draw. This is surpising given that the evaluation of the previous positions. Black's material advantage is the same as the position above.
    \\ \hline
    \begin{minipage}{0.3\textwidth}
    \captionsetup{justification=centering}
    \caption*{\tiny{8/p4p2/kpP2p2/qrp1bp2/rpP2p2/pP6/P3K3/8 w - -}}
      \includegraphics[width=0.9\linewidth,]{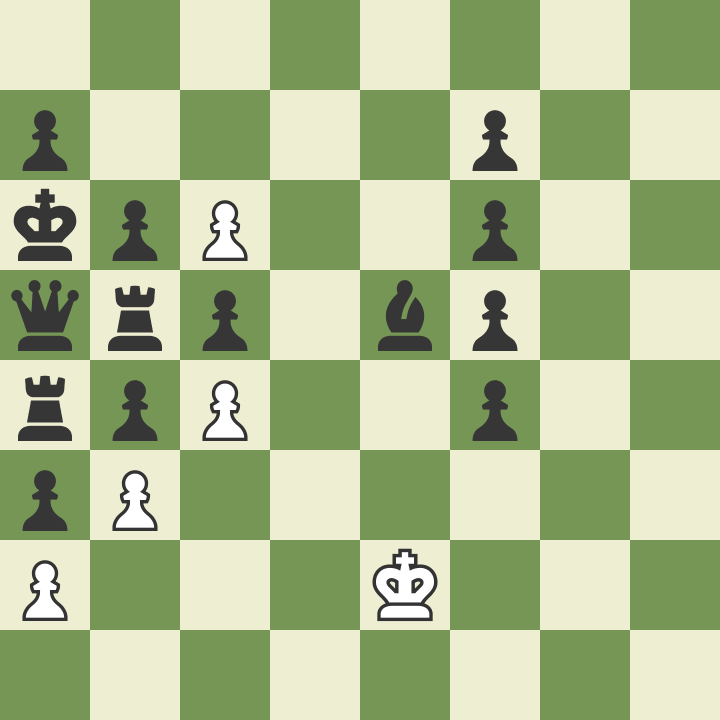}
    \end{minipage}
    & Draw & White &
    SF does not understand that this is a draw, suggesting that there is a material threshold beyond which point engines may become confused.
    \\ \hline
    \begin{minipage}{0.3\textwidth}
    \captionsetup{justification=centering}
    \caption*{\tiny{8/p7/kpP5/qrpKb3/rpP2b2/pP4b1/P7/8 w - -}}
      \includegraphics[width=0.9\linewidth,]{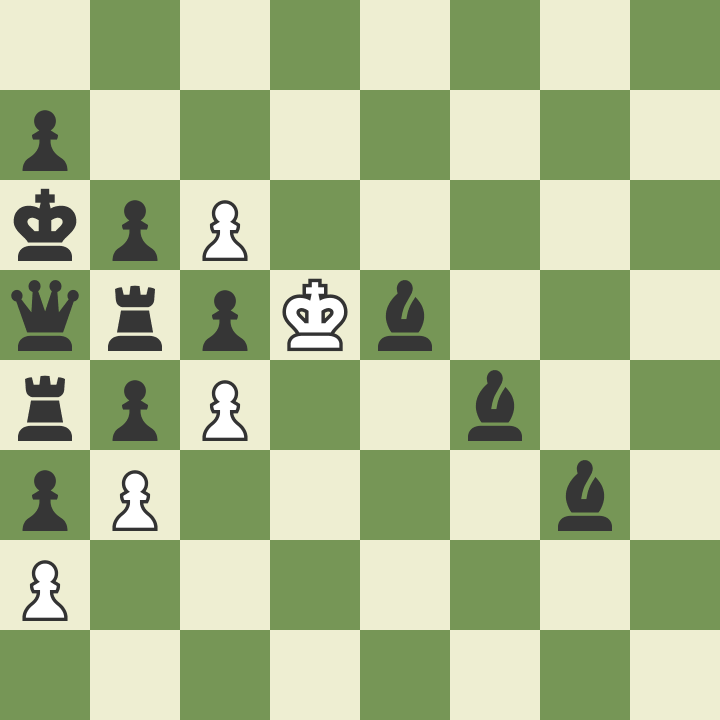}
    \end{minipage}
    & Draw & White &
    The king's position in the center of the board makes SF evaluate the position even worse for white (compared to the original Penrose position). However, the kind's position does not affect the result. 
    \\ \hline
  \end{tabular}
  \captionsetup{justification=centering}
  \caption{Penrose set positions, Penrose variation 2, positions 5-8}\label{tbl:penrose1_5}
\end{table}\clearpage

\begin{table}[h!]
  \centering
  \begin{tabular}{ | c | m{1.5cm} | m{1.5cm} | m{6cm} |}
    \hline
    Board & Solution & To Play & Comment \\ \hline
    \begin{minipage}{0.3\textwidth}
    \captionsetup{justification=centering}
    \caption*{\tiny{8/p7/kpP5/1rp1b3/rpB2b2/pP4b1/P3K3/8 w - -}}
      \includegraphics[width=0.9\linewidth,]{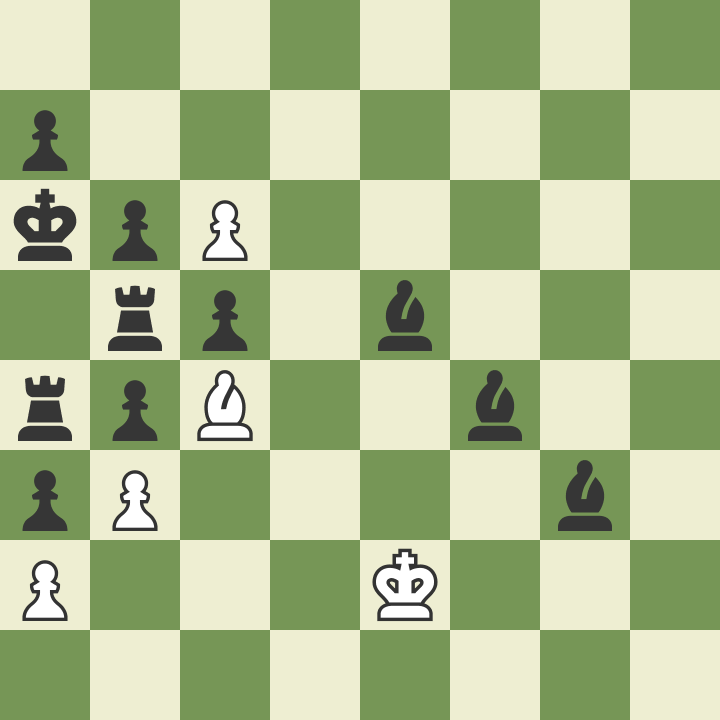}
    \end{minipage}
    & Draw & White &
    Black has more space on the king side, compared to the original Penrose position, however it does not help Black. This position still evaluates to a draw.
    \\ \hline
    \begin{minipage}{0.3\textwidth}
    \captionsetup{justification=centering}
    \caption*{\tiny{8/p7/kpP5/1rpKb3/rpB1Bb2/ pP1B1Bb1/P1B1B1B1/3B1B2 w - -}}
      \includegraphics[width=0.9\linewidth,]{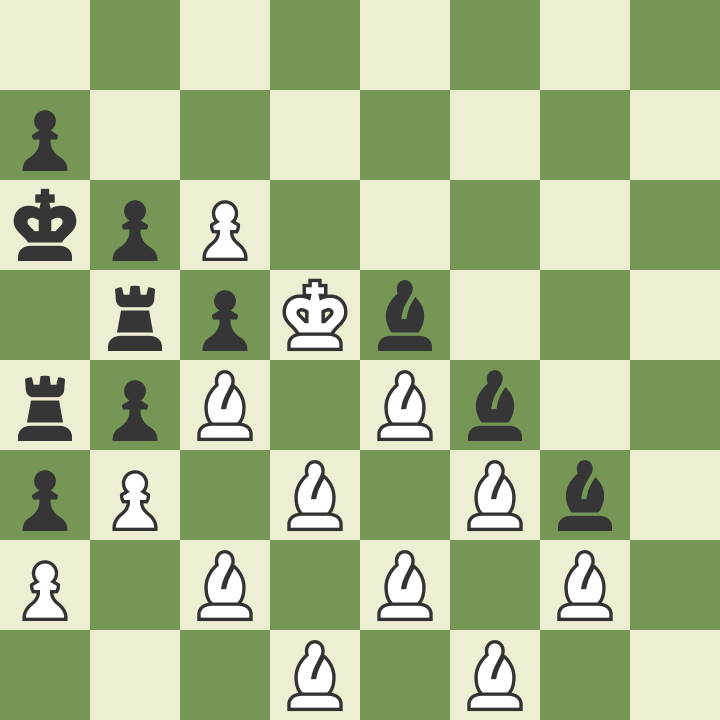}
    \end{minipage}
    & Draw & White &
    In this position, we change the Penrose position to add more material for white, such that white has a material advantage. However, it is still a draw – white only controls the white squares. As black has control over the black squares, white cannot progress. SF thinks that white is winning. 
    \\ \hline
    \begin{minipage}{0.3\textwidth}
    \captionsetup{justification=centering}
    \caption*{\tiny{8/p7/kpP5/1rpKb3/rpB1Bb2/pP3Bb1/P7/8 w - -}}
      \includegraphics[width=0.9\linewidth,]{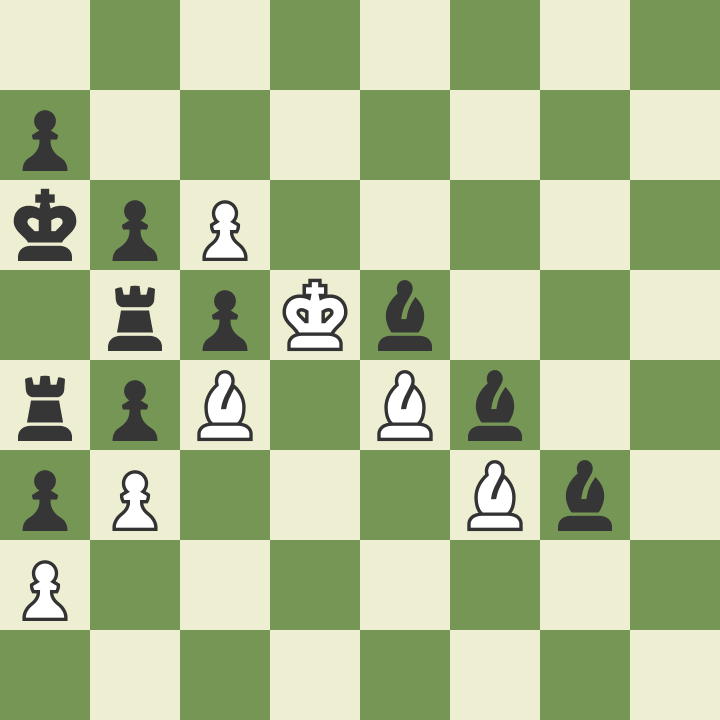}
    \end{minipage}
    & Draw & White &
    White and black get extra bishops compared to the original Penrose position, however the bishops are on opposite colors and the position is still a draw. SF correctly evaluates this position as a draw. 
    \\ \hline
    \begin{minipage}{0.3\textwidth}
    \captionsetup{justification=centering}
    \caption*{\tiny{8/p7/kpPb4/qrpP4/rpB2b2/pP4b1/P3K3/8 w - -}}
      \includegraphics[width=0.9\linewidth,]{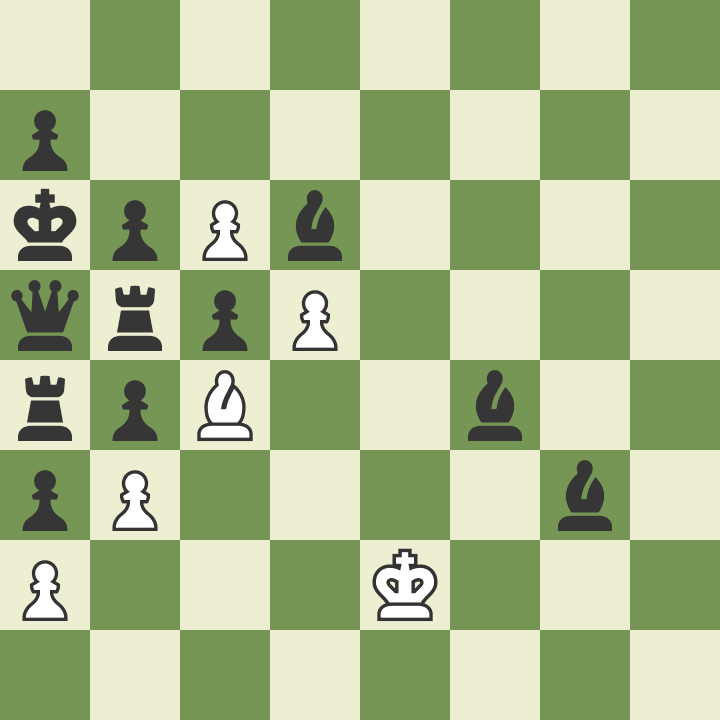}
    \end{minipage}
    & Draw & White &
    This position is a draw, however black needs to be accurate it their play. SF does not see the danger in the position, and plays incorrectly (resulting in a draw). We elaborate on this in the two positions below.
    \\ \hline
  \end{tabular}
  \captionsetup{justification=centering}
  \caption{Penrose set positions, Penrose variation 2, positions 9-12}\label{tbl:penrose1_6}
\end{table}\clearpage

\begin{table}[h!]
  \centering
  \begin{tabular}{ | c | m{1.5cm} | m{1.5cm} | m{6cm} |}
    \hline
    Board & Solution & To Play & Comment \\ \hline
    \begin{minipage}{0.3\textwidth}
    \captionsetup{justification=centering}
    \caption*{\tiny{2K5/p1b5/kpP5/qrpPb3/rpB5/pP2b3/P7/8 w - -}}
      \includegraphics[width=0.9\linewidth,]{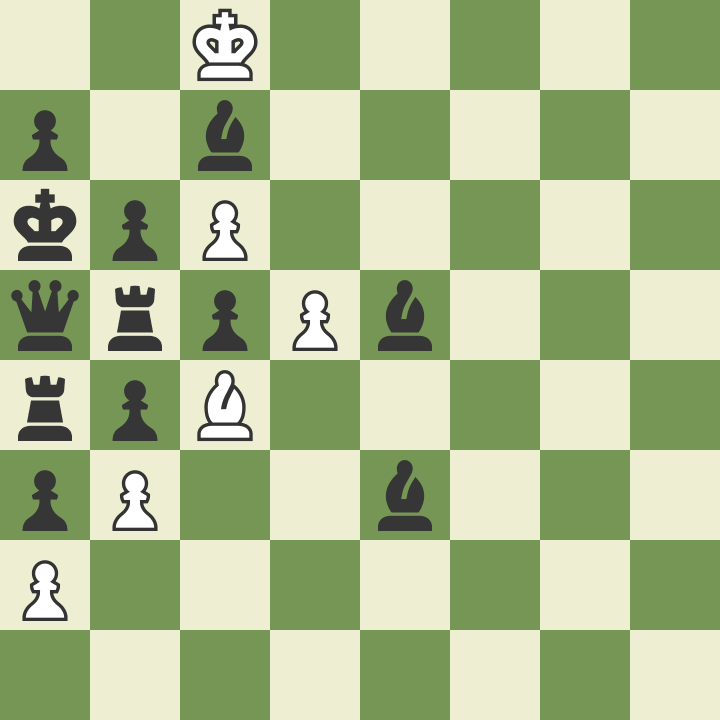}
    \end{minipage}
    & Win & White &
    Here, SF still thinks black is winning. However, white has mate in 5 with 1. d6.
    \\ \hline
    \begin{minipage}{0.3\textwidth}
    \captionsetup{justification=centering}
    \caption*{\tiny{2K5/p1b5/kpPP4/qrp1b3/rpB5/pP2b3/P7/8 b - -}}
      \includegraphics[width=0.9\linewidth,]{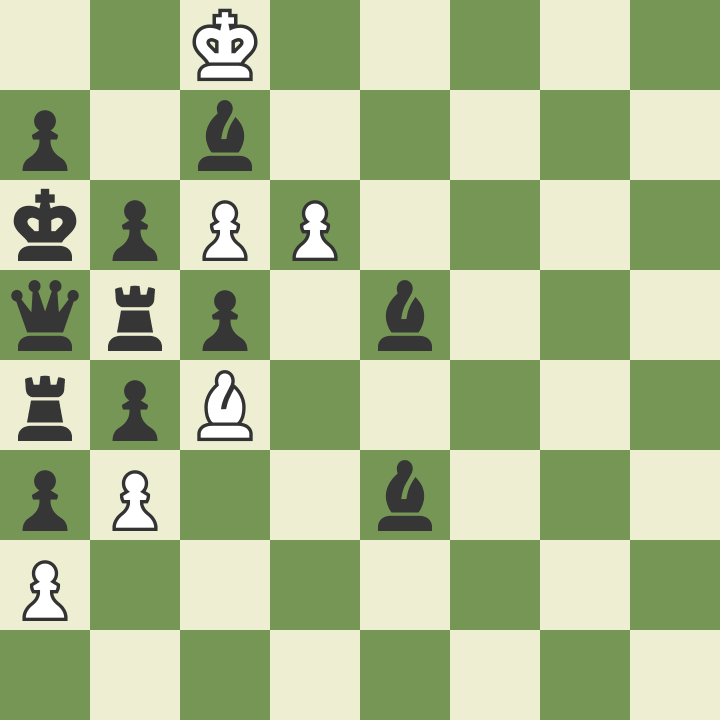}
    \end{minipage}
    & Lose & Black &
    Here, white is winning and SF’s evaluation jumps to winning for white. White can now use the e6 square for its bishop to mate black. 
    \\ \hline
    \begin{minipage}{0.3\textwidth}
    \captionsetup{justification=centering}
    \caption*{\tiny{8/p7/kpP5/rrp1b3/rpP2b2/pP4b1/P3K3/8 w - -}}
      \includegraphics[width=0.9\linewidth,]{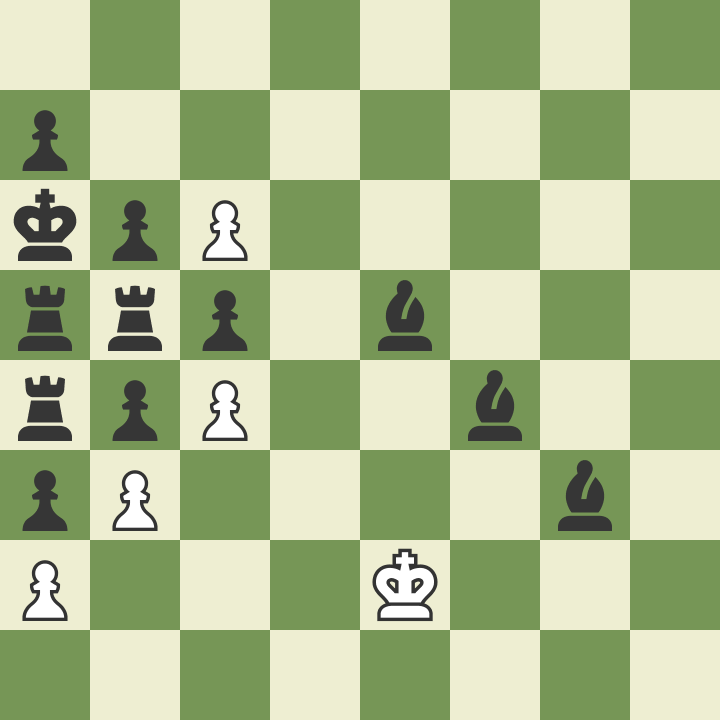}
    \end{minipage}
    & Draw & White &
    SF thinks black is winning. However having a rook, rather than a queen as in the original Penrose position, does not affect the evaluation.
    \\ \hline
    \begin{minipage}{0.3\textwidth}
    \captionsetup{justification=centering}
    \caption*{\tiny{8/p7/kpP5/rrp1b3/rpP2b2/pP6/P3K3/8 w - -}}
      \includegraphics[width=0.9\linewidth,]{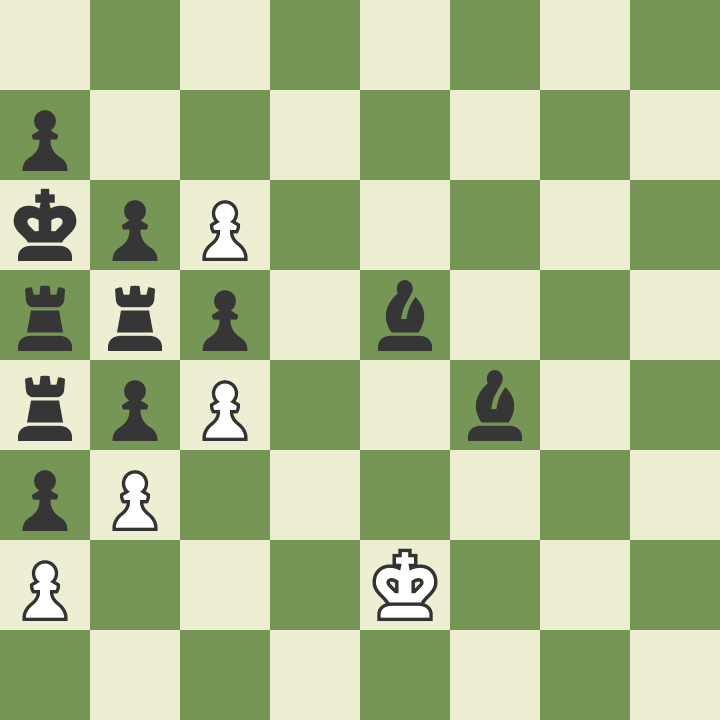}
    \end{minipage}
    & Draw & White &
    Similar to the position above, SF thinks black is winning. 
    \\ \hline
  \end{tabular}
  \captionsetup{justification=centering}
  \caption{Penrose set positions, Penrose variation 2, positions 13-16}\label{tbl:penrose1_7}
\end{table}\clearpage

\begin{table}[h!]
  \centering
  \begin{tabular}{ | c | m{1.5cm} | m{1.5cm} | m{6cm} |}
    \hline
    Board & Solution & To Play & Comment \\ \hline
    \begin{minipage}{0.3\textwidth}
    \captionsetup{justification=centering}
    \caption*{\tiny{8/p7/kpP5/rrp1b3/rpP5/pP6/P3K3/8 w - -}}
      \includegraphics[width=0.9\linewidth,]{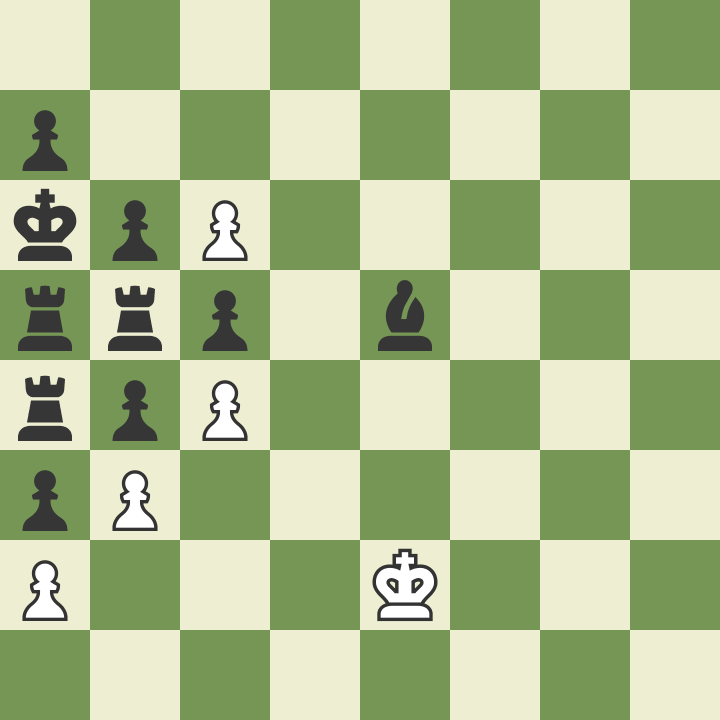}
    \end{minipage}
    & Draw & White &
    SF recognizes this position is a draw – the trend when comparing against the previous two positions is interesting because it suggests that there’s a limit to material advantage.
    \\ \hline
  \end{tabular}
  \captionsetup{justification=centering}
  \caption{Penrose set positions, Penrose variation 2, position 17}\label{tbl:penrose1_8}
\end{table}\clearpage

\begin{table}[h!]
  \centering
  \begin{tabular}{ | c | m{1.5cm} | m{1.5cm} | m{6cm} |}
    \hline
    Board & Solution & To Play & Comment \\ \hline
    \begin{minipage}{0.3\textwidth}
    \captionsetup{justification=centering}
    \caption*{\tiny{n1B4r/r1p2k2/BpPp3p/pP1PpBpP/P2bPpPb/ 1R1RbPb1/R1K1R1R1/1R3R2 w - -}}
      \includegraphics[width=0.9\linewidth,]{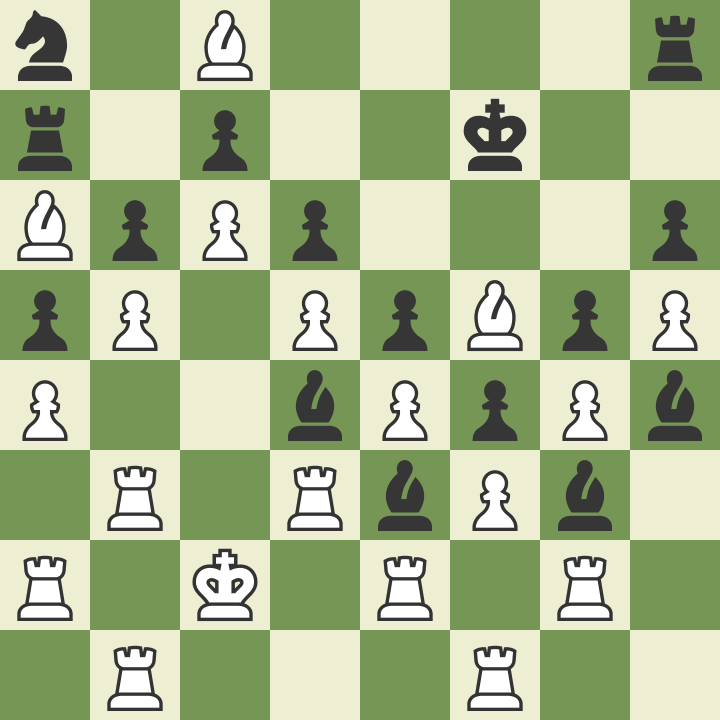}
    \end{minipage}
    & Draw & White &
    As the pawn structure is locked, and White is unable to remove any of the Black pawns by force, this is a trivial draw. Should White capture one of Black's dark-squared bishops, Black should avoid recapturing, and keep the blockade.
    \\ \hline
    \begin{minipage}{0.3\textwidth}
    \captionsetup{justification=centering}
    \caption*{\tiny{n1B4r/r1p2k2/BpPp3p/pP1PpBpP/P2bPpPb/ 1R1RbPb1/R1K1B1R1/1R3R2 w - -}}
      \includegraphics[width=0.9\linewidth,]{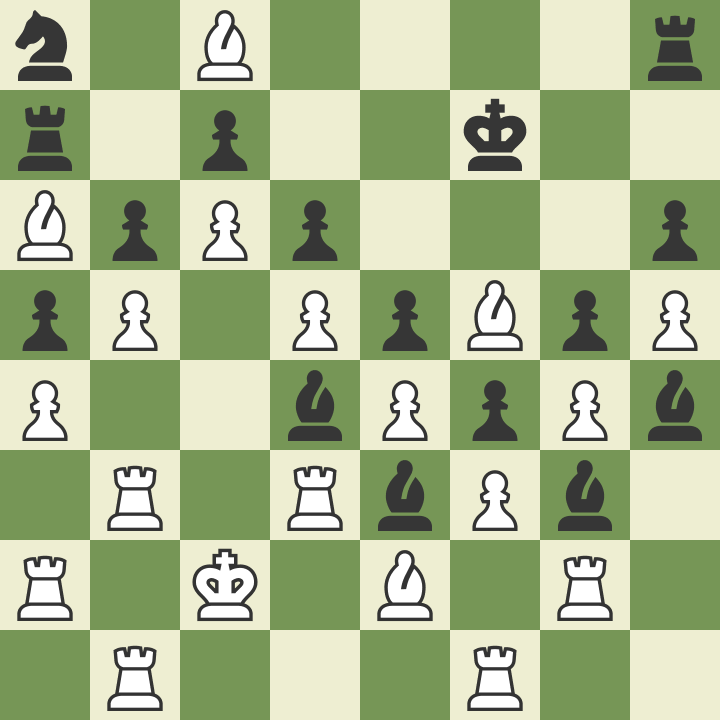}
    \end{minipage}
    & Draw & White &
    As the pawn structure is locked, and White is unable to remove any of the Black pawns by force, this is a trivial draw. Should White capture one of Black's dark-squared bishops, Black should avoid recapturing, and keep the blockade.
    \\ \hline
    \begin{minipage}{0.3\textwidth}
    \captionsetup{justification=centering}
    \caption*{\tiny{n1B4r/r1p2k2/BpPp3p/pP1PpBpP/P1RbPpPb/ 1R1RbPbB/R1K3R1/1R3R1R w - -}}
      \includegraphics[width=0.9\linewidth,]{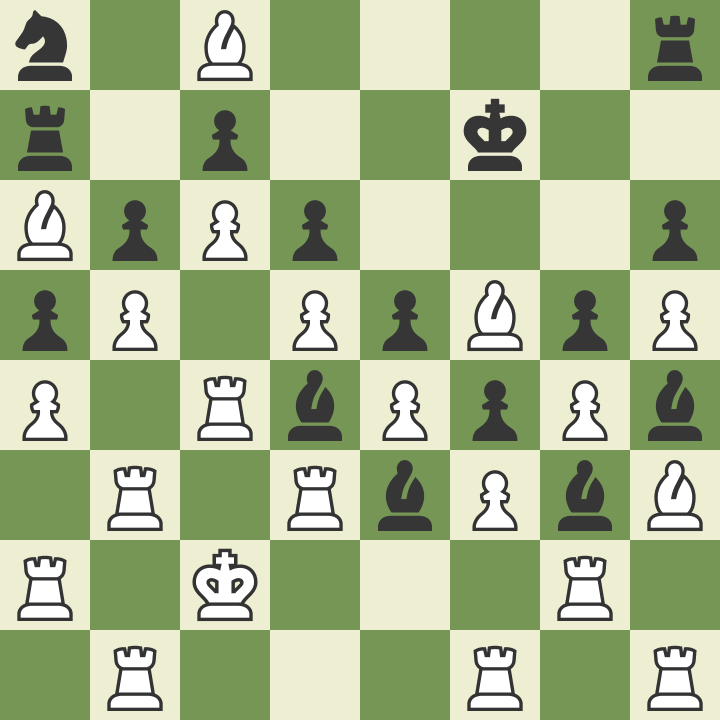}
    \end{minipage}
    & Draw & White &
    As the pawn structure is locked, and White is unable to remove any of the Black pawns by force, this is a trivial draw. Should White capture one of Black's dark-squared bishops, Black should avoid recapturing, and keep the blockade.
    \\ \hline
  \end{tabular}
  \captionsetup{justification=centering}
  \caption{Penrose set positions, Blocked set variation 1, positions 1-3. These represent the same theme (blockade), albeit with a slight variation in the total amount of material on each side, and therefore -- the branching factor in search.}\label{tbl:blocked1}
\end{table}\clearpage

\begin{table}[h!]
  \centering
  \begin{tabular}{ | c | m{1.5cm} | m{1.5cm} | m{6cm} |}
    \hline
    Board & Solution & To Play & Comment \\ \hline
    \begin{minipage}{0.3\textwidth}
    \captionsetup{justification=centering}
    \caption*{\tiny{2rBb3/1kBbP3/1BbP3B/BbP1B3/bP5B/ P1B1B3/5K1B/B1B1B3 w - -}}
      \includegraphics[width=0.9\linewidth,]{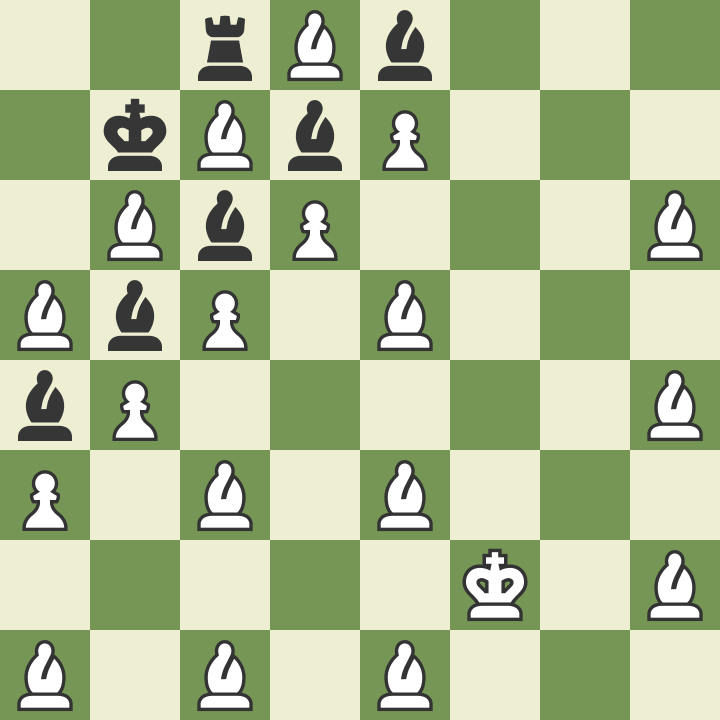}
    \end{minipage}
    & Draw & White &
    Despite the overwhelming material advantage, White is unable to penetrate on the light squares controlled by Black light-squared bishops. The position is also out of distribution in that it would not normally be possible to have as many bishops on the board.
    \\ \hline
    \begin{minipage}{0.3\textwidth}
    \captionsetup{justification=centering}
    \caption*{\tiny{3Bb2R/1kBbP3/1BbP3B/BbP1B3/bP5B/ P1B1B3/5K1B/2B1B3 w - -}}
      \includegraphics[width=0.9\linewidth,]{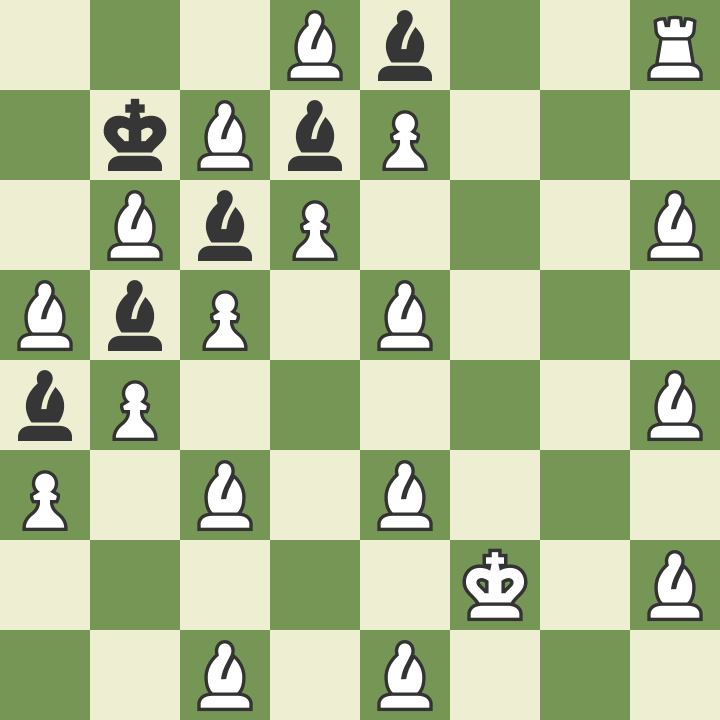}
    \end{minipage}
    & Draw & White &
    Despite the overwhelming material advantage, White is unable to penetrate on the light squares controlled by Black light-squared bishops. The position is also out of distribution in that it would not normally be possible to have as many bishops on the board. The addition of the rook makes it possible to exchange one bishop off, but doesn't change the evaluation.
    \\ \hline
    \begin{minipage}{0.3\textwidth}
    \captionsetup{justification=centering}
    \caption*{\tiny{3Bb2R/1kBbP3/nBbP3B/BbP1B3/bP5B/ P1B1B3/5K1B/2B1B3 w - -}}
      \includegraphics[width=0.9\linewidth,]{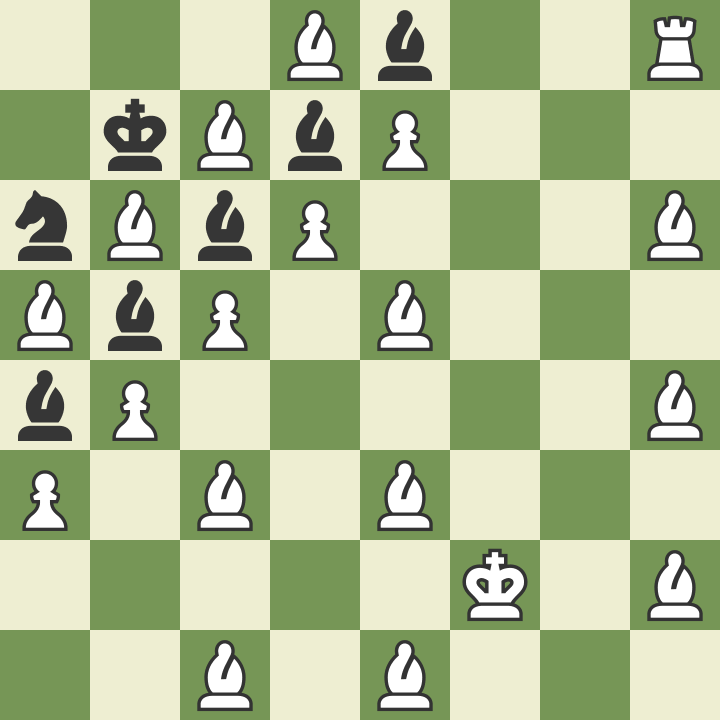}
    \end{minipage}
    & Draw & White &
    Despite the overwhelming material advantage, White is unable to penetrate on the light squares controlled by Black light-squared bishops. The position is also out of distribution in that it would not normally be possible to have as many bishops on the board. A further addition of the Knight on a6 should not affect the evaluation.
    \\ \hline
  \end{tabular}
  \captionsetup{justification=centering}
  \caption{Penrose set positions, Blocked set variation 2, positions 1-3}\label{tbl:blocked2}
\end{table}\clearpage

\begin{table}[h!]
  \centering
  \begin{tabular}{ | c | m{1.5cm} | m{1.5cm} | m{6cm} |}
    \hline
    Board & Solution & To Play & Comment \\ \hline
    \begin{minipage}{0.3\textwidth}
    \captionsetup{justification=centering}
    \caption*{\tiny{5k2/5Bp1/5pP1/1K2pP2/4P3/1B3B2/8/5B2 w - -}}
      \includegraphics[width=0.9\linewidth,]{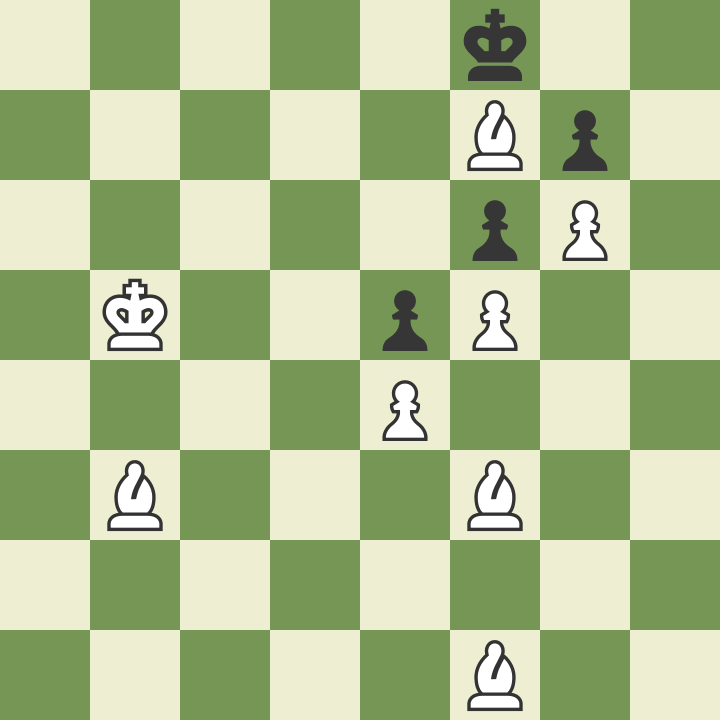}
    \end{minipage}
    & Draw & White &
    White has no way of challenging the dark squares and the pawn on g7. Black should just move the king Kf8-e7-f8, and the game would end in a draw. If White vacates the a2-g8 diagonal, the Black king could be forced to g8 or h8, but that wouldn't change the outcome. Different amounts of extra material test for whether having extra light-squared bishops affects the engine's judgement.
    \\ \hline
    \begin{minipage}{0.3\textwidth}
    \captionsetup{justification=centering}
    \caption*{\tiny{5k2/5Bp1/5pP1/1K2pP2/4P3/8/8/8 w - -}}
      \includegraphics[width=0.9\linewidth,]{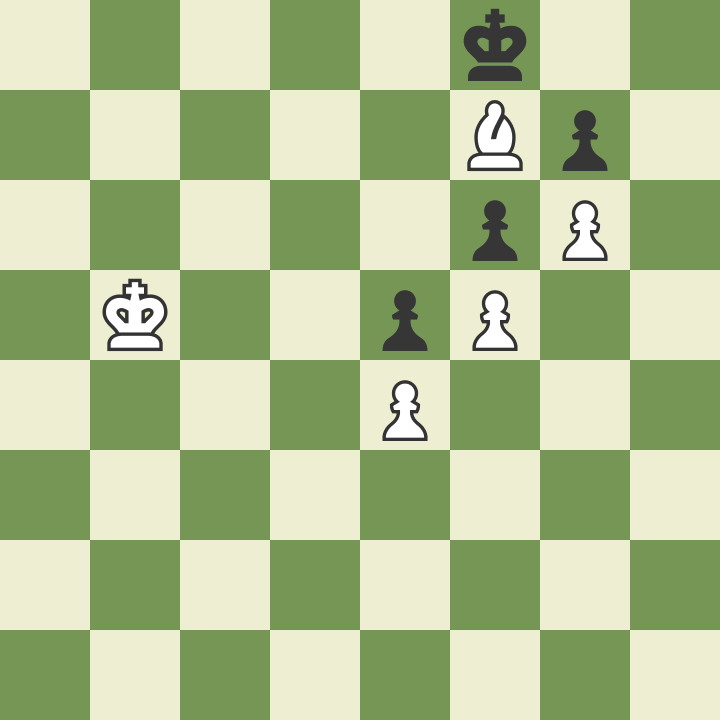}
    \end{minipage}
    & Draw & White &
    White has no way of challenging the dark squares and the pawn on g7. Black should just move the king Kf8-e7-f8, and the game would end in a draw. If White vacates the a2-g8 diagonal, the Black king could be forced to g8 or h8, but that wouldn't change the outcome. Different amounts of extra material test for whether having extra light-squared bishops affects the engine's judgement.
    \\ \hline
    \begin{minipage}{0.3\textwidth}
    \captionsetup{justification=centering}
    \caption*{\tiny{B4k2/5BpB/2B2pP1/1K2pP1B/4P3/3B1B1B/B7/8 w - -}}
      \includegraphics[width=0.9\linewidth,]{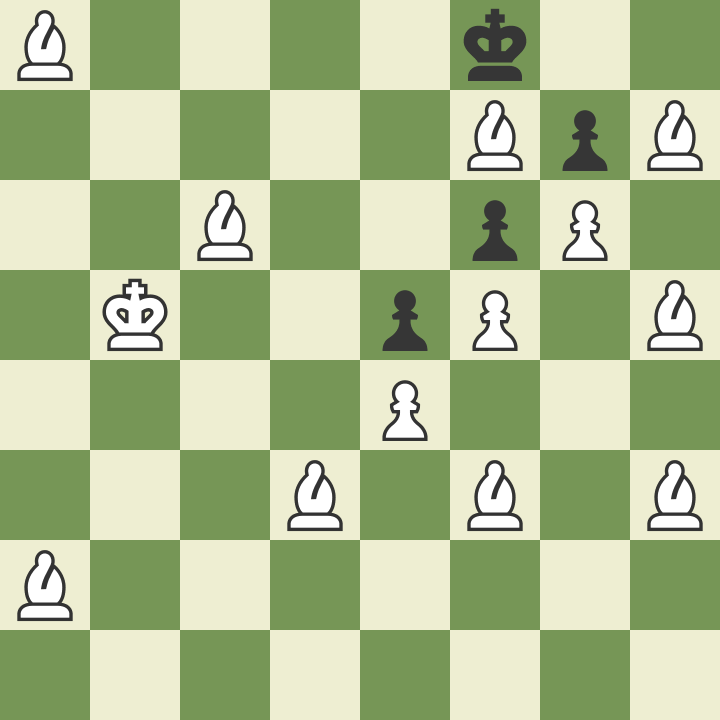}
    \end{minipage}
    & Draw & White &
    White has no way of challenging the dark squares and the pawn on g7. Black should just move the king Kf8-e7-f8, and the game would end in a draw. If White vacates the a2-g8 diagonal, the Black king could be forced to g8 or h8, but that wouldn't change the outcome. Different amounts of extra material test for whether having extra light-squared bishops affects the engine's judgement.
    \\ \hline
  \end{tabular}
  \captionsetup{justification=centering}
  \caption{Penrose set positions, Blocked set variation 3, positions 1-3}\label{tbl:blocked3}
\end{table}\clearpage

\begin{table}[h!]
  \centering
  \begin{tabular}{ | c | m{1.5cm} | m{1.5cm} | m{6cm} |}
    \hline
    Board & Solution & To Play & Comment \\ \hline
    \begin{minipage}{0.3\textwidth}
    \captionsetup{justification=centering}
    \caption*{\tiny{rb1b3k/2b1p3/1p1pP3/pPpP1B2/P1P5/4K3/8/8 w - -}}
      \includegraphics[width=0.9\linewidth,]{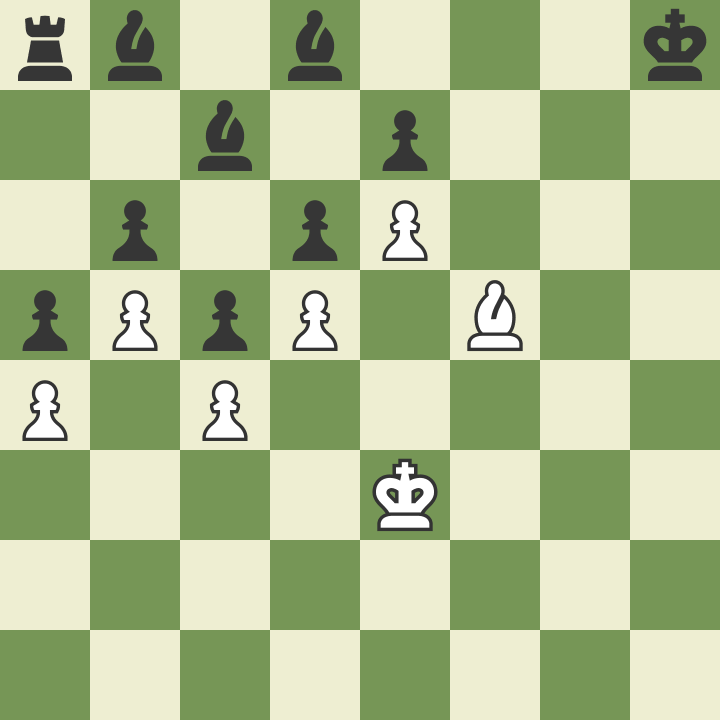}
    \end{minipage}
    & Draw & White &
    By getting the king to g5, White blocks the activation of the Black king, which cannot make further progress. Thanks to the light-squared bishop, which will penetrate the light squares on the kingside, Black will not be able to safely get the rook out, and the dark-squared bishops cannot escape. Sample line to illustrate the point: 1. Kf4 Ba7 2. Bg6 Rb8 3. Be8 Rb7 4. Bc6
    \\ \hline
  \end{tabular}
  \captionsetup{justification=centering}
  \caption{Penrose set positions, Blocked set variation 4, position 1}\label{tbl:blocked4}
\end{table}\clearpage

\begin{table}[h!]
  \centering
  \begin{tabular}{ | c | m{1.5cm} | m{1.5cm} | m{6cm} |}
    \hline
    Board & Solution & To Play & Comment \\ \hline
    \begin{minipage}{0.3\textwidth}
    \captionsetup{justification=centering}
    \caption*{\tiny{2b1rr2/1p1p1r1k/1PpPp2r/2P1Pp1r/5Pp1/3K2Pp/7P/8 w - -}}
      \includegraphics[width=0.9\linewidth,]{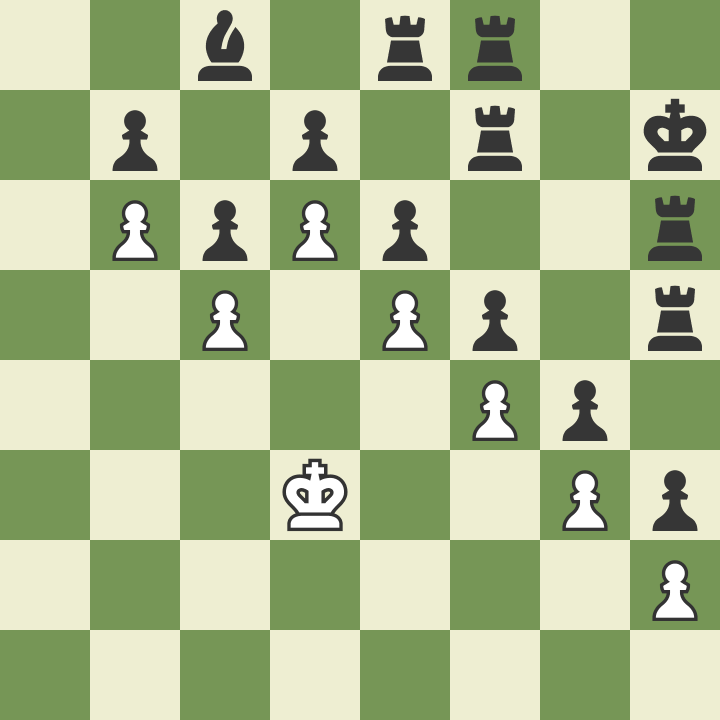}
    \end{minipage}
    & Draw & White &
    Despite the extra material, the bishop on c8 makes it impossible for Black to utilize any of the extra pieces, resulting in a fully blocked position, and therefore a trivial draw.
    \\ \hline
    \begin{minipage}{0.3\textwidth}
    \captionsetup{justification=centering}
    \caption*{\tiny{2b1rr2/1p1p1r1k/1PpPp2r/2P1Pp1r/1R3Pp1/ 2RK2Pp/5B1P/2B5 w - -}}
      \includegraphics[width=0.9\linewidth,]{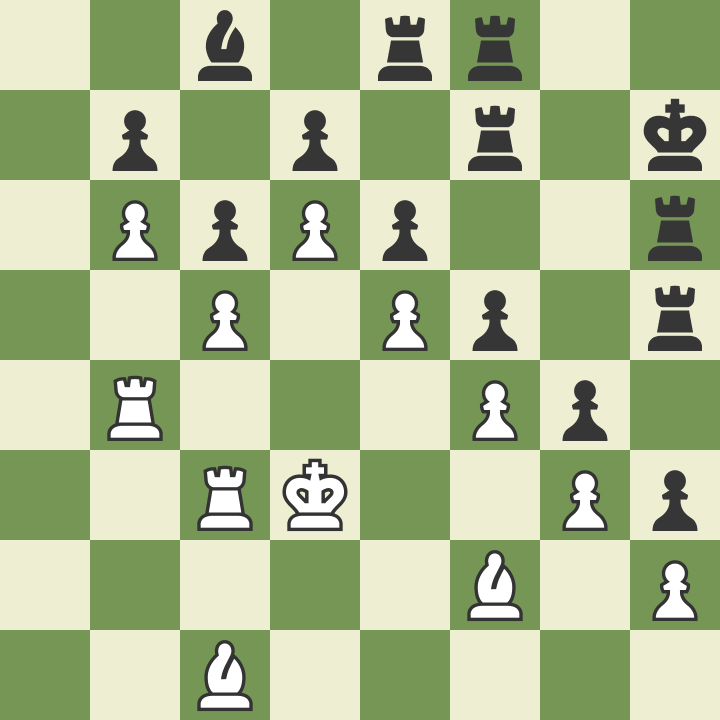}
    \end{minipage}
    & Draw & White &
    Same as the previous position, with some extra pieces for White. Black has sufficient extra material to defend against any potential exchange on b7, which is the only additional dynamic option introduced by the White rooks. Therefore, the evaluation is still a trivial draw, as neither side can make progress.
    \\ \hline
    \begin{minipage}{0.3\textwidth}
    \captionsetup{justification=centering}
    \caption*{\tiny{2b1rr2/1p1p1r1k/1PpPp2r/2P1Pp1r/1R3Pp1/ 3K2Pp/5B1P/8 w - -}}
      \includegraphics[width=0.9\linewidth,]{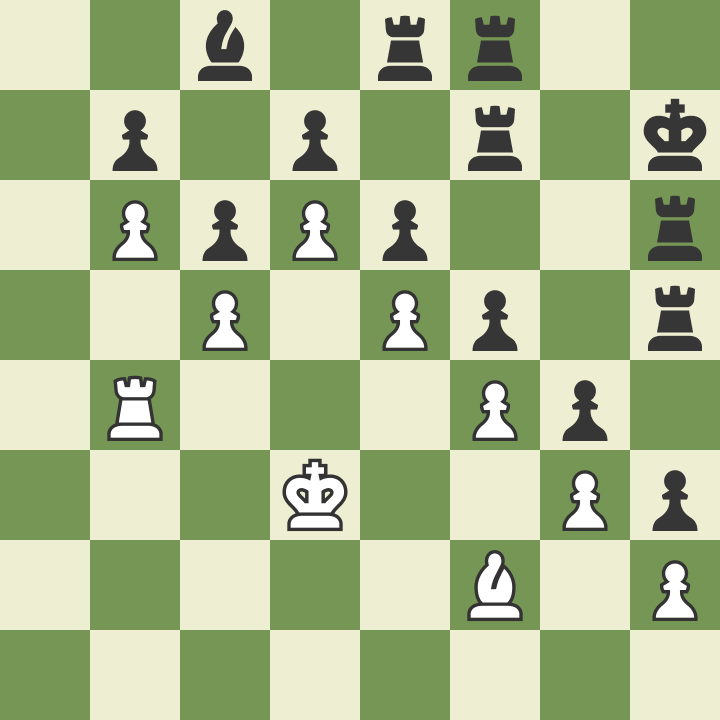}
    \end{minipage}
    & Draw & White &
    As above, though with less material for White, to better evaluate how the extent of material differences may affect engine evaluation.
    \\ \hline
  \end{tabular}
  \captionsetup{justification=centering}
  \caption{Penrose set positions, Blocked set variation 5, positions 1-3}\label{tbl:blocked5_1}
\end{table}\clearpage

\begin{table}[h!]
  \centering
  \begin{tabular}{ | c | m{1.5cm} | m{1.5cm} | m{6cm} |}
    \hline
    Board & Solution & To Play & Comment \\ \hline
    \begin{minipage}{0.3\textwidth}
    \captionsetup{justification=centering}
    \caption*{\tiny{2b2r2/1p1p3k/1PpPp3/2P1Pp1r/5Pp1/3K2Pp/7P/8 w - -}}
      \includegraphics[width=0.9\linewidth,]{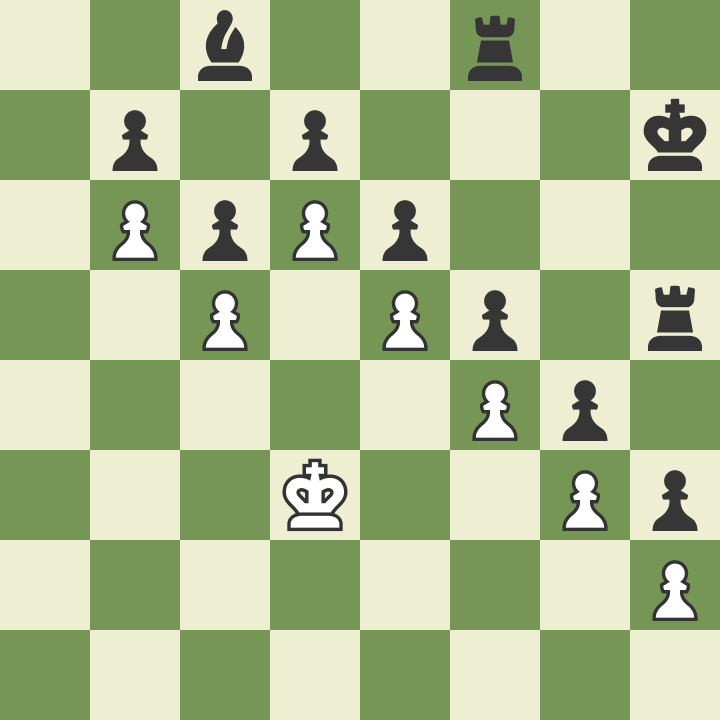}
    \end{minipage}
    & Draw & White &
    The bishop on c8 makes it impossible to utilize Black rooks, resulting in a trivial draw.
    \\ \hline
    \begin{minipage}{0.3\textwidth}
    \captionsetup{justification=centering}
    \caption*{\tiny{2b2r2/1p1p3k/1PpPp3/2P1Pp2/5Pp1/3K2Pp/7P/8 w - -}}
      \includegraphics[width=0.9\linewidth,]{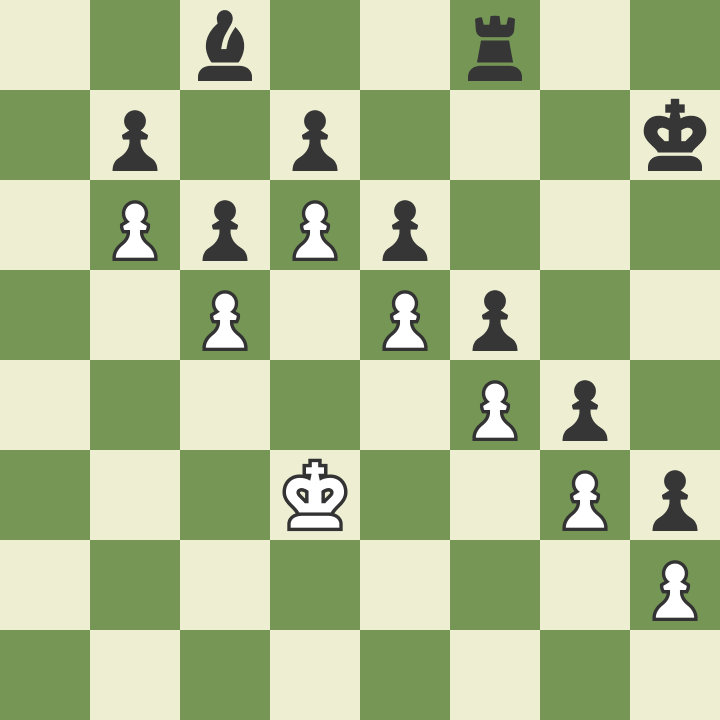}
    \end{minipage}
    & Draw & White &
    The bishop on c8 makes it impossible to utilize Black rook, resulting in a trivial draw.
    \\ \hline
  \end{tabular}
  \captionsetup{justification=centering}
  \caption{Penrose set positions, Blocked set variation 5, positions 4-5}\label{tbl:blocked5_2}
\end{table}\clearpage

\begin{table}[h!]
  \centering
  \begin{tabular}{ | c | m{1.5cm} | m{1.5cm} | m{6cm} |}
    \hline
    Board & Solution & To Play & Comment \\ \hline
    \begin{minipage}{0.3\textwidth}
    \captionsetup{justification=centering}
    \caption*{\tiny{6b1/2k3P1/4b1PQ/3b2P1/b1b3PQ/ 1b2p1PQ/b3P1P1/4KBRR w - -}}
      \includegraphics[width=0.9\linewidth,]{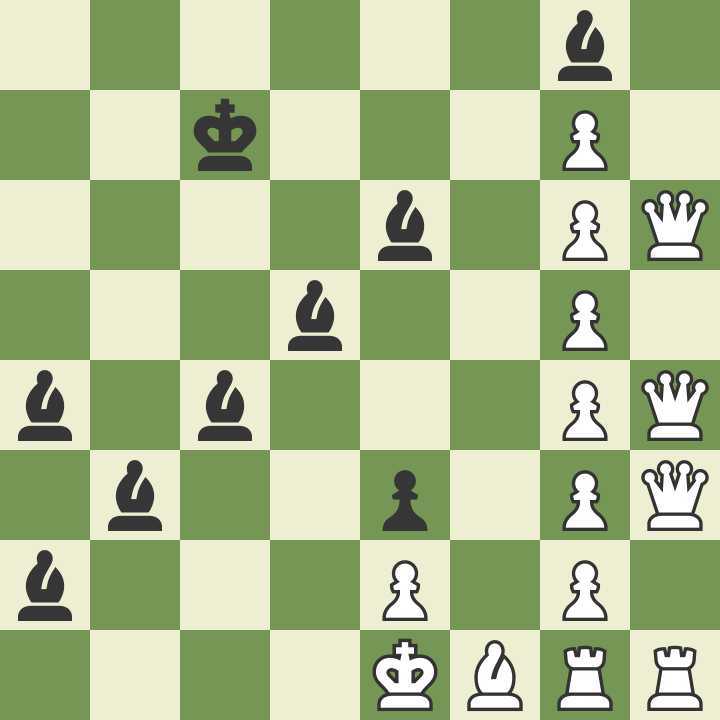}
    \end{minipage}
    & Draw & White &
    White is unable to utilize the overwhelming material advantage. White can exchange some extra material on the g8 square, though counting the pieces should make it obvious that at the end of all those exchanges, Black will remain with a bishop on g8. The bishop on a4 also makes it impossible to activate White's king.
    \\ \hline
    \begin{minipage}{0.3\textwidth}
    \captionsetup{justification=centering}
    \caption*{\tiny{6bQ/1b1b2P1/4b1PQ/3b2P1/2b3PQ/ 1b2p1PQ/b3P1P1/2k1KBRR w - -}}
      \includegraphics[width=0.9\linewidth,]{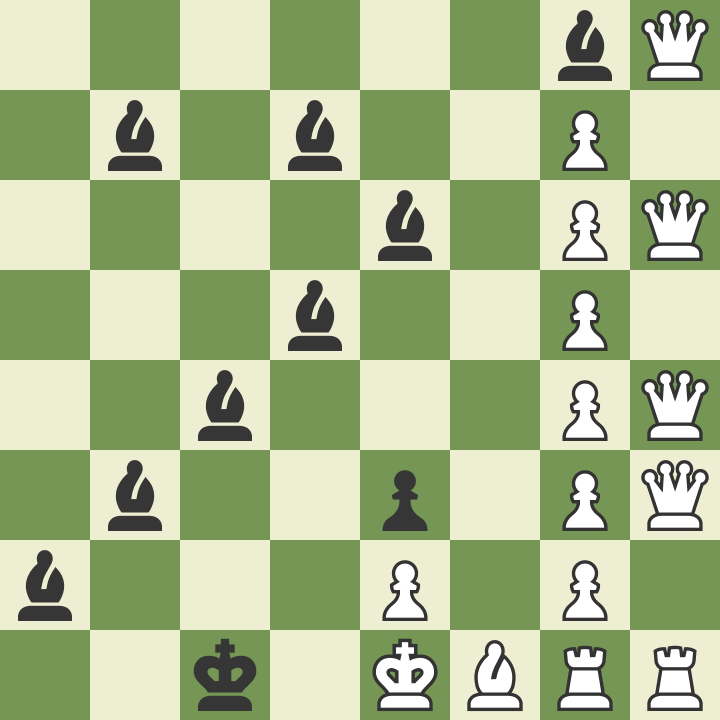}
    \end{minipage}
    & Draw & White &
    As above, piece counting with respect to possible captures on g8 makes it clear that White's pieces on the h-file will not be able to activate. Compared to the previous position, there is a larger material advantage here, but the evaluation is the same.
    \\ \hline
    \begin{minipage}{0.3\textwidth}
    \captionsetup{justification=centering}
    \caption*{\tiny{6bQ/1b1b2PR/2b1b1PQ/3b2P1/2b3PQ/ 1b2p1PQ/b3P1P1/2k1KBRR w - -}}
      \includegraphics[width=0.9\linewidth,]{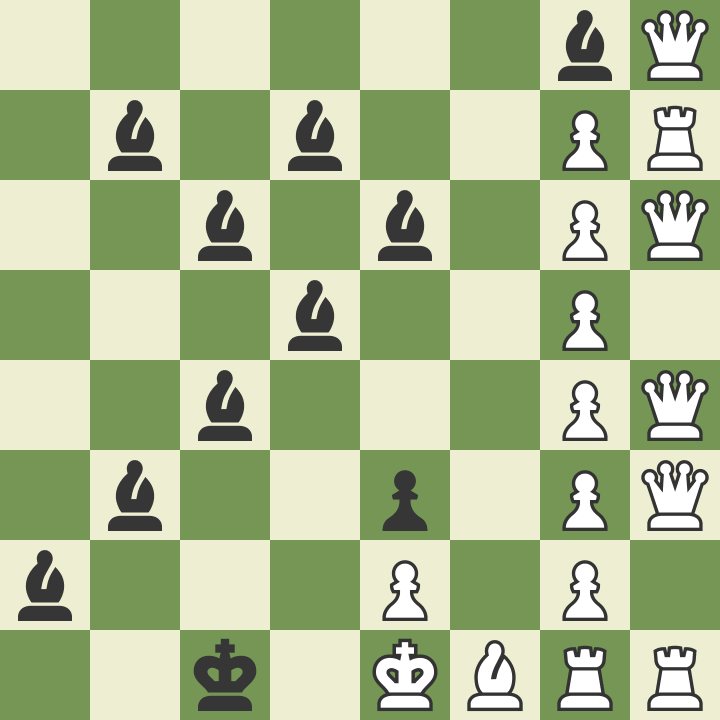}
    \end{minipage}
    & Draw & White &
    As above, piece counting with respect to possible captures on g8 makes it clear that White's pieces on the h-file will not be able to activate. Compared to the previous position, there is a larger material advantage here, but the evaluation is the same.
    \\ \hline
  \end{tabular}
  \captionsetup{justification=centering}
  \caption{Penrose set positions, Blocked set variation 6, positions 1-3}\label{tbl:blocked6}
\end{table}\clearpage

\begin{table}[h!]
  \centering
  \begin{tabular}{ | c | m{1.5cm} | m{1.5cm} | m{6cm} |}
    \hline
    Board & Solution & To Play & Comment \\ \hline
    \begin{minipage}{0.3\textwidth}
    \captionsetup{justification=centering}
    \caption*{\tiny{2k5/2p5/1q1p4/pPpPp1pp/N1P1Pp2/P4PbP/KQ4P1/8 w - -}}
      \includegraphics[width=0.9\linewidth,]{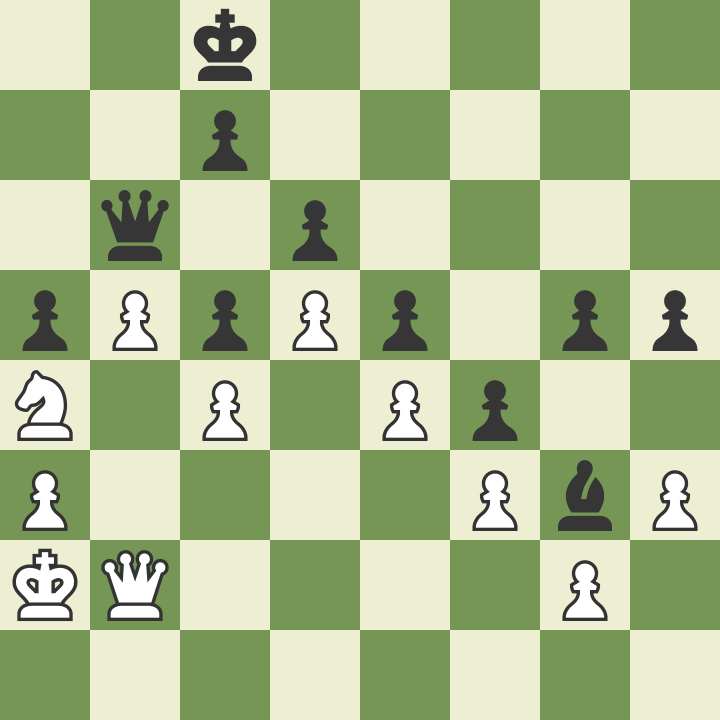}
    \end{minipage}
    & Win & White &
    A. Petrosian-Hazai, 1970. Black is strategically lost, and tries his last chance by offering a queen sacrifices. The trap worked, but white should have instead fought over the a5 pawn and win. 
    \\ \hline
    \begin{minipage}{0.3\textwidth}
    \captionsetup{justification=centering}
    \caption*{\tiny{2k5/8/1p1p4/pPpPp1pp/2P1Pp2/P4PbP/KQ4P1/8 w - -}}
      \includegraphics[width=0.9\linewidth,]{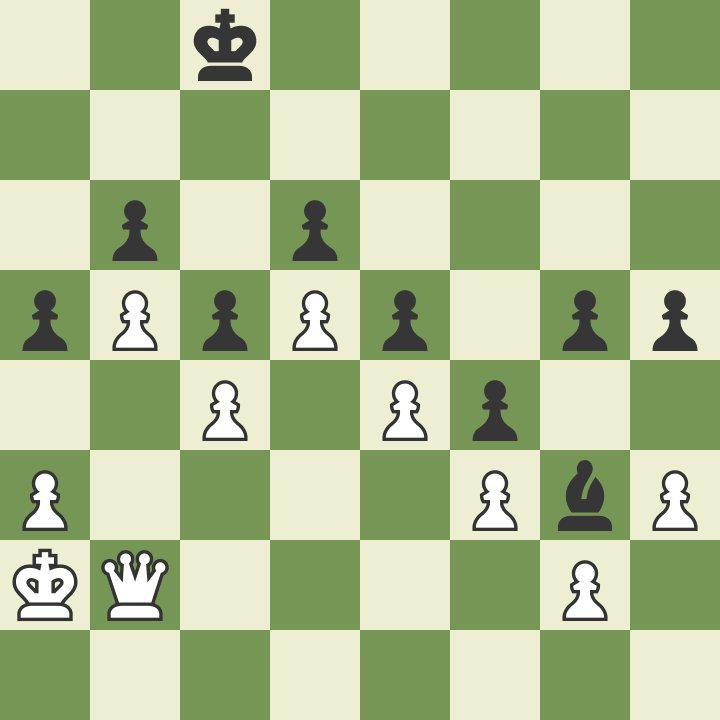}
    \end{minipage}
    & Draw & White &
    Black threatens to seal the kingside with ...h4 but White still has a window of opportunity to open entry lines into the black position: 1.h4 gxh4 when the h3-square is freed for the white queen. However after 2.Qb1 h3!! 3.gxh3 h4, the door is sealed shut once more!   
    \\ \hline
    \begin{minipage}{0.3\textwidth}
    \captionsetup{justification=centering}
    \caption*{\tiny{7r/8/8/p3p1k1/Pp1pPp2/1PpP1Pp1/2P1K1P1/8 b - -}}
      \includegraphics[width=0.9\linewidth,]{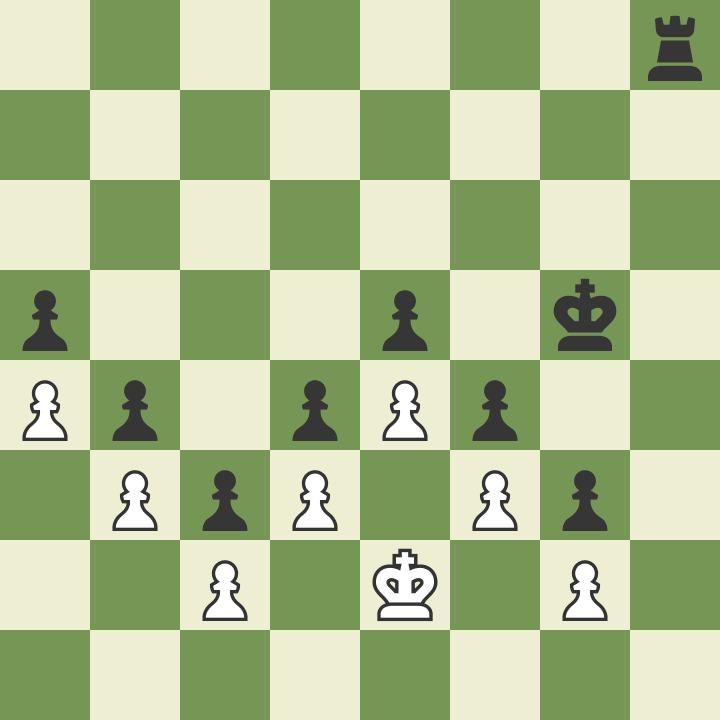}
    \end{minipage}
    & Draw & Black &
    Black's problem is that the rook can only invade by entering White's first rank (e.g. via 1...Rh1) but that also turns out to be stalemate! 
    \\ \hline
  \end{tabular}
  \captionsetup{justification=centering}
  \caption{Penrose set positions from \citep{gidel2018frank}, positions 1-3}\label{tbl:fortress1}
\end{table}\clearpage

\begin{table}[h!]
  \centering
  \begin{tabular}{ | c | m{1.5cm} | m{1.5cm} | m{6cm} |}
    \hline
    Board & Solution & To Play & Comment \\ \hline
    \begin{minipage}{0.3\textwidth}
    \captionsetup{justification=centering}
    \caption*{\tiny{8/8/6k1/8/4N2p/7P/3N2K1/q7 b - -}}
      \includegraphics[width=0.9\linewidth,]{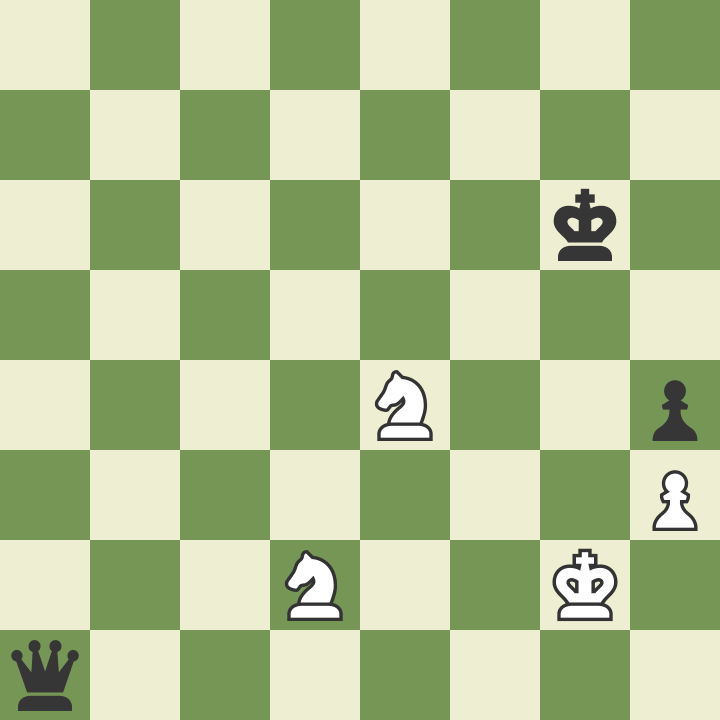}
    \end{minipage}
    & Draw & Black &
    P. Pechenkin, 1953. The black king cannot go further than e2 and f4, while the queen cannot come close enough to the white king to deny him of the free squares in the corner. 
    \\ \hline
    \begin{minipage}{0.3\textwidth}
    \captionsetup{justification=centering}
    \caption*{\tiny{8/B1p5/2Pp4/3Pp1k1/4P3/5PK1/3q4/8 b - -}}
      \includegraphics[width=0.9\linewidth,]{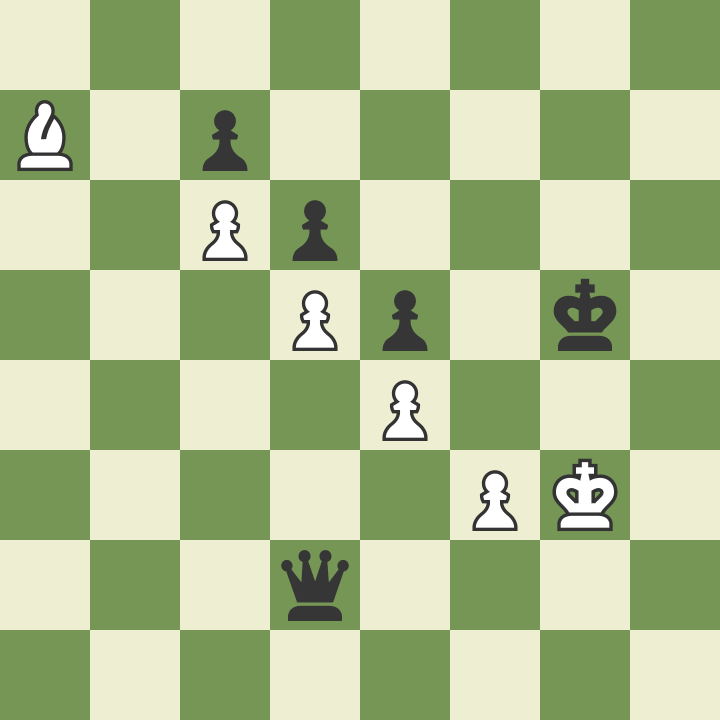}
    \end{minipage}
    & Draw & Black &
    Chekhover, 1948. Black king breakthrough on the king side is difficult, and it cannot march to the queen side since the white bishop controls the a7 square. 
    \\ \hline
    \begin{minipage}{0.3\textwidth}
    \captionsetup{justification=centering}
    \caption*{\tiny{4K3/5p1N/2k2PpB/6P1/8/8/b7/7q b - -}}
      \includegraphics[width=0.9\linewidth,]{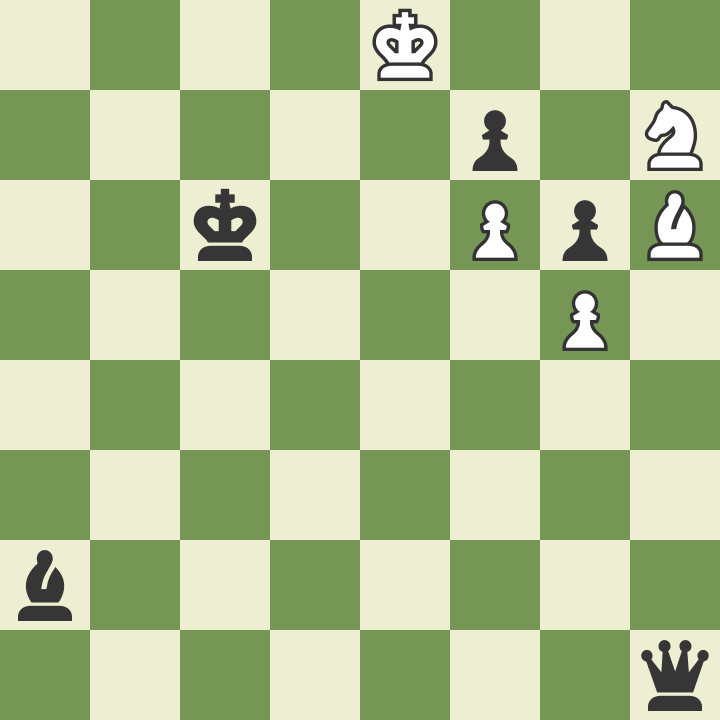}
    \end{minipage}
    & Draw & Black &
    Nothing can stop the white king from heading to the fortified security of the g7 square!
    \\ \hline
  \end{tabular}
  \captionsetup{justification=centering}
  \caption{Penrose set positions from \citep{gidel2018frank} (top and middle) and \citep{steingrimsson2021chess} (bottom), positions 4-6}\label{tbl:fortress2}
\end{table}\clearpage

\begin{table}[h!]
  \centering
  \begin{tabular}{ | c | m{1.5cm} | m{1.5cm} | m{6cm} |}
    \hline
    Board & Solution & Comment \\ \hline
    \begin{minipage}{0.3\textwidth}
    \captionsetup{justification=centering}
    \caption*{\tiny{1B3B1B/2B5/p6B/8/8/8/8/1k1K4 w - -}}
      \includegraphics[width=0.9\linewidth,]{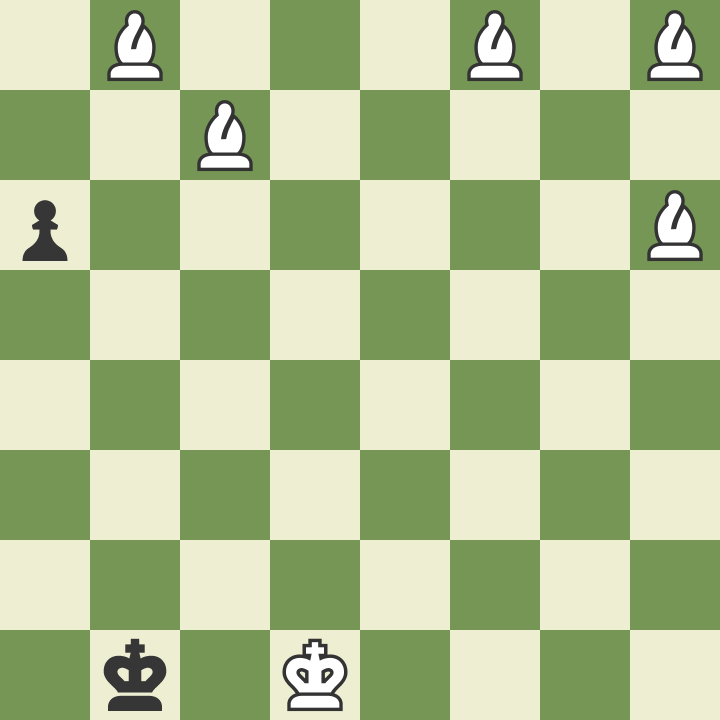}
    \end{minipage}
    & Win & White &
    Troitsky, 1916. To win, white must force the black king to the a1 square, the black pawn to a2, the white king has to control the b1 square (via c1 or c2), and have a bishop checking the king on the long diagonal. This can be achieved via a sequence of maneuvers, zugzwangs (that force the black pawn to march to a2) and a discovered check. Otherwise, white cannot mate black with the same color bishops. Engines miss this long mating sequence. Source: \citep{Troitsky}.

    \\ \hline
    \begin{minipage}{0.3\textwidth}
    \captionsetup{justification=centering}
    \caption*{\tiny{r7/7k/5R2/p3p3/Pp1pPp2/ 1PpP1Pp1/K1P3P1/8 w - -}}
      \includegraphics[width=0.9\linewidth,]{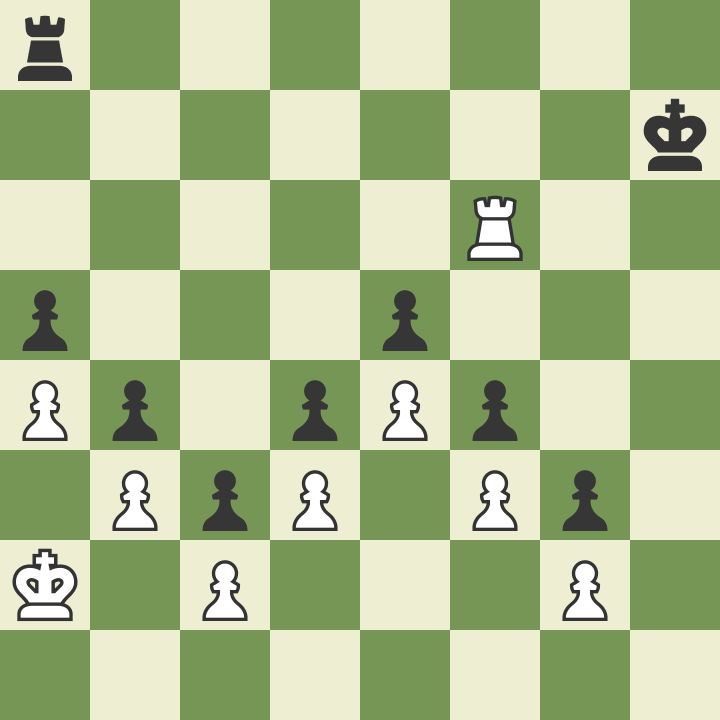}
    \end{minipage}
    & Draw & White &
    Hasek vs. Unknown player. White sacrifices its rook and builds a fortress. Source: \citep{10positions}.
    \\ \hline
    \begin{minipage}{0.3\textwidth}
    \captionsetup{justification=centering}
    \caption*{\tiny{8/1p1q1k2/1Pp5/p1Pp4/P2Pp1p1/ 4PpPp/1N3P1P/3B2K1 w - -}}
      \includegraphics[width=0.9\linewidth,]{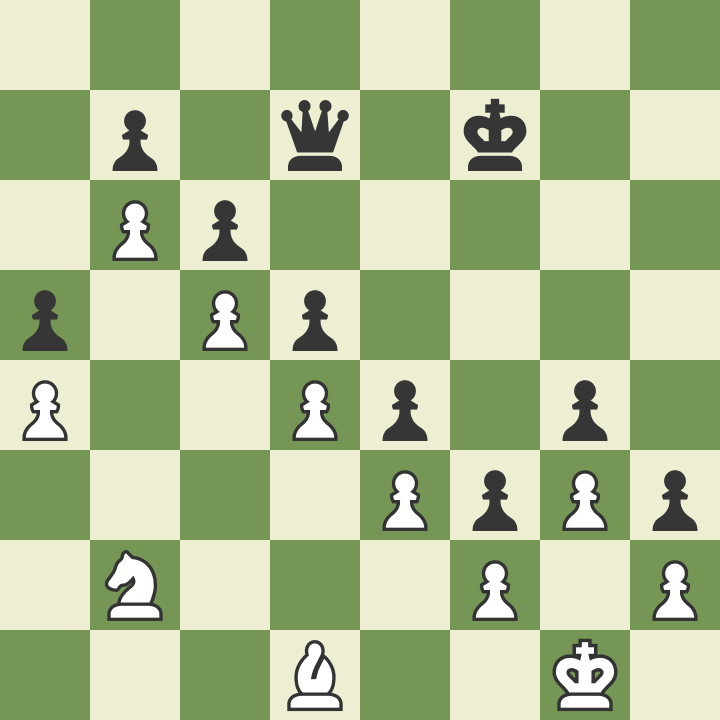}
    \end{minipage}
    & Draw & White &
    One of those puzzles where the solution is still difficult to reconstruct even if you've solved it before! White has a choice of pieces to sacrifices and where to sacrifice them to create a fortress and the solution is the least obvious of all!  Source: \citep{10positions}.
    \\ \hline
  \end{tabular}
  \captionsetup{justification=centering}
  \caption{Challenge set positions, positions 1-3}\label{tbl:Challenge1}
\end{table}\clearpage

\begin{table}[h!]
  \centering
  \begin{tabular}{ | c | m{1.5cm} | m{1.5cm} | m{6cm} |}
    \hline
    Board & Solution & Comment \\ \hline
    \begin{minipage}{0.3\textwidth}
    \captionsetup{justification=centering}
    \caption*{\tiny{8/3P3k/n2K3p/2p3n1/1b4N1/2p1p1P1/8/3B4 w - - }}
      \includegraphics[width=0.9\linewidth,]{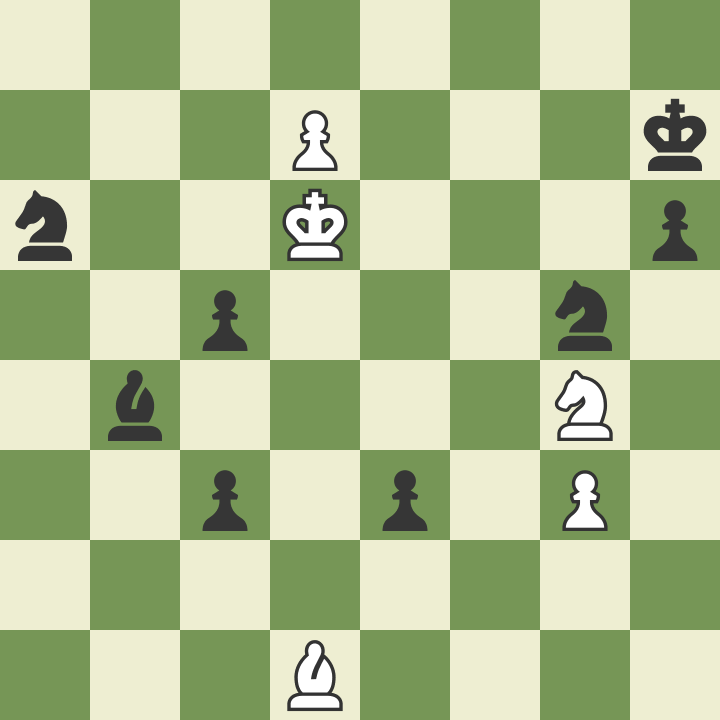}
    \end{minipage}
    & Win & White &
    \href{https://en.chessbase.com/post/solution-to-a-truly-remarkable-study}{Solution to a truly remarkable study}, \href{https://www.youtube.com/watch?v=8wCJalNkTEI}{Grandmasters and Engines Couldn't Solve This Puzzle. Then Came The Magician}.
    \\ \hline
    \begin{minipage}{0.3\textwidth}
    \captionsetup{justification=centering}
    \caption*{\tiny{4b1k1/P4pPp/1R3P1P/2r5/8/1P6/1K6/8 w - -}}
      \includegraphics[width=0.9\linewidth,]{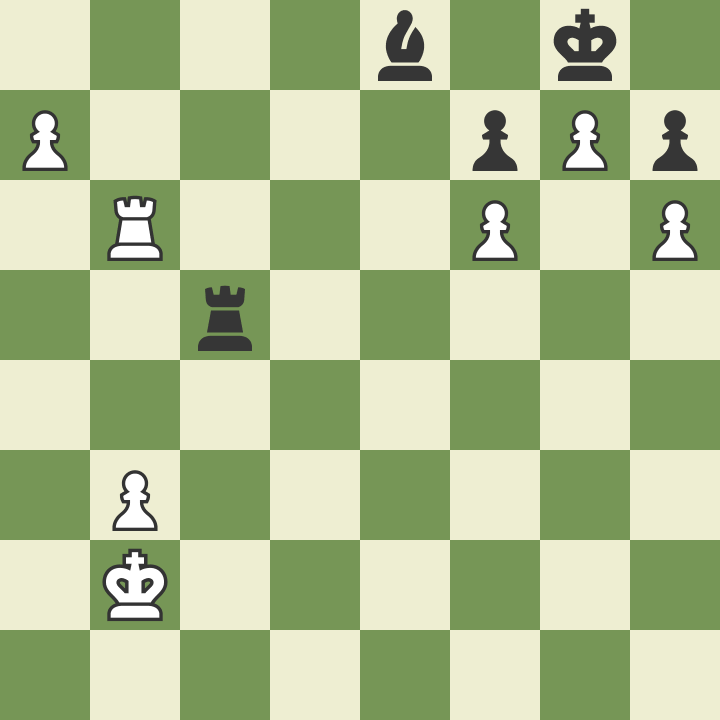}
    \end{minipage}
    & Win & White &
    Sokolsky vs. Ravinsky, USSR 1938. Source: Tim Krabbe's blog, \href{https://timkr.home.xs4all.nl/chess2/minor.htm}{Underpromotion In Games}. White promotes thr a7 pawn to a bishop on its way to a win. Other promotions end up with a stalemate. 
    \\ \hline
    \begin{minipage}{0.3\textwidth}
    \captionsetup{justification=centering}
    \caption*{\tiny{n2Bqk2/5p1p/Q4KP1/p7/8/8/8/8 w - -}}
      \includegraphics[width=0.9\linewidth,]{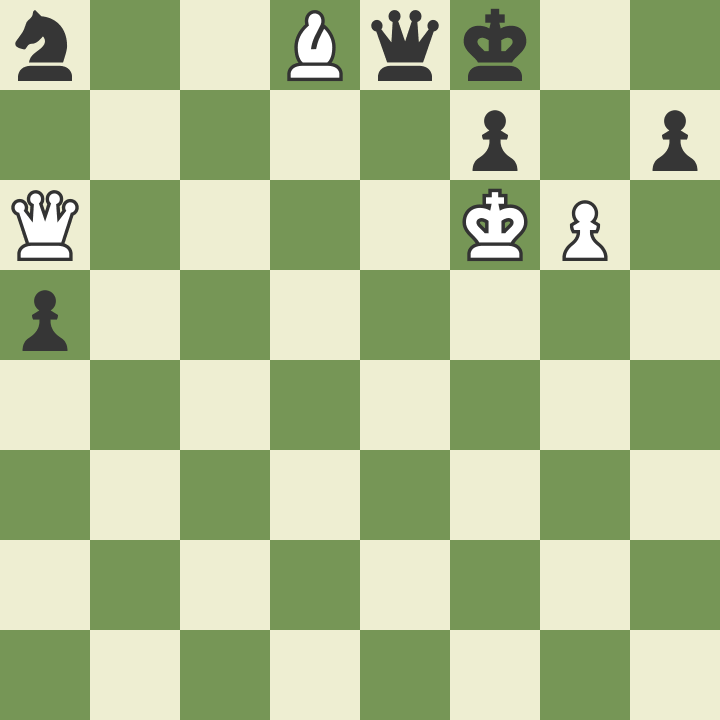}
    \end{minipage}
    & Win & White &
    Most chess engines struggle with this beautiful mating pattern in which Black has many delaying moves to calculate, but a human can clearly see they will not help \citep{10positions}.
    \\ \hline
  \end{tabular}
  \captionsetup{justification=centering}
  \caption{Challenge set positions, positions 4-6}\label{tbl:Challenge2}
\end{table}\clearpage

\begin{table}[h!]
  \centering
  \begin{tabular}{ | c | m{1.5cm} | m{1.5cm} | m{6cm} |}
    \hline
    Board & Solution & Comment \\ \hline
    \begin{minipage}{0.3\textwidth}
    \captionsetup{justification=centering}
    \caption*{\tiny{3B4/1r2p3/r2p1p2/bkp1P1p1/1p1P1PPp/ p1P4P/PPB1K3/8 w - -}}
      \includegraphics[width=0.9\linewidth,]{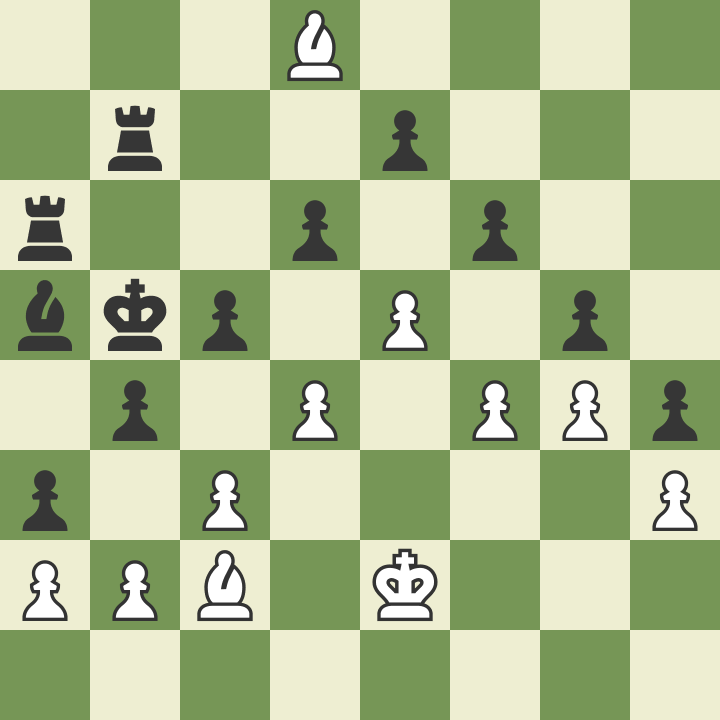}
    \end{minipage}
    & Draw & White &
    William Rudolph vs. Unknown player. White sacrifices the bishop, marches its pawns while delivering checks, and closes the position. 
    \\ \hline
    \begin{minipage}{0.3\textwidth}
    \captionsetup{justification=centering}
    \caption*{\tiny{4K3/4Bp1N/2k3p1/5PP1/8/7p/b7/8 w - -}}
      \includegraphics[width=0.9\linewidth,]{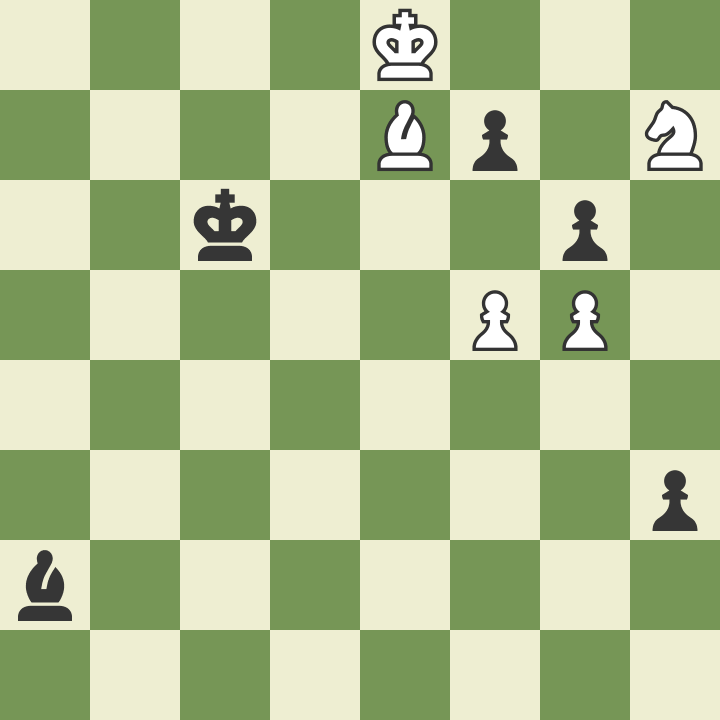}
    \end{minipage}
    & Draw & White &
    It's clear that White's cannot stop the h-pawn or create a passed pawn of his own. The dilemma is to find the most solid setup on the kingside to be able to hide the white king on g7 and keep the g5-pawn safe. 
    \\ \hline
    \begin{minipage}{0.3\textwidth}
    \captionsetup{justification=centering}
    \caption*{\tiny{8/6B1/5N2/3p1k2/1bb3p1/6Pp/5K1P/8 w - - }}
      \includegraphics[width=0.9\linewidth,]{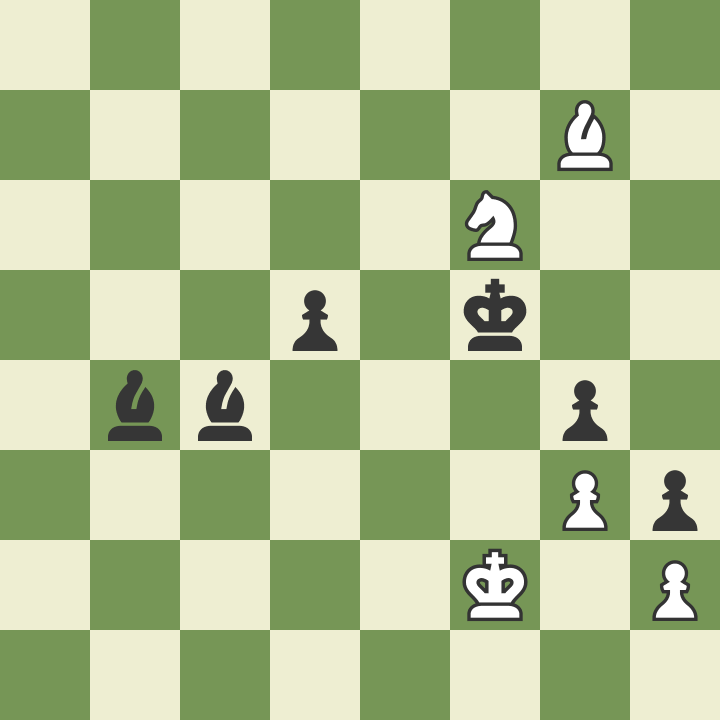}
    \end{minipage}
    & Draw & White &
    The resulting fortress position is not unfamiliar to human players but the added material makes it strangely difficult to visualise. It seems particularly strange that 2 mighty black bishops cannot drive the white king away from the kingside pawns. 
    \\ \hline
  \end{tabular}
  \captionsetup{justification=centering}
  \caption{Challenge set positions from \citep{steingrimsson2021chess}, positions 7-9}\label{tbl:Challenge3}
\end{table}\clearpage

\begin{table}[h!]
  \centering
  \begin{tabular}{ | c | m{1.5cm} | m{1.5cm} | m{6cm} |}
    \hline
    Board & Solution & Comment \\ \hline
    \begin{minipage}{0.3\textwidth}
    \captionsetup{justification=centering}
    \caption*{\tiny{7r/p3k3/2p5/1pPp4/3P4/PP4P1/3P1PB1/2K5 w - -}}
      \includegraphics[width=0.9\linewidth,]{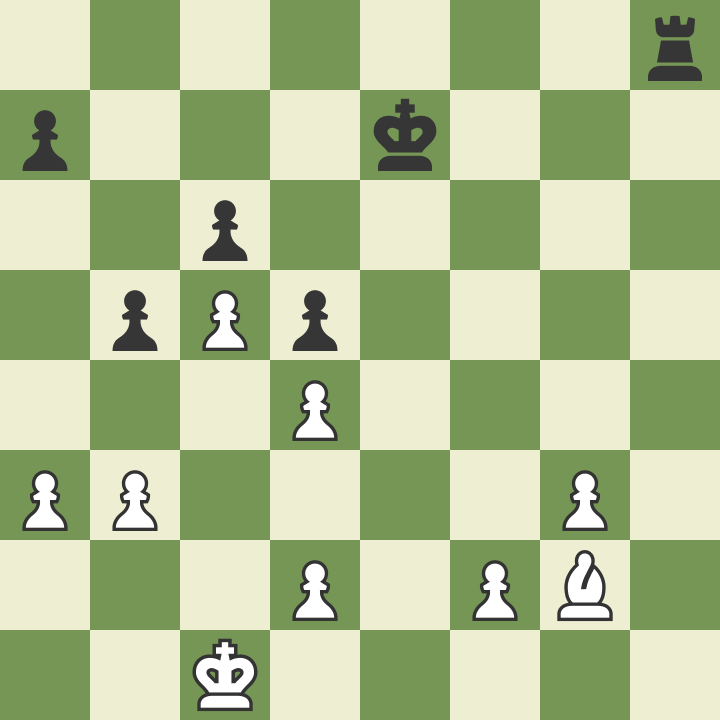}
    \end{minipage}
    & Draw & White &
    Chekhover, 1947. The Black Rook can infiltrate White’s position and start capturing pawns. Yet White has a fortress like resource in his dispose. 
    \\ \hline
    \begin{minipage}{0.3\textwidth}
    \captionsetup{justification=centering}
    \caption*{\tiny{4knQ1/7r/3p2p1/2bP1pP1/5P1N/6K1/8/8 b - -}}
      \includegraphics[width=0.9\linewidth,]{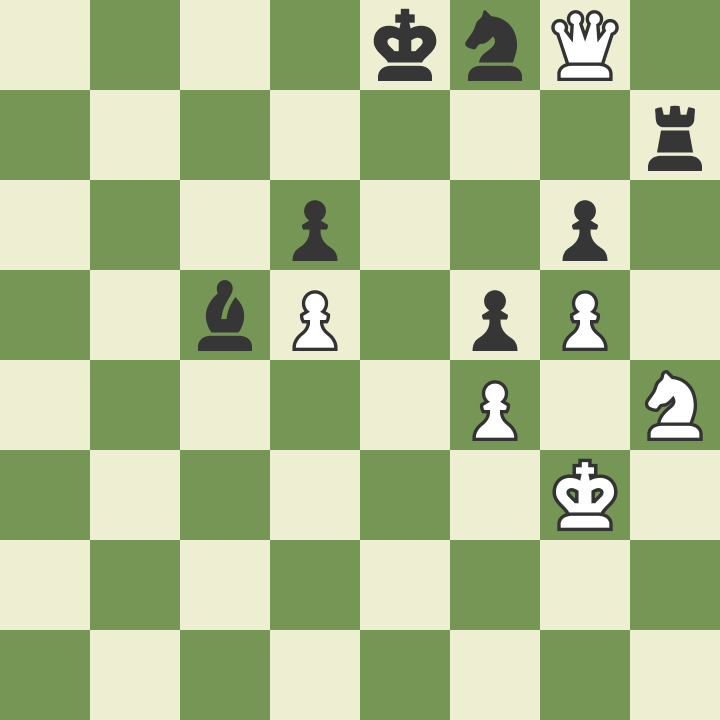}
    \end{minipage}
    & Draw & Black &
    White's queen is exerting a lot of pressure on Black's position but 1...Rxh4 2.Kxh4 Bd4! denies the queen access to any squares! Without any queen mobility, White is powerless to break the black fortress. 
    \\ \hline
    \begin{minipage}{0.3\textwidth}
    \captionsetup{justification=centering}
    \caption*{\tiny{7k/4p3/5p2/1p1p1p2/1PpP1PpB/1pP3P1/7P/6NK w - -}}
      \includegraphics[width=0.9\linewidth,]{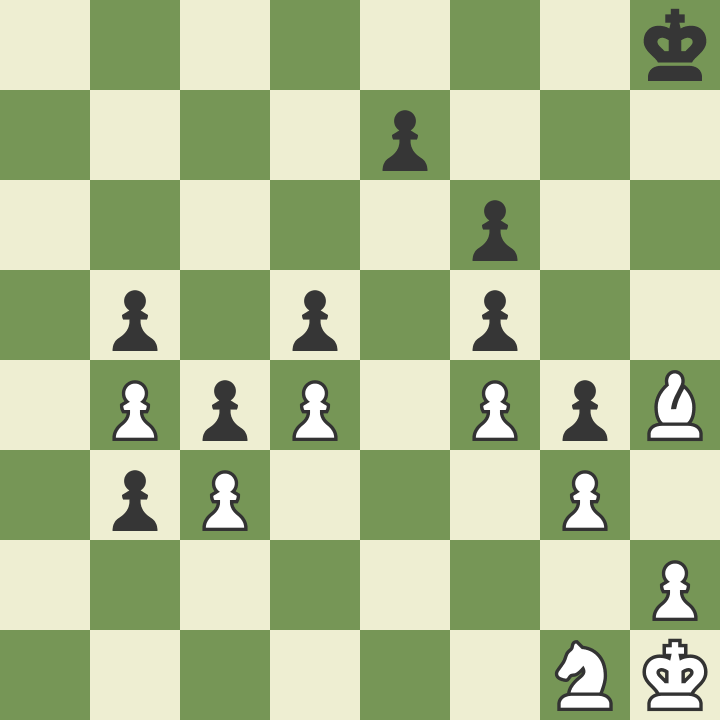}
    \end{minipage}
    & Draw & White &
    White's line to draw 1.Bxf6+ ef 2.h4 b2 3.Kg2 b1(Q) 4.Kf2 followed by Ne2.
    \\ \hline
  \end{tabular}
  \captionsetup{justification=centering}
  \caption{Challenge set positions from \citep{steingrimsson2021chess}, positions 10-12}\label{tbl:Challenge4}
\end{table}\clearpage

\begin{table}[h!]
  \centering
  \begin{tabular}{ | c | m{1.5cm} | m{1.5cm} | m{6cm} |}
    \hline
    Board & Solution & Comment \\ \hline
    \begin{minipage}{0.3\textwidth}
    \captionsetup{justification=centering}
    \caption*{\tiny{8/8/4kpp1/3p1b2/p6P/2B5/6P1/6K1 b - -}}
      \includegraphics[width=0.9\linewidth,]{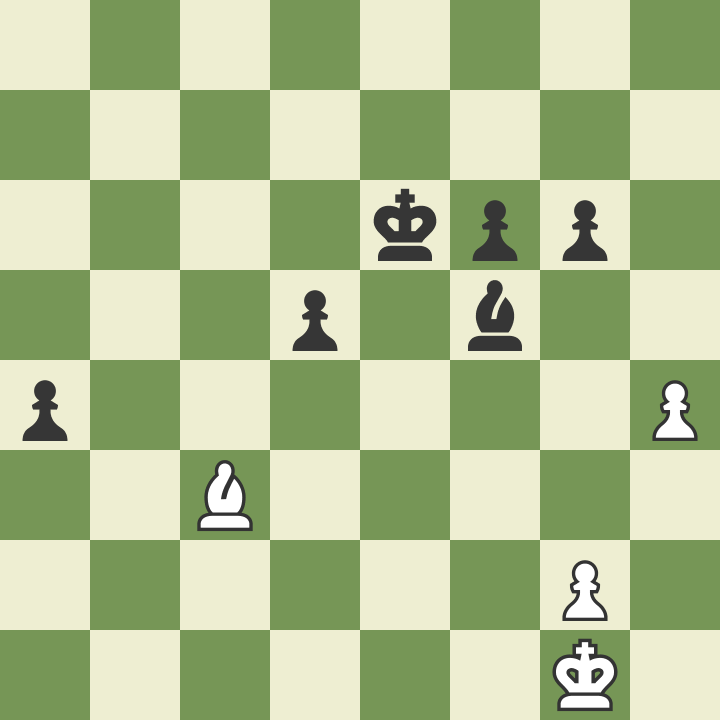}
    \end{minipage}
    & Win & Black &
    Topalov-Shirov, Linares 1998 (Round 10). Here, the best move for black is to continue with Bh3. However, the uncoventional sacrifice is difficult for engines to see, and therefore they miss the win. 
    \\ \hline
    \begin{minipage}{0.3\textwidth}
    \captionsetup{justification=centering}
    \caption*{\tiny{8/8/2p5/2k5/p1p1B3/PpP5/rP6/2NK4 w - -}}
      \includegraphics[width=0.9\linewidth,]{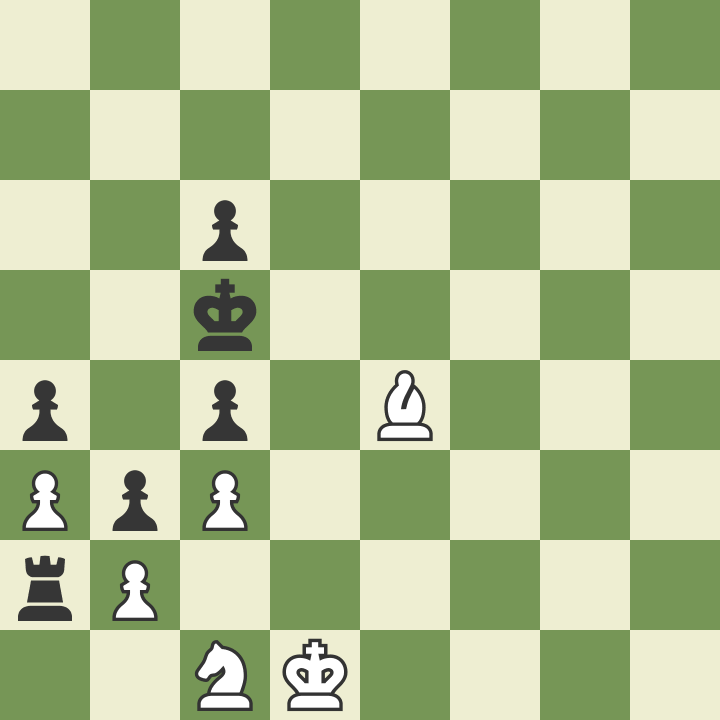}
    \end{minipage}
    & Draw & White &
    Prophylaxis, the prevention of an opponent's ideas, is one of the most difficult things in chess. It's no surprise then that engines may also struggle with this concept. Prophylaxis is naturally most difficult when the opponent's ideas are well-hidden. In the following position, Black has a dangerous idea to play ...Ra2 and win the b2-pawn. White has one clever defense that requires anticipating the strength of ...Ra2 \citep{10positions}. 
    \\ \hline
    \begin{minipage}{0.3\textwidth}
    \captionsetup{justification=centering}
    \caption*{\tiny{2k5/2p5/1q1p4/pPpPp1pp/N1P1Pp2/P4PbP/KQ4P1/8 w - -}}
      \includegraphics[width=0.9\linewidth,]{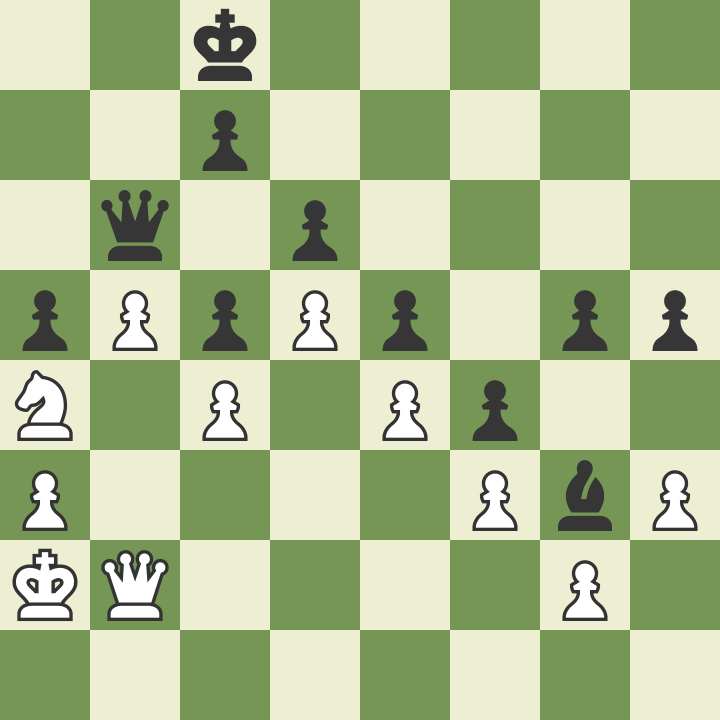}
    \end{minipage}
    & Win & White &
    Arshak B Petrosian vs. Laszlo Hazai. GM Laszlo Hazai offered a queen sacrifice, and GM Arshak Petrosian took the bait. Many chess engines fall for the same trap, while a careful play for white leads to a win \citep{10positions}. 
    \\ \hline
  \end{tabular}
  \captionsetup{justification=centering}
  \caption{Challenge set positions, positions 13-15}\label{tbl:Challenge5}
\end{table}
\clearpage

\bibliography{main}

\begin{thebibliography}{83}
\providecommand{\natexlab}[1]{#1}
\providecommand{\url}[1]{\texttt{#1}}
\expandafter\ifx\csname urlstyle\endcsname\relax
  \providecommand{\doi}[1]{doi: #1}\else
  \providecommand{\doi}{doi: \begingroup \urlstyle{rm}\Url}\fi

\bibitem[Anschel et~al.(2017)Anschel, Baram, and Shimkin]{anschel2017averaged}
O.~Anschel, N.~Baram, and N.~Shimkin.
\newblock Averaged-dqn: Variance reduction and stabilization for deep
  reinforcement learning.
\newblock \emph{International conference on machine learning}, pages 176--185,
  2017.

\bibitem[Balduzzi et~al.(2019)Balduzzi, Garnelo, Bachrach, Czarnecki, Perolat,
  Jaderberg, and Graepel]{balduzzi2019open}
D.~Balduzzi, M.~Garnelo, Y.~Bachrach, W.~Czarnecki, J.~Perolat, M.~Jaderberg,
  and T.~Graepel.
\newblock Open-ended learning in symmetric zero-sum games.
\newblock In \emph{International Conference on Machine Learning}, pages
  434--443. PMLR, 2019.

\bibitem[Barreto et~al.(2017)Barreto, Dabney, Munos, Hunt, Schaul, van Hasselt,
  and Silver]{barreto2017successor}
A.~Barreto, W.~Dabney, R.~Munos, J.~J. Hunt, T.~Schaul, H.~P. van Hasselt, and
  D.~Silver.
\newblock Successor features for transfer in reinforcement learning.
\newblock In \emph{Advances in neural information processing systems}, pages
  4055--4065, 2017.

\bibitem[Brown(1951)]{brown1951iterative}
G.~W. Brown.
\newblock Iterative solution of games by fictitious play.
\newblock \emph{Act. Anal. Prod Allocation}, 13\penalty0 (1):\penalty0 374,
  1951.

\bibitem[Brown and Sandholm(2019)]{brown2019superhuman}
N.~Brown and T.~Sandholm.
\newblock Superhuman ai for multiplayer poker.
\newblock \emph{Science}, 365\penalty0 (6456):\penalty0 885--890, 2019.

\bibitem[Carroll et~al.(2019)Carroll, Shah, Ho, Griffiths, Seshia, Abbeel, and
  Dragan]{carroll2019utility}
M.~Carroll, R.~Shah, M.~K. Ho, T.~Griffiths, S.~Seshia, P.~Abbeel, and
  A.~Dragan.
\newblock On the utility of learning about humans for human-ai coordination.
\newblock \emph{Advances in neural information processing systems}, 32, 2019.

\bibitem[Copeland(2021)]{10positions}
S.~Copeland.
\newblock 10 positions chess engines just don't understand, 2021.
\newblock URL
  \url{https://www.chess.com/article/view/10-positions-chess-engines-just-dont-understand}.

\bibitem[Cully and Demiris(2017)]{cully2017quality}
A.~Cully and Y.~Demiris.
\newblock Quality and diversity optimization: A unifying modular framework.
\newblock \emph{IEEE Transactions on Evolutionary Computation}, 22\penalty0
  (2):\penalty0 245--259, 2017.

\bibitem[Czarnecki et~al.(2020)Czarnecki, Gidel, Tracey, Tuyls, Omidshafiei,
  Balduzzi, and Jaderberg]{czarnecki2020real}
W.~M. Czarnecki, G.~Gidel, B.~Tracey, K.~Tuyls, S.~Omidshafiei, D.~Balduzzi,
  and M.~Jaderberg.
\newblock Real world games look like spinning tops.
\newblock \emph{Advances in Neural Information Processing Systems},
  33:\penalty0 17443--17454, 2020.

\bibitem[Doggers(2017)]{penrose}
P.~Doggers.
\newblock Will this position help understand human consciousness?, 2017.
\newblock URL
  \url{https://www.chess.com/news/view/will-this-position-help-to-understand-human-consciousness-4298}.

\bibitem[Dvoretsky(2010)]{dvoretsky2010dvoretsky}
M.~Dvoretsky.
\newblock \emph{Dvoretsky's endgame manual}.
\newblock SCB Distributors, 2010.

\bibitem[Eysenbach et~al.(2019)Eysenbach, Gupta, Ibarz, and
  Levine]{eysenbach2018diversity}
B.~Eysenbach, A.~Gupta, J.~Ibarz, and S.~Levine.
\newblock Diversity is all you need: Learning skills without a reward function.
\newblock In \emph{International Conference on Learning Representations}, 2019.
\newblock URL \url{https://openreview.net/forum?id=SJx63jRqFm}.

\bibitem[Fawzi et~al.(2022)Fawzi, Balog, Huang, Hubert, Romera-Paredes,
  Barekatain, Novikov, R~Ruiz, Schrittwieser, Swirszcz,
  et~al.]{fawzi2022discovering}
A.~Fawzi, M.~Balog, A.~Huang, T.~Hubert, B.~Romera-Paredes, M.~Barekatain,
  A.~Novikov, F.~J. R~Ruiz, J.~Schrittwieser, G.~Swirszcz, et~al.
\newblock Discovering faster matrix multiplication algorithms with
  reinforcement learning.
\newblock \emph{Nature}, 610\penalty0 (7930):\penalty0 47--53, 2022.

\bibitem[Flet-Berliac et~al.(2021)Flet-Berliac, Ferret, Pietquin, Preux, and
  Geist]{flet2021adversarially}
Y.~Flet-Berliac, J.~Ferret, O.~Pietquin, P.~Preux, and M.~Geist.
\newblock Adversarially guided actor-critic.
\newblock In \emph{International Conference on Learning Representations}, 2021.
\newblock URL \url{https://openreview.net/forum?id=_mQp5cr_iNy}.

\bibitem[Gidel et~al.(2018)Gidel, Pedregosa, and
  Lacoste-Julien]{gidel2018frank}
G.~Gidel, F.~Pedregosa, and S.~Lacoste-Julien.
\newblock Frank-wolfe splitting via augmented lagrangian method.
\newblock \emph{arXiv preprint arXiv:1804.03176}, 2018.

\bibitem[Gigerenzer(1991)]{gigerenzer1991tools}
G.~Gigerenzer.
\newblock From tools to theories: A heuristic of discovery in cognitive
  psychology.
\newblock \emph{Psychological review}, 98\penalty0 (2):\penalty0 254, 1991.

\bibitem[Gonz{\'a}lez-D{\'\i}az and
  Palacios-Huerta(2022)]{gonzalez2022alphazero}
J.~Gonz{\'a}lez-D{\'\i}az and I.~Palacios-Huerta.
\newblock Alphazero ideas, 2022.

\bibitem[Gregor et~al.(2017)Gregor, Rezende, and
  Wierstra]{gregor2016variational}
K.~Gregor, D.~J. Rezende, and D.~Wierstra.
\newblock Variational intrinsic control.
\newblock In \emph{5th International Conference on Learning Representations,
  {ICLR} 2017, Toulon, France, April 24-26, 2017, Workshop Track Proceedings}.
  OpenReview.net, 2017.
\newblock URL \url{https://openreview.net/forum?id=Skc-Fo4Yg}.

\bibitem[Guid and Bratko(2012)]{guid2012detecting}
M.~Guid and I.~Bratko.
\newblock Detecting fortresses in chess.
\newblock \emph{Elektrotehni{\v{s}}ki Vestnik}, 79\penalty0 (1-2):\penalty0
  35--40, 2012.

\bibitem[Hansen and Salamon(1990)]{hansen1990neural}
L.~K. Hansen and P.~Salamon.
\newblock Neural network ensembles.
\newblock \emph{IEEE transactions on pattern analysis and machine
  intelligence}, 12\penalty0 (10):\penalty0 993--1001, 1990.

\bibitem[Jaderberg et~al.(2017)Jaderberg, Dalibard, Osindero, Czarnecki,
  Donahue, Razavi, Vinyals, Green, Dunning, Simonyan, Fernando, and
  Kavukcuoglu]{jaderberg2017population}
M.~Jaderberg, V.~Dalibard, S.~Osindero, W.~M. Czarnecki, J.~Donahue, A.~Razavi,
  O.~Vinyals, T.~Green, I.~Dunning, K.~Simonyan, C.~Fernando, and
  K.~Kavukcuoglu.
\newblock Population based training of neural networks, 2017.

\bibitem[Kasparov(2017)]{deepthinking}
G.~Kasparov.
\newblock \emph{Deep Thinking: Where Machine Intelligence Ends and Human
  Creativity Begins}.
\newblock John Murray, 2017.

\bibitem[Kasparov(2018)]{chess_dorosopilia}
G.~Kasparov.
\newblock Chess, a <i>drosophila</i> of reasoning.
\newblock \emph{Science}, 362\penalty0 (6419):\penalty0 1087--1087, 2018.
\newblock \doi{10.1126/science.aaw2221}.
\newblock URL \url{https://www.science.org/doi/abs/10.1126/science.aaw2221}.

\bibitem[Kasparov and King(2000)]{kasparov2000kasparov}
G.~K. Kasparov and D.~King.
\newblock \emph{Kasparov against the world: the story of the greatest online
  challenge}.
\newblock KasparovChess Online, Incorporated, 2000.

\bibitem[Katz et~al.(2023)Katz, Bommarito, Gao, and Arredondo]{katz2023gpt}
D.~M. Katz, M.~J. Bommarito, S.~Gao, and P.~Arredondo.
\newblock Gpt-4 passes the bar exam, 2023.

\bibitem[Kemmerling et~al.(2023)Kemmerling, Lütticke, and
  Schmitt]{kemmerling2023beyond}
M.~Kemmerling, D.~Lütticke, and R.~H. Schmitt.
\newblock Beyond games: A systematic review of neural monte carlo tree search
  applications, 2023.

\bibitem[Kittler et~al.(1998)Kittler, Hatef, Duin, and
  Matas]{kittler1998combining}
J.~Kittler, M.~Hatef, R.~P. Duin, and J.~Matas.
\newblock On combining classifiers.
\newblock \emph{IEEE transactions on pattern analysis and machine
  intelligence}, 20\penalty0 (3):\penalty0 226--239, 1998.

\bibitem[Klissarov and Machado(2023)]{klissarov2023deep}
M.~Klissarov and M.~C. Machado.
\newblock Deep laplacian-based options for temporally-extended exploration.
\newblock \emph{International Conference on Machine Learning (ICML)}, 2023.

\bibitem[Kumar et~al.(2020)Kumar, Kumar, Levine, and Finn]{kumar2020one}
S.~Kumar, A.~Kumar, S.~Levine, and C.~Finn.
\newblock One solution is not all you need: Few-shot extrapolation via
  structured maxent rl.
\newblock \emph{Advances in Neural Information Processing Systems},
  33:\penalty0 8198--8210, 2020.

\bibitem[Lanctot et~al.(2017)Lanctot, Zambaldi, Gruslys, Lazaridou, Tuyls,
  P{\'e}rolat, Silver, and Graepel]{lanctot2017unified}
M.~Lanctot, V.~Zambaldi, A.~Gruslys, A.~Lazaridou, K.~Tuyls, J.~P{\'e}rolat,
  D.~Silver, and T.~Graepel.
\newblock A unified game-theoretic approach to multiagent reinforcement
  learning.
\newblock \emph{Advances in neural information processing systems}, 30, 2017.

\bibitem[Lehman and Stanley(2011)]{lehman2011evolving}
J.~Lehman and K.~O. Stanley.
\newblock Evolving a diversity of virtual creatures through novelty search and
  local competition.
\newblock In \emph{Proceedings of the 13th annual conference on Genetic and
  evolutionary computation}, pages 211--218, 2011.

\bibitem[Leslie and Collins(2006)]{leslie2006generalised}
D.~S. Leslie and E.~J. Collins.
\newblock Generalised weakened fictitious play.
\newblock \emph{Games and Economic Behavior}, 56\penalty0 (2):\penalty0
  285--298, 2006.

\bibitem[Lewis et~al.(2014)Lewis, Howes, and Singh]{lewis2014computational}
R.~L. Lewis, A.~Howes, and S.~Singh.
\newblock Computational rationality: Linking mechanism and behavior through
  bounded utility maximization.
\newblock \emph{Topics in cognitive science}, 6\penalty0 (2):\penalty0
  279--311, 2014.

\bibitem[Liu et~al.(2022{\natexlab{a}})Liu, Lanctot, Marris, and
  Heess]{pmlr-v162-liu22h}
S.~Liu, M.~Lanctot, L.~Marris, and N.~Heess.
\newblock Simplex neural population learning: Any-mixture {B}ayes-optimality in
  symmetric zero-sum games.
\newblock In K.~Chaudhuri, S.~Jegelka, L.~Song, C.~Szepesvari, G.~Niu, and
  S.~Sabato, editors, \emph{Proceedings of the 39th International Conference on
  Machine Learning}, volume 162 of \emph{Proceedings of Machine Learning
  Research}, pages 13793--13806. PMLR, 17--23 Jul 2022{\natexlab{a}}.
\newblock URL \url{https://proceedings.mlr.press/v162/liu22h.html}.

\bibitem[Liu et~al.(2022{\natexlab{b}})Liu, Marris, Hennes, Merel, Heess, and
  Graepel]{liu2022neupl}
S.~Liu, L.~Marris, D.~Hennes, J.~Merel, N.~Heess, and T.~Graepel.
\newblock Neu{PL}: Neural population learning.
\newblock In \emph{International Conference on Learning Representations},
  2022{\natexlab{b}}.
\newblock URL \url{https://openreview.net/forum?id=MIX3fJkl_1}.

\bibitem[Liu et~al.(2021)Liu, Jia, Wen, Hu, Chen, Fan, Hu, and
  Yang]{liu2021towards}
X.~Liu, H.~Jia, Y.~Wen, Y.~Hu, Y.~Chen, C.~Fan, Z.~Hu, and Y.~Yang.
\newblock Towards unifying behavioral and response diversity for open-ended
  learning in zero-sum games.
\newblock \emph{Advances in Neural Information Processing Systems},
  34:\penalty0 941--952, 2021.

\bibitem[Liu et~al.(2017)Liu, Ramachandran, Liu, and Peng]{liu2017stein}
Y.~Liu, P.~Ramachandran, Q.~Liu, and J.~Peng.
\newblock Stein variational policy gradient.
\newblock In \emph{33rd Conference on Uncertainty in Artificial Intelligence,
  UAI 2017}, 2017.

\bibitem[Lupu et~al.(2021)Lupu, Cui, Hu, and Foerster]{pmlr-v139-lupu21a}
A.~Lupu, B.~Cui, H.~Hu, and J.~Foerster.
\newblock Trajectory diversity for zero-shot coordination.
\newblock In M.~Meila and T.~Zhang, editors, \emph{Proceedings of the 38th
  International Conference on Machine Learning}, volume 139 of
  \emph{Proceedings of Machine Learning Research}, pages 7204--7213. PMLR,
  18--24 Jul 2021.
\newblock URL \url{https://proceedings.mlr.press/v139/lupu21a.html}.

\bibitem[Machado et~al.(2023)Machado, Barreto, Precup, and
  Bowling]{machado2023temporal}
M.~C. Machado, A.~Barreto, D.~Precup, and M.~Bowling.
\newblock Temporal abstraction in reinforcement learning with the successor
  representation.
\newblock \emph{Journal of Machine Learning Research}, 24\penalty0
  (80):\penalty0 1--69, 2023.

\bibitem[Mandhane et~al.(2022)Mandhane, Zhernov, Rauh, Gu, Wang, Xue, Shang,
  Pang, Claus, Chiang, Chen, Han, Chen, Mankowitz, Broshear, Schrittwieser,
  Hubert, Vinyals, and Mann]{mandhane2022muzero}
A.~Mandhane, A.~Zhernov, M.~Rauh, C.~Gu, M.~Wang, F.~Xue, W.~Shang, D.~Pang,
  R.~Claus, C.-H. Chiang, C.~Chen, J.~Han, A.~Chen, D.~J. Mankowitz,
  J.~Broshear, J.~Schrittwieser, T.~Hubert, O.~Vinyals, and T.~Mann.
\newblock Muzero with self-competition for rate control in vp9 video
  compression, 2022.

\bibitem[Masood and Doshi-Velez(2019)]{masood2019diversity}
M.~Masood and F.~Doshi-Velez.
\newblock Diversity-inducing policy gradient: Using maximum mean discrepancy to
  find a set of diverse policies.
\newblock In \emph{Proceedings of the Twenty-Eighth International Joint
  Conference on Artificial Intelligence, {IJCAI-19}}, pages 5923--5929.
  International Joint Conferences on Artificial Intelligence Organization, 7
  2019.
\newblock \doi{10.24963/ijcai.2019/821}.
\newblock URL \url{https://doi.org/10.24963/ijcai.2019/821}.

\bibitem[McGrath et~al.(2022)McGrath, Kapishnikov, Toma{\vv{s}}ev, Pearce,
  Wattenberg, Hassabis, Kim, Paquet, and Kramnik]{mcgrath2022acquisition}
T.~McGrath, A.~Kapishnikov, N.~Toma{\vv{s}}ev, A.~Pearce, M.~Wattenberg,
  D.~Hassabis, B.~Kim, U.~Paquet, and V.~Kramnik.
\newblock Acquisition of chess knowledge in alphazero.
\newblock \emph{Proceedings of the National Academy of Sciences}, 119\penalty0
  (47):\penalty0 e2206625119, 2022.

\bibitem[McIlroy-Young et~al.(2020)McIlroy-Young, Sen, Kleinberg, and
  Anderson]{mcilroy2020aligning}
R.~McIlroy-Young, S.~Sen, J.~Kleinberg, and A.~Anderson.
\newblock Aligning superhuman ai with human behavior: Chess as a model system.
\newblock \emph{Proceedings of the 26th ACM SIGKDD International Conference on
  Knowledge Discovery \& Data Mining}, pages 1677--1687, 2020.

\bibitem[Metz(2016)]{move37b}
C.~Metz.
\newblock Alphago versus lee sedol, 2016.
\newblock URL
  \url{https://www.wired.com/2016/03/googles-ai-wins-pivotal-game-two-match-go-grandmaster/}.

\bibitem[Mnih et~al.(2015)Mnih, Kavukcuoglu, Silver, Rusu, Veness, Bellemare,
  Graves, Riedmiller, Fidjeland, Ostrovski, et~al.]{mnih2015human}
V.~Mnih, K.~Kavukcuoglu, D.~Silver, A.~A. Rusu, J.~Veness, M.~G. Bellemare,
  A.~Graves, M.~Riedmiller, A.~K. Fidjeland, G.~Ostrovski, et~al.
\newblock Human-level control through deep reinforcement learning.
\newblock \emph{nature}, 518\penalty0 (7540):\penalty0 529--533, 2015.

\bibitem[Morav{\v{c}}{\'\i}k et~al.(2017)Morav{\v{c}}{\'\i}k, Schmid, Burch,
  Lis{\`y}, Morrill, Bard, Davis, Waugh, Johanson, and
  Bowling]{moravvcik2017deepstack}
M.~Morav{\v{c}}{\'\i}k, M.~Schmid, N.~Burch, V.~Lis{\`y}, D.~Morrill, N.~Bard,
  T.~Davis, K.~Waugh, M.~Johanson, and M.~Bowling.
\newblock Deepstack: Expert-level artificial intelligence in heads-up no-limit
  poker.
\newblock \emph{Science}, 356\penalty0 (6337):\penalty0 508--513, 2017.

\bibitem[Mouret and Clune(2015)]{mouret2015illuminating}
J.~Mouret and J.~Clune.
\newblock Illuminating search spaces by mapping elites.
\newblock \emph{CoRR}, abs/1504.04909, 2015.
\newblock URL \url{http://arxiv.org/abs/1504.04909}.

\bibitem[OpenAI(2023)]{openai2023gpt4}
OpenAI.
\newblock Gpt-4 technical report, 2023.

\bibitem[Osband et~al.(2016)Osband, Blundell, Pritzel, and
  Van~Roy]{osband2016deep}
I.~Osband, C.~Blundell, A.~Pritzel, and B.~Van~Roy.
\newblock Deep exploration via bootstrapped dqn.
\newblock \emph{Advances in neural information processing systems}, 29, 2016.

\bibitem[Page(2019)]{page2019diversity}
S.~E. Page.
\newblock \emph{The diversity bonus: How great teams pay off in the knowledge
  economy}.
\newblock Princeton University Press, 2019.

\bibitem[Parker-Holder et~al.(2020)Parker-Holder, Pacchiano, Choromanski, and
  Roberts]{parker2020effective}
J.~Parker-Holder, A.~Pacchiano, K.~M. Choromanski, and S.~J. Roberts.
\newblock Effective diversity in population based reinforcement learning.
\newblock \emph{Advances in Neural Information Processing Systems}, 33, 2020.

\bibitem[Peer et~al.(2021)Peer, Tessler, Merlis, and Meir]{peer2021ensemble}
O.~Peer, C.~Tessler, N.~Merlis, and R.~Meir.
\newblock Ensemble bootstrapping for q-learning.
\newblock \emph{International Conference on Machine Learning}, pages
  8454--8463, 2021.

\bibitem[Penrose(1994)]{penrose1994shadows}
R.~Penrose.
\newblock \emph{Shadows of the Mind}, volume~4.
\newblock Oxford University Press Oxford, 1994.

\bibitem[Penrose and Mermin(1990)]{penrose1990emperor}
R.~Penrose and N.~D. Mermin.
\newblock The emperor’s new mind: Concerning computers, minds, and the laws
  of physics, 1990.

\bibitem[Penrose(2020)]{penrose2}
S.~R. Penrose.
\newblock Sir roger penrose and dr. stuart hameroff: Consciousness and the
  physics of the brain, 2020.
\newblock URL \url{https://www.youtube.com/watch?v=xGbgDf4HCHU}.

\bibitem[Perez et~al.(2018)Perez, Strub, De~Vries, Dumoulin, and
  Courville]{perez2018film}
E.~Perez, F.~Strub, H.~De~Vries, V.~Dumoulin, and A.~Courville.
\newblock Film: Visual reasoning with a general conditioning layer.
\newblock \emph{Proceedings of the AAAI conference on artificial intelligence},
  32\penalty0 (1), 2018.

\bibitem[Polikar(2006)]{polikar2006ensemble}
R.~Polikar.
\newblock Ensemble based systems in decision making.
\newblock \emph{IEEE Circuits and systems magazine}, 6\penalty0 (3):\penalty0
  21--45, 2006.

\bibitem[Pugh et~al.(2016)Pugh, Soros, and Stanley]{pugh2016quality}
J.~K. Pugh, L.~B. Soros, and K.~O. Stanley.
\newblock Quality diversity: A new frontier for evolutionary computation.
\newblock \emph{Frontiers in Robotics and AI}, 3:\penalty0 40, 2016.

\bibitem[Rhoads(2019)]{Troitsky}
G.~C. Rhoads.
\newblock What are some of the strangest chess puzzles you've ever seen?, 2019.
\newblock URL
  \url{https://www.quora.com/What-are-some-of-the-strangest-chess-puzzles-youve-ever-seen/answer/Glenn-C-Rhoads?ch=10&oid=120233484&share=942d35cc&target_type=answer}.

\bibitem[Rosin(2011)]{rosin2011multi}
C.~D. Rosin.
\newblock Multi-armed bandits with episode context.
\newblock \emph{Annals of Mathematics and Artificial Intelligence}, 61\penalty0
  (3):\penalty0 203--230, 2011.

\bibitem[Sanjaya et~al.(2022)Sanjaya, Wang, and Yang]{sanjaya2022measuring}
R.~Sanjaya, J.~Wang, and Y.~Yang.
\newblock Measuring the non-transitivity in chess.
\newblock \emph{Algorithms}, 15\penalty0 (5):\penalty0 152, 2022.

\bibitem[Schrittwieser et~al.(2020)Schrittwieser, Antonoglou, Hubert, Simonyan,
  Sifre, Schmitt, Guez, Lockhart, Hassabis, Graepel,
  et~al.]{schrittwieser2020mastering}
J.~Schrittwieser, I.~Antonoglou, T.~Hubert, K.~Simonyan, L.~Sifre, S.~Schmitt,
  A.~Guez, E.~Lockhart, D.~Hassabis, T.~Graepel, et~al.
\newblock Mastering atari, go, chess and shogi by planning with a learned
  model.
\newblock \emph{Nature}, 588\penalty0 (7839):\penalty0 604--609, 2020.

\bibitem[Shead(2020)]{demis}
S.~Shead.
\newblock How deepmind boss demis hassabis used chess to get billionaire peter
  thiel to ‘take notice’ of his ai lab, 2020.
\newblock URL
  \url{https://www.cnbc.com/2020/12/07/deepminds-demis-hassabis-used-chess-to-get-peter-thiels-attention.html}.

\bibitem[Shin et~al.(2023)Shin, Kim, van Opheusden, and
  Griffiths]{shin2023superhuman}
M.~Shin, J.~Kim, B.~van Opheusden, and T.~L. Griffiths.
\newblock Superhuman artificial intelligence can improve human decision-making
  by increasing novelty.
\newblock \emph{Proceedings of the National Academy of Sciences}, 120\penalty0
  (12):\penalty0 e2214840120, 2023.

\bibitem[Silver et~al.(2017)Silver, Schrittwieser, Simonyan, Antonoglou, Huang,
  Guez, Hubert, Baker, Lai, Bolton, et~al.]{silver2017mastering}
D.~Silver, J.~Schrittwieser, K.~Simonyan, I.~Antonoglou, A.~Huang, A.~Guez,
  T.~Hubert, L.~Baker, M.~Lai, A.~Bolton, et~al.
\newblock Mastering the game of go without human knowledge.
\newblock \emph{nature}, 550\penalty0 (7676):\penalty0 354--359, 2017.

\bibitem[Silver et~al.(2018{\natexlab{a}})Silver, Hubert, Schrittwieser,
  Antonoglou, Lai, Guez, Lanctot, Sifre, Kumaran, Graepel, Lillicrap, Simonyan,
  and Hassabis]{az}
D.~Silver, T.~Hubert, J.~Schrittwieser, I.~Antonoglou, M.~Lai, A.~Guez,
  M.~Lanctot, L.~Sifre, D.~Kumaran, T.~Graepel, T.~Lillicrap, K.~Simonyan, and
  D.~Hassabis.
\newblock A general reinforcement learning algorithm that masters chess, shogi,
  and go through self-play.
\newblock \emph{Science}, 362\penalty0 (6419):\penalty0 1140--1144,
  2018{\natexlab{a}}.
\newblock \doi{10.1126/science.aar6404}.
\newblock URL \url{https://www.science.org/doi/abs/10.1126/science.aar6404}.

\bibitem[Silver et~al.(2018{\natexlab{b}})Silver, Hubert, Schrittwieser,
  Antonoglou, Lai, Guez, Lanctot, Sifre, Kumaran, Graepel,
  et~al.]{silver2018general}
D.~Silver, T.~Hubert, J.~Schrittwieser, I.~Antonoglou, M.~Lai, A.~Guez,
  M.~Lanctot, L.~Sifre, D.~Kumaran, T.~Graepel, et~al.
\newblock A general reinforcement learning algorithm that masters chess, shogi,
  and go through self-play.
\newblock \emph{Science}, 362\penalty0 (6419):\penalty0 1140--1144,
  2018{\natexlab{b}}.

\bibitem[Singhal et~al.(2023)Singhal, Tu, Gottweis, Sayres, Wulczyn, Hou,
  Clark, Pfohl, Cole-Lewis, Neal, Schaekermann, Wang, Amin, Lachgar, Mansfield,
  Prakash, Green, Dominowska, y~Arcas, Tomasev, Liu, Wong, Semturs, Mahdavi,
  Barral, Webster, Corrado, Matias, Azizi, Karthikesalingam, and
  Natarajan]{singhal2023expertlevel}
K.~Singhal, T.~Tu, J.~Gottweis, R.~Sayres, E.~Wulczyn, L.~Hou, K.~Clark,
  S.~Pfohl, H.~Cole-Lewis, D.~Neal, M.~Schaekermann, A.~Wang, M.~Amin,
  S.~Lachgar, P.~Mansfield, S.~Prakash, B.~Green, E.~Dominowska, B.~A. y~Arcas,
  N.~Tomasev, Y.~Liu, R.~Wong, C.~Semturs, S.~S. Mahdavi, J.~Barral,
  D.~Webster, G.~S. Corrado, Y.~Matias, S.~Azizi, A.~Karthikesalingam, and
  V.~Natarajan.
\newblock Towards expert-level medical question answering with large language
  models, 2023.

\bibitem[Steingrimsson(2021)]{steingrimsson2021chess}
H.~Steingrimsson.
\newblock Chess fortresses, a causal test for state of the art symbolic [neuro]
  architectures.
\newblock In \emph{2021 IEEE Conference on Games (CoG)}, pages 1--8. IEEE,
  2021.

\bibitem[Stokes(2005)]{stokes2005creativity}
P.~D. Stokes.
\newblock \emph{Creativity from constraints: The psychology of breakthrough}.
\newblock Springer Publishing Company, 2005.

\bibitem[Strouse et~al.(2021)Strouse, McKee, Botvinick, Hughes, and
  Everett]{NEURIPS2021_797134c3}
D.~Strouse, K.~McKee, M.~Botvinick, E.~Hughes, and R.~Everett.
\newblock Collaborating with humans without human data.
\newblock \emph{Advances in Neural Information Processing Systems},
  34:\penalty0 14502--14515, 2021.
\newblock URL
  \url{https://proceedings.neurips.cc/paper_files/paper/2021/file/797134c3e42371bb4979a462eb2f042a-Paper.pdf}.

\bibitem[Sutton and Barto(2018)]{sutton2018reinforcement}
R.~S. Sutton and A.~G. Barto.
\newblock \emph{Reinforcement learning: An introduction}.
\newblock MIT press, 2018.

\bibitem[Timbers et~al.(2022)Timbers, Bard, Lockhart, Lanctot, Schmid, Burch,
  Schrittwieser, Hubert, and Bowling]{timbers2022approximate}
F.~Timbers, N.~Bard, E.~Lockhart, M.~Lanctot, M.~Schmid, N.~Burch,
  J.~Schrittwieser, T.~Hubert, and M.~Bowling.
\newblock Approximate exploitability: learning a best response.
\newblock In \emph{Proceedings of the International Joint Conference on
  Artificial Intelligence (IJCAI)}, pages 3487--3493, 2022.

\bibitem[Toma{\vv{s}}ev et~al.(2020)Toma{\vv{s}}ev, Paquet, Hassabis, and
  Kramnik]{tomavsev2020assessing}
N.~Toma{\vv{s}}ev, U.~Paquet, D.~Hassabis, and V.~Kramnik.
\newblock Assessing game balance with alphazero: Exploring alternative rule
  sets in chess, 2020.

\bibitem[Turing(1948)]{Turing}
A.~Turing.
\newblock Intelligent machinery.
\newblock \emph{Philosophia Mathematica}, page 395, 1948.

\bibitem[Van~Hasselt et~al.(2016)Van~Hasselt, Guez, and Silver]{van2016deep}
H.~Van~Hasselt, A.~Guez, and D.~Silver.
\newblock Deep reinforcement learning with double q-learning.
\newblock In \emph{Proceedings of the AAAI conference on artificial
  intelligence}, volume~30, 2016.

\bibitem[Vassiliades et~al.(2017)Vassiliades, Chatzilygeroudis, and
  Mouret]{vassiliades2016scaling}
V.~Vassiliades, K.~Chatzilygeroudis, and J.-B. Mouret.
\newblock Using centroidal voronoi tessellations to scale up the
  multi-dimensional archive of phenotypic elites algorithm, 2017.

\bibitem[Vinyals et~al.(2019)Vinyals, Babuschkin, Chung, Mathieu, Jaderberg,
  Czarnecki, Dudzik, Huang, Georgiev, Powell, et~al.]{vinyals2019alphastar}
O.~Vinyals, I.~Babuschkin, J.~Chung, M.~Mathieu, M.~Jaderberg, W.~M. Czarnecki,
  A.~Dudzik, A.~Huang, P.~Georgiev, R.~Powell, et~al.
\newblock Alphastar: Mastering the real-time strategy game starcraft ii.
\newblock \emph{DeepMind blog}, 2:\penalty0 20, 2019.

\bibitem[Wang et~al.(2022)Wang, Gleave, Belrose, Tseng, Miller, Dennis, Duan,
  Pogrebniak, Levine, and Russell]{wang2023adversarial}
T.~T. Wang, A.~Gleave, N.~Belrose, T.~Tseng, J.~Miller, M.~D. Dennis, Y.~Duan,
  V.~Pogrebniak, S.~Levine, and S.~Russell.
\newblock Adversarial policies beat professional-level go ais.
\newblock \emph{arXiv preprint arXiv:2211.00241}, 2022.

\bibitem[Zahavy et~al.(2020)Zahavy, Hasidim, Kaplan, and
  Mansour]{zahavy2020planning}
T.~Zahavy, A.~Hasidim, H.~Kaplan, and Y.~Mansour.
\newblock Planning in hierarchical reinforcement learning: Guarantees for using
  local policies.
\newblock In \emph{Algorithmic Learning Theory}, pages 906--934, 2020.

\bibitem[Zahavy et~al.(2021{\natexlab{a}})Zahavy, Barreto, Mankowitz, Hou,
  O'Donoghue, Kemaev, and Singh]{zahavy2021discovering}
T.~Zahavy, A.~Barreto, D.~J. Mankowitz, S.~Hou, B.~O'Donoghue, I.~Kemaev, and
  S.~Singh.
\newblock Discovering a set of policies for the worst case reward.
\newblock In \emph{International Conference on Learning Representations},
  2021{\natexlab{a}}.
\newblock URL \url{https://openreview.net/forum?id=PUkhWz65dy5}.

\bibitem[Zahavy et~al.(2021{\natexlab{b}})Zahavy, O'Donoghue, Desjardins, and
  Singh]{zahavy2021reward}
T.~Zahavy, B.~O'Donoghue, G.~Desjardins, and S.~Singh.
\newblock Reward is enough for convex mdps.
\newblock \emph{Advances in Neural Information Processing Systems},
  34:\penalty0 25746--25759, 2021{\natexlab{b}}.

\bibitem[Zahavy et~al.(2023)Zahavy, Schroecker, Behbahani, Baumli, Flennerhag,
  Hou, and Singh]{zahavy2022discovering}
T.~Zahavy, Y.~Schroecker, F.~Behbahani, K.~Baumli, S.~Flennerhag, S.~Hou, and
  S.~Singh.
\newblock Discovering policies with {DOM}i{NO}: Diversity optimization
  maintaining near optimality.
\newblock In \emph{The Eleventh International Conference on Learning
  Representations}, 2023.
\newblock URL \url{https://openreview.net/forum?id=kjkdzBW3b8p}.

\end{thebibliography}
\end{document}